\pdfoutput=1
\documentclass[11pt]{article}

\usepackage[preprint]{acl}

\usepackage{times}
\usepackage{latexsym}

\usepackage[T1]{fontenc}
\usepackage[utf8]{inputenc}

\usepackage{microtype}

\usepackage{inconsolata}

\usepackage{graphicx}
\usepackage{subcaption}

\usepackage{tikz}
\usepackage{booktabs}
\usepackage{pifont}
\usepackage{array}
\usepackage{multirow}
\usepackage{colortbl}
\usepackage{makecell}
\usepackage{diagbox}
\usepackage{xcolor}
\usepackage{graphicx}
\usepackage{threeparttable}
\usepackage[most]{tcolorbox}
\usepackage{algorithm}
\usepackage{algpseudocode}
\usepackage{hyperref}
\usepackage{xurl}
\usepackage{xspace}
\newcommand{\bench}{\textsc{\textbf{VGI-Bench}}\xspace}
\newcommand{\benchtitle}{\texorpdfstring{\textcolor{benchsky}{\textsc{\textbf{VGI-Bench}}}}{VGI-Bench}}

\definecolor{Green}{RGB}{110,140,120}
\definecolor{GreenLight}{RGB}{242,247,243}
\definecolor{Blue}{RGB}{100,125,150}
\definecolor{BlueLight}{RGB}{242,245,249}

\definecolor{Yellow}{RGB}{170,150,95}
\definecolor{YellowLight}{RGB}{249,247,240}

\definecolor{Red}{RGB}{145,104,104}
\definecolor{RedLight}{RGB}{248,243,243}

\definecolor{Purple}{RGB}{118,108,140}
\definecolor{PurpleLight}{RGB}{245,243,248}

\definecolor{hlgreen}{RGB}{92,158,84}
\definecolor{hlred}{RGB}{165,100,95}

\definecolor{hlbestb}{HTML}{B6CBE4}
\definecolor{hlsecb}{HTML}{E4EDF7}

\providecolor{piegreen}{RGB}{102,166,142}
\providecolor{piegrey}{RGB}{140,140,140}
\providecolor{piered}{RGB}{204,92,92}
\providecolor{pieblue}{RGB}{120,160,195}
\providecolor{pieyellow}{RGB}{214,157,62}  % muted academic yellow / amber

\newcommand{\cmark}{\textcolor{pieblue}{\Large\ding{51}}}
\newcommand{\xmark}{\textcolor{piered}{\Large\ding{55}}}

\newcommand{\pmark}{%
  \textcolor{pieyellow}{%
    \ooalign{%
      \hidewidth\raisebox{0.15ex}{\small\ding{55}}\hidewidth\cr
      \Large\ding{51}\cr
    }%
  }%
}

\usepackage{fontawesome5}
\providecolor{darkblue}{rgb}{0, 0, 0.5}
\colorlet{projectlink}{darkblue}             % = acl.sty's urlcolor
\newcommand{\globeicon}{%
  \raisebox{0.03em}{\scalebox{0.81}{\faGlobe}}}
\newcommand{\githubicon}{%
  \raisebox{-0.05em}{\includegraphics[height=0.79em]{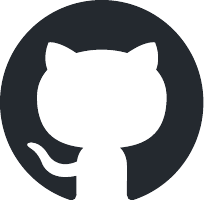}}}
\newcommand{\hficon}{%
  \raisebox{-0.08em}{\includegraphics[height=0.85em]{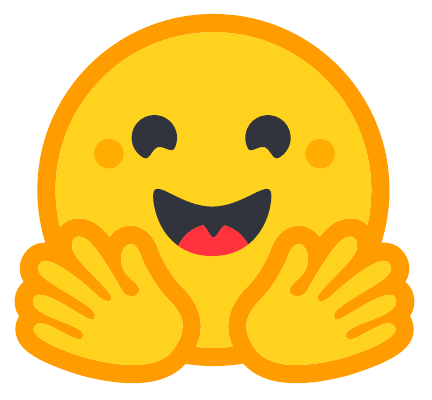}}}

\makeatletter
\tcbset{contbreak/.style={%
  title after break={\kvtcb@title\ \mbox{\normalfont\itshape(continued)}}}}

\makeatother

\newcommand{\boxednum}[1]{\tikz[baseline=(bn.base)]{%
  \node[draw, rectangle, inner sep=1.6pt, line width=0.6pt] (bn) {\footnotesize\textbf{#1}};}}

\newcommand{\catOne}{\raisebox{-0.10ex}{\large\ding{171}}}
\newcommand{\catTwo}{\raisebox{-0.10ex}{\large\ding{168}}}
\newcommand{\catThree}{\raisebox{-0.10ex}{\large\ding{110}}}

\DeclareFontFamily{T1}{optimistic}{}
\DeclareFontShape{T1}{optimistic}{m}{n}{<-> s * [0.88] optimistic}{}
\DeclareFontShape{T1}{optimistic}{b}{n}{<-> s * [0.88] optimistic}{}
\DeclareFontShape{T1}{optimistic}{bx}{n}{<-> ssub * optimistic/b/n}{}
\DeclareFontShape{T1}{optimistic}{m}{sc}{<-> ssub * optimistic/m/n}{}
\DeclareFontShape{T1}{optimistic}{b}{sc}{<-> ssub * optimistic/b/n}{}
\DeclareFontShape{T1}{optimistic}{bx}{sc}{<-> ssub * optimistic/b/n}{}
\pdfmapline{+optimistic < Optimistic.ttf <T1-WGL4.enc}
\newcommand{\optimistic}{\fontfamily{optimistic}\selectfont}

\newcommand{\fauxboldwidth}{0.35}
\newcommand{\fauxbold}[1]{%
  \pdfliteral{2 Tr \fauxboldwidth\space w}#1\pdfliteral{0 Tr}}

\newcommand{\afflfont}{\fontsize{10pt}{12pt}\selectfont}

\definecolor{metafg}{HTML}{1C2B33}
\definecolor{metabg}{HTML}{E9EEF5}
\definecolor{benchsky}{HTML}{0A85C4}

\newtcolorbox{titlecard}{
  enhanced, frame hidden, colback=metabg,
  arc=10pt, boxsep=0pt,
  left=0.55cm, right=0.55cm, top=0.5cm, bottom=0.5cm,
  width=\textwidth,
  grow to left by=0.55cm, grow to right by=0.55cm,
}

\makeatletter
\newcommand{\@abstracttext}{}
\newcommand{\abstracttext}[1]{\gdef\@abstracttext{#1}}

\AtBeginDocument{%
  \def\@maketitle{%
    \begin{titlecard}
      \color{metafg}
      \centering
      {\optimistic\fontsize{16pt}{19pt}\selectfont \fauxbold{\@title} \par}
      \vskip 0.16in
      {\fontsize{10.8pt}{13pt}\selectfont
       \renewcommand{\textbf}[1]{{\optimistic ##1}}%
       \begin{tabular}[t]{c}\@author\end{tabular}\par}
      \vskip 0.20in
      \begin{minipage}{\dimexpr\linewidth-0.5cm\relax}
        \normalsize\@setsize\normalsize{12pt}\xpt\@xpt
        \setlength{\parindent}{0pt}
        \@abstracttext
      \end{minipage}
      \vskip 0.02in
    \end{titlecard}%
  }%
}
\makeatother
\title{\benchtitle: Probing Visual Intelligence in Video Generation Models}

\author{
 \textbf{Xuan He\textsuperscript{1 $\dagger$ $\S$}},\enspace
 \textbf{Cong Wei\textsuperscript{3,6 $\dagger$}},\enspace
 \textbf{Yuhao Cheng\textsuperscript{1 $\dagger$}},\enspace
 \textbf{Linrui Ma\textsuperscript{2,4 $\dagger$}},\enspace
 \textbf{Yuxuan Zhang\textsuperscript{5,6,10 $\dagger$}},
\\
 \textbf{Zuojun Li\textsuperscript{2}},\enspace
 \textbf{Yuhao Wen\textsuperscript{2}},\enspace
  \textbf{Jize Jiang\textsuperscript{1}},\enspace
 \textbf{Zeyi Liu\textsuperscript{1}},\enspace
 \textbf{Yuren Hao\textsuperscript{1}},\enspace
 \textbf{Songcheng Cai\textsuperscript{3,6}},
\\
\textbf{Keming Wu\textsuperscript{2}},\enspace
 \textbf{Penghui Du\textsuperscript{10}},\enspace
 \textbf{Kai Zou\textsuperscript{9}},\enspace
 \textbf{Rui Yang\textsuperscript{1}},\enspace
 \textbf{Chenkai Sun\textsuperscript{8}},\enspace
 \textbf{Ke Yang\textsuperscript{1,7}},\enspace
 \textbf{Ping Nie\textsuperscript{3}},
\\
 \textbf{Kelsey R. Allen\textsuperscript{5,6}},\enspace
 \textbf{Chenglong Wang\textsuperscript{7}},\enspace
 \textbf{Michel Galley\textsuperscript{7}},\enspace
 \textbf{Jianfeng Gao\textsuperscript{7}},\enspace
 \textbf{ChengXiang Zhai\textsuperscript{1}}
\\
\\
 \afflfont
 \textsuperscript{1}University of Illinois Urbana Champaign,\enspace
 \textsuperscript{2}Tsinghua University,\enspace
 \textsuperscript{3}University of Waterloo,
 \\
 \afflfont
 \textsuperscript{4}Massachusetts Institute of Technology,\enspace
 \textsuperscript{5}University of British Columbia,\enspace
 \textsuperscript{6}Vector Institute,
 \\
 \afflfont
 \textsuperscript{7}Microsoft Research,\enspace
 \textsuperscript{8}Independent,\enspace
 \textsuperscript{9}NetMind.ai,\enspace
 \textsuperscript{10}Etude AI,
\\[0.5em]
{\normalsize\color{projectlink}
\href{https://hexuan21.github.io/VGI-Bench/}{\globeicon\ Project Page}
\quad
\href{https://huggingface.co/datasets/hexuan21/VGI-Bench}{\hficon\ Data}
\quad
\href{https://github.com/hexuan21/VGI-Bench}{\githubicon\ Code}
}
}

\abstracttext{%
Recent studies suggest that video generation models can exhibit certain forms of zero-shot visual reasoning through generated frames. Yet reliable evaluation remains challenging: benchmarks should adopt inputs aligned with the visual priors of current video models, require valid evolving processes rather than only plausible final states, and calibrate task difficulty to remain challenging yet partly feasible. 
To this end, we introduce \bench, containing 27 tasks and 810 instances, organized by a two-level taxonomy of task domains and skill tags for fine-grained evaluation of visual reasoning capabilities of video generation models.
Our evaluations show that current generative systems can solve a subset of visually grounded reasoning tasks, but remain far from reliable, with even the strongest model, Seedance~2.0, achieving only 51.0\% under our evaluation criteria.
Our analysis further explore the output failure modes, input condition sensitivity, performance transfer boundary from synthetic fine-tuning, and internal denoising perspective revealing limited self-correction, where later steps mainly refine early hypotheses rather than correct reasoning errors. 
We hope \bench will help stimulate the development of next-generation video generation models.
}

\begin{document}

% \twocolumn[{%
%     \renewcommand\twocolumn[1][]{#1}
%     \maketitle
%     \centering
%     \vspace{0.5em}
%     \begin{center}
%         \centering
%         \includegraphics[width=\linewidth]{Figures/teaser.pdf}
%         \vspace{0.5em}
%         \captionof{figure}{Overview of representative tasks in our benchmark. \bench adopts a \textbf{double-level taxonomy}: the first level groups every task into one of four mutually exclusive task \textbf{domains}, 
%         while the second level annotates each task with one or more (non-exclusive) \textbf{skill tags}, as shown in the legend at the bottom. Each panel shows a representative real-scene input image; the icons under the task name encode that task's skill tags.}
%         \label{fig:teaser}
%     \end{center}
% }]

\maketitle
\begin{figure*}[ht]
    \centering
    \includegraphics[width=0.9\linewidth]{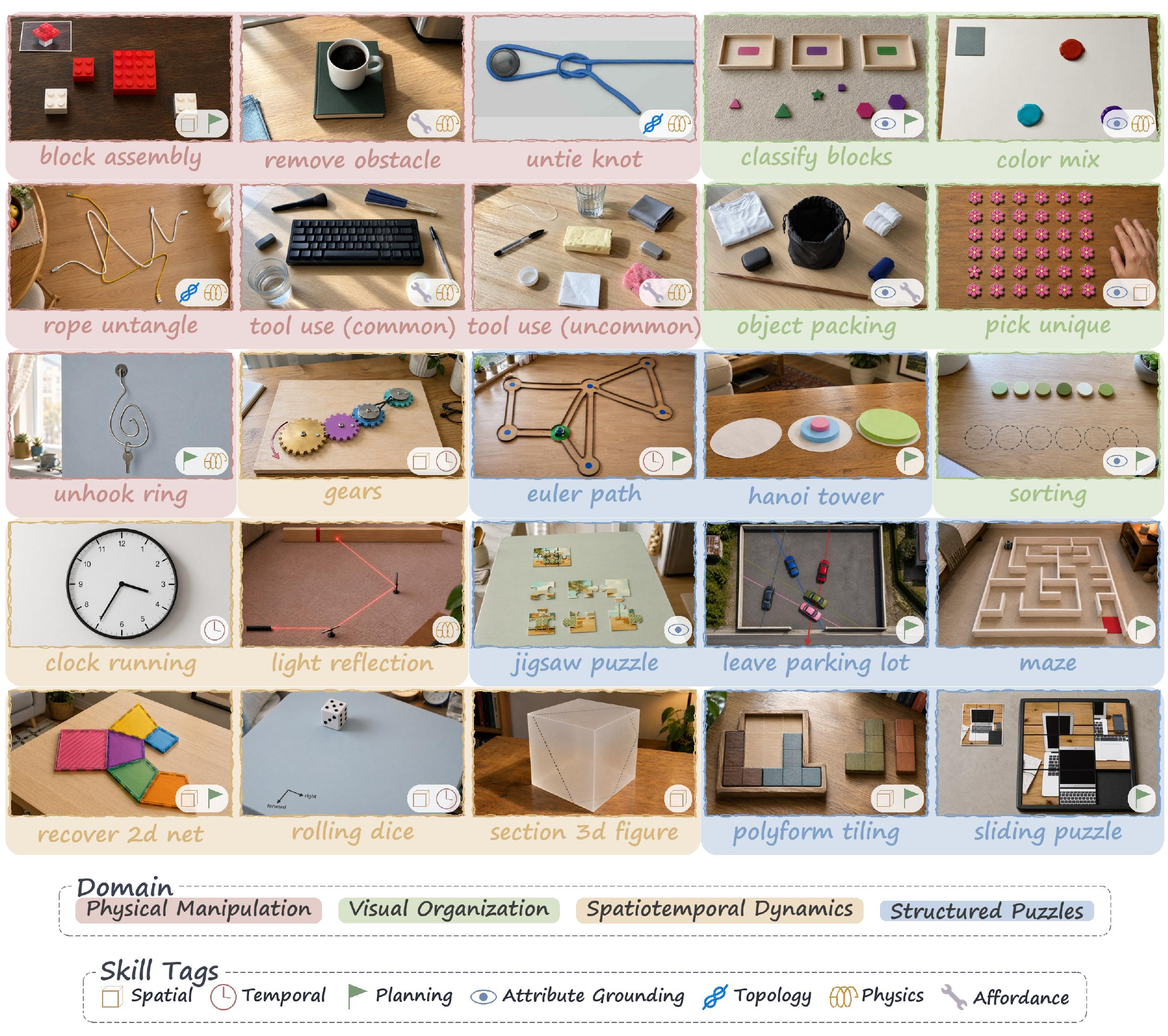}
    \caption{Overview of representative tasks in our benchmark. \bench adopts a \textbf{double-level taxonomy}: the first level groups every task into one of four mutually exclusive task \textbf{domains}, 
    while the second level annotates each task with one or more (non-exclusive) \textbf{skill tags}, as shown in the legend at the bottom. Each panel shows a representative real-scene input image; the icons under the task name encode that task's skill tags.}
    \label{fig:teaser}
\end{figure*}

\begingroup
\renewcommand{\thefootnote}{}
\footnotetext{$\dagger$ Main Contributor. $\S$ Project Lead.}
\endgroup

\definecolor{clsPhysical}{RGB}{239,224,222}
\definecolor{clsVisual}{RGB}{231,238,222}
\definecolor{clsSpatio}{RGB}{244,234,219}
\definecolor{clsStructured}{RGB}{219,228,239}

\section{Introduction}
\label{1_intro}

video generation models are increasingly viewed as visual world simulators~\citep{openai2024sora,  bruce2024genie, qin2024worldsimbench, huang2025vid2world}, capable of synthesizing plausible evolutions of visual scenes. Recent studies~\citep{wiedemer2025video,tong2025thinking} further suggest video generation models may encode more than low-level world priors like appearances and motions: they can acquire structured representations of spatial-temporal relations, rule constraints, and action-outcome dependencies from large-scale training, enabling certain forms of zero-shot visual reasoning. This has motivated a broader view of video generation models, from \textbf{passive visual simulators} toward \textbf{potential visual reasoners} that express reasoning through frame sequence. Beyond video synthesis itself, such emergent visual intelligence suggests its potential as vision foundation models, with implications for tasks including scene understanding, controllable editing, and downstream embodied learning~\citep{gabeur2026image, wang2026videogenerationmodelsgeneralpurpose, yang2025unified, zheng2026v, liang2025video, gao2026dreamdojo, ye2026worldactionmodelszeroshot}. 
These developments raise two key questions: how much visual intelligence is encoded in current video generation models, and how well these models can solve downstream tasks by “imagining” their step-by-step progression through generated video frames.

% =====================================================================
% Tables/related_works.tex  ---  Comparison of video-generation benchmarks.
%
% REQUIRED PREAMBLE (add to acl_latex.tex):
%   \usepackage{tikz}
%   \usepackage{booktabs}
%   \usepackage{pifont}
%   \usepackage{array}
%
% Then include with:  \input{Tables/related_works}
% =====================================================================

% Fallback for \bench — define this in your preamble (e.g.,
%   \newcommand{\bench}{YourBenchName}) to override.
\providecommand{\bench}{OurBench}

% ----- Pie chart  \pie{n}{d}  (n/d green, rest red) ------------------
% 0/d => all red (nothing satisfied); d/d => all green (fully satisfied)
\providecommand{\pie}[2]{%
  \tikz[baseline=-1.0ex]{%
    \def\r{1.3ex}%
    \pgfmathsetmacro{\stopangle}{90 - 360*(#1/#2)}%
    \ifdim#1pt=0pt\relax
      \fill[piered] (0,0) circle (\r);%
    \else
      \ifnum#1=#2\relax
        \fill[pieblue] (0,0) circle (\r);%
      \else
        \fill[pieblue] (0,0) -- (90:\r)
          arc[start angle=90, end angle=\stopangle, radius=\r] -- cycle;%
        \fill[piered] (0,0) -- (\stopangle:\r)
          arc[start angle=\stopangle, end angle=-270, radius=\r] -- cycle;%
      \fi
    \fi
  }%
}

% Two-line header helper
\providecommand{\hdr}[2]{\shortstack{\textbf{#1}\\\textbf{#2}}}

% ----- Modality icons: image / video --------------------------------
% Bumped up via TikZ scale (1x -> 1.8x); baseline tied to the picture's
% own bounding-box centre so the icon sits vertically centred in its cell.
\providecommand{\imgicon}{%
  \tikz[baseline={([yshift=-0.7ex]current bounding box.center)}, scale=1.8]{%
    \draw[piegrey,line width=0.5pt,rounded corners=0.5pt] (0,0) rectangle (1.6ex,1.2ex);%
    \fill[piegrey] (0.4ex,0.85ex) circle (0.16ex);%
    \fill[piegrey] (0.15ex,0.12ex) -- (0.7ex,0.7ex) -- (1.05ex,0.4ex) -- (1.45ex,0.85ex) -- (1.45ex,0.12ex) -- cycle;%
  }%
}
\providecommand{\vidicon}{%
  \tikz[baseline={([yshift=-0.7ex]current bounding box.center)}, scale=1.8]{%
    \draw[piegrey,line width=0.5pt,rounded corners=0.5pt] (0,0) rectangle (1.6ex,1.2ex);%
    \fill[piegrey] (0.55ex,0.3ex) -- (0.55ex,0.9ex) -- (1.15ex,0.6ex) -- cycle;%
  }%
}

\vspace{-0.5em}
\begin{table}[ht]
\centering
\setlength{\tabcolsep}{4pt}
\renewcommand{\arraystretch}{1.15}
\resizebox{\columnwidth}{!}{
\begin{tabular}{l c c c c}
\toprule
\textbf{Bench} & \hdr{Reasoning}{Demand} & \hdr{Appearance}{\textcolor{piered}{Abst.}, \textcolor{pieblue}{Real.}} & \hdr{Process-sensitive}{\textcolor{piered}{No}, \textcolor{pieblue}{Yes}} & \hdr{Difficulty}{Control} \\
\midrule
% VBench           & Low         & \pie{4}{4}  & \pie{4}{4} & \textcolor{piered}{\ding{55}} \\
PhysGenBench     & \textcolor{piered}{Low} / \textcolor{pieyellow}{Mid}     & \pie{4}{4}  & \pie{4}{4} & \pmark \\
WorldSimBench  & \textcolor{piered}{Low} / \textcolor{pieyellow}{Mid}     & \pie{4}{4}  & \pie{4}{4} & \cmark \\
% VideoThinkBench & High        & \pie{0}{4}  & \pie{0}{4} & \textcolor{piered}{\ding{55}} \\
TiVi-Bench$^{\dag}$ & \textcolor{pieblue}{High}     & \pie{6}{24}  & \pie{15}{24} & \pmark \\
V-ReasonBench            & \textcolor{pieblue}{High}    & \pie{38}{328}  & \pie{108}{328}  & \xmark \\
VBVR-Bench       & \textcolor{pieblue}{High}        & \pie{0}{4}  & \pie{47}{100}          & \xmark \\
\midrule
\textbf{Ours}    & \textcolor{pieblue}{High}        & \pie{2}{2}  & \pie{2}{2} & \cmark \\
\bottomrule
\end{tabular}%
}

\begin{tablenotes}[flushleft]
\scriptsize
\item \parbox{0.48\textwidth}{
$^{\dag}$Data has not been publicly released and statistics are inferred from the paper.\par
}
\end{tablenotes}
\vspace{-0.3em}

\caption{Comparison with related video benchmarks. Each pie encodes the fraction of tasks that \textcolor{pieblue}{satisfy a desideratum} versus \textcolor{piered}{do not}. 
For input appearance, \textcolor{pieblue}{Real.} denotes photorealistic-style inputs, while \textcolor{piered}{Abs.} denotes synthetic inputs, like line-art or schematic images. Appendix~\ref{apdx:bench_comparisons} details the ratio assignment. 
}
\label{tab:related_works}
\end{table}
\vspace{-0.5em}

Answering these questions requires benchmarks that go beyond visual fidelity and test whether video models can use their learned visual priors for reasoning. Recent efforts have begun to evaluate video models as zero-shot visual reasoners~\citep{chen2025tivibench, liu2025can, vbvr2026, luo2025vreasonbenchunifiedreasoningbenchmark}, but still leave several important gaps:
% \raisebox{0.2ex}{\fbox{\scriptsize \textbf{1}}}
\boxednum{1} \textbf{Distribution-mismatched visual appearances}. 
Many benchmarks use line-art or abstract inputs for scalability and controllability, but such inputs can deviate far from the natural-image priors of video generation models. In our controlled comparisons, abstract inputs more often lead to collapse and constraint ignorance than the visually realistic counterparts. Such failures may reflect visual domain mismatch more than reasoning limitations, weakening validity of existing evaluations.
\boxednum{2} \textbf{Limited demand for visual rollout reasoning}.
Many existing visual reasoning or visual QA tasks can be answered directly from the input, without requiring the model to simulate how the scene evolves. Therefore, they do not adequately evaluate whether a video generation model can solve a task by explicitly rolling out its visual progression and grounding the final answer in the generated trajectory.
\boxednum{3} \textbf{Uncontrolled task difficulty and feasibility}. 
Existing benchmarks include tasks that are far beyond their feasible regime, making failures less diagnostic. Examples include long-horizon tasks exceeding practical video duration and knowledge-heavy tasks relying on non-visual domain expertise like medical knowledge. A more diagnostic evaluation should calibrate task feasibility near the current capability boundary and organize tasks into graded difficulty levels. The limitations above are summarized in Table~\ref{tab:related_works} and detailed in Appendix~\ref{apdx:bench_comparisons}.

To bridge these gaps, we introduce \bench, a benchmark for evaluating visual intelligence in video generation models through meticulously designed downstream tasks. 
Our benchmark addresses previous issues by \boxednum{1} using photorealistic-style inputs to reduce visual-domain mismatch, \boxednum{2} filtering tasks whose success depends on valid intermediate trajectories rather than final states alone, and \boxednum{3} calibrating difficulty through pre-generation filtering and human review to keep tasks challenging yet partially feasible for current models, as detailed in Section~\ref{3_benchmark}. 
\bench further adopts a two-level taxonomy covering both task domains and skill requirements. Each task is assigned to one mutually exclusive domain based on its visual characteristics, and then annotated with one or more skill tags, as in Figure~\ref{fig:teaser}.

Leveraging \bench, we systematically evaluated a wide range of representative video models to understand their reasoning capacity. The results show that  
%A broad evaluation of 
the current generative models exhibit both emerging reasoning abilities and substantial gaps toward general-purpose visual intelligence. 
Even the strongest model, Seedance~2.0, achieves only 51.0 under our criteria. Current models can make partial progress on visual goals, but often fail to maintain coherent multi-step execution, with common failure modes including physical collapse, rule violation, and object/state inconsistency. 
Beyond performance, our diagnostic analyses examine video reasoning failure from several complementary angles. At the inference level, input conditions like prompts and visual styles substantially affect performance, with open-source models especially sensitive to the visual style gaps. At the training level, large-scale synthetic fine-tuning can transfer from abstract data to realistic tasks, but the gains are bounded by how well the training distribution covers the skill requirements. At the internal level, denoising trajectories show that current video models tend to refine early visual hypotheses, while reliable self-correction of erroneous states remains limited. Together, these analyses provide a more detailed view of reasoning behavior and help identify the factors that shape, limit, and potentially improve it in video generation models.

In summary, our contributions are threefold:
\boxednum{1} \bench: a benchmark for evaluating visual intelligence in video generation models around their current capability boundary. 
\boxednum{2} Broad evaluation of contemporary generative models covering both image and video generation, showing their emerging ability while exposing limitations on general-purpose intelligence. 
\boxednum{3} Multi-faceted analyses of video model reasoning, covering failure modes, input sensitivity, performance transfer of synthetic fine-tuning , and denoising dynamics, yielding insights into the factors that shape and limit visual reasoning performance.

% Our key findings are summarized as follows:

% \boxednum{1} Current video generation models exhibit emerging reasoning ability, but still far from general-purpose visual intelligence. 
% \boxednum{2} Input conditions like prompt formulation and visual style can substantially affect performance, particularly for open-source models. 
% \boxednum{3} Large-scale synthetic fine-tuning provides useful supervision, but its transfer to realistic tasks is bounded by the training data's coverage of target structures and skills.

\section{Related Works}
\label{2_related_works}

\vspace{-0.2em}
\subsection{Video Generation Models}
\vspace{-0.2em}

Video generation models were first developed and evaluated primarily as content creation systems, with progress measured by visual quality, motion realism, and condition alignment~\citep{yang2025cogvideox,  kong2024hunyuanvideo, wan2025wan}. As temporal coherence and physical plausibility improve, they are increasingly viewed as visual world simulators: generated videos can serve as explicit predictions of how scenes, objects, and interactions evolve over time~\citep{brooks2024sora,qin2024worldsimbench}. While more recent studies further suggest another role beyond the above: they may act as zero-shot visual reasoners, expressing solutions through generated frame sequences~\citep{wiedemer2025video,tong2025thinking}, showing a promising paradigm for multi-modal reasoning. 

% \vspace{-0.2em}
% \subsection{Generative Visual Priors}
% \vspace{-0.2em}

% A growing line of work shows generative models encode reusable visual priors beyond synthesis. Image diffusion representations have been found to capture semantics, spatial layout, and dense correspondence, supporting downstream tasks like semantic matching, segmentation, and depth estimation~\citep{tang2023emergent, he2025lotus, ke2025marigold, gabeur2026image}. Recent studies further suggest that video diffusion models encode temporal correspondence and dynamic scene structure across frames~\citep{velez2025image, nam2026emergent}. These findings motivate a natural question: whether such generative priors can be used not only for perception or reconstruction, but also for general reasoning intelligence in video generation.

\vspace{-0.2em}
\subsection{Probing Reasoning in Video Generation}
\vspace{-0.2em}

Studies suggest video generation models can exhibit non-trivial zero-shot reasoning through generated frames~\citep{wiedemer2025video,tong2025thinking}, showing the potential of visual reasoners beyond simulators. This motivated a series of follow-up evaluations, including TiVi-Bench~\citep{chen2025tivibench}, V-ReasonBench~\citep{luo2025vreasonbenchunifiedreasoningbenchmark}, and MMGR~\citep{cai2025mmgrmultimodalgenerativereasoning}, which evaluate generative reasoning across spatial, physical, logical, and other tasks. VBVR~\citep{vbvr2026} further scales this direction from evaluation to adaptation, pairing a large task collection with benchmark-specific supervision for LoRA fine-tuning. Complementary to these efforts, \bench evaluates the reasoning capability encoded in video generation models through naturalistic and goal-directed procedural tasks with feasibility controls.

\section{\bench}
\label{3_benchmark}

% domain-name chips: backgrounds match the teaser figure (lightened class colors)
\definecolor{domVO}{HTML}{D7E3CA}   % visual_organization   — dusty sage
\definecolor{domPM}{HTML}{E5CDCB}   % physical_manipulation — dusty rose
\definecolor{domSP}{HTML}{C5D4E6}   % structured_puzzles    — soft denim blue
\definecolor{domSD}{HTML}{EDDEC6}   % spatiotemporal_dynamics — warm sand
\newcommand{\dombox}[2]{{\setlength{\fboxsep}{1.5pt}\colorbox{#1}{\textbf{#2}}}}

% inline skill-tag glyph; matches the per-skill icons in the teaser
\newcommand{\skillicon}[1]{\raisebox{-0.25ex}{\includegraphics[height=1.05em]{Figures/skill_icons/skill_#1.png}}}

\vspace{-0.2em}
\subsection{Taxonomy}
\label{3_taxonomy}
\vspace{-0.2em}

To cover different aspects of visual intelligence, we 
%adopt
propose a two-level taxonomy: \textbf{Domains} and \textbf{Skill-tags}. The first level organizes tasks into four mutually exclusive domains from their visual characteristics:  \dombox{domVO}{Visual Organization} tasks require models to arrange, group, or select objects based on visual attributes and cues.
 \dombox{domPM}{Physical Manipulation} tasks involve object-level actions such as moving, placing, stacking, or using tools, where success depends on plausible physical interaction.  \dombox{domSP}{Structured Puzzles} focus on rule-governed visual puzzles, where models must follow explicit constraints to transform an initial state into the valid target. \dombox{domSD}{Spatiotemporal Dynamics} tasks require reasoning about how states evolve over time, including ordering and temporal dependency.

Beyond domain-level organization, we further annotate each task with one or more \textbf{Skill Tags} that capture the underlying capabilities for solving it. These tags are inspired by visual cognition theories and provide a capability-level view complementary to the domain taxonomy. Skill tags are non-exclusive and seven tags are used: \skillicon{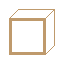} \textbf{Spatial}, \skillicon{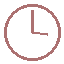} \textbf{Temporal}, \skillicon{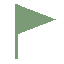} \textbf{Planning}, \skillicon{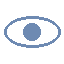} \textbf{Attribute Grounding}, \skillicon{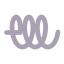} \textbf{Physics}, \skillicon{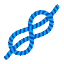} \textbf{Topology}, and \skillicon{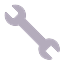} \textbf{Affordance}. This two-level design enables both coarse- and fine-grained diagnosis of model capabilities.

\vspace{-0.1em}
\subsection{Task Collection}
\label{3_task_collection}
\vspace{-0.1em}

\textbf{Task Proposal.\quad}
Our tasks are designed to evaluate visually grounded reasoning processes that can be naturally expressed through video. Each task follows a unified I/O format: the model receives a text prompt and an input image as the first frame, then generates a video completing the specified visual procedure. We focus on \textbf{reasoning-intensive} objectives and avoid low-level recognition or localization tasks. Each task is required to be \textbf{process-sensitive}, where success depends on intermediate state evolution and rule-preserving trajectory, rather than final-state correctness alone. We also constrain the expected solution length to match the typical generation duration of current video models.

Each task is instantiated at three difficulty levels, with roughly ten instances per level. Every instance consists of an input image, a text prompt, and task-specific evaluation criteria~\ref{3_eval_criteria}. The resulting suite is representative rather than exhaustive: we prioritize tasks near the capability boundary of current models, yielding sharper diagnostic signals for current systems while remaining meaningful and challenging for future models.

\noindent
\textbf{Input Image and Prompt.\quad}
Each instance contains an input image and a text prompt. Input images are collected from web images or existing datasets, or generated with image generation models such as GPT-Image-2~\citep{openai_gpt_image_2} and Nano Banana Pro~\citep{google2025nanobananapro}. For generated inputs, we use a human-in-the-loop process: since image generation models may sometimes introduce unintended artifacts, images are manually reviewed and iterated until matching the intended task design. All images are standardized to a 16:9 aspect ratio. The text prompt specifies the task goal and the constraints for the generated videos. It describes the relevant objects, attributes, allowed actions, prohibited shortcuts, and task-specific rules that must be preserved during generation. We also includes task-agnostic controls on background, layout, camera motion, and video speed, so that the generated video remains focused on the intended procedural reasoning rather than irrelevant visual variation.

\noindent
\textbf{Reference Solution.\quad}
Each task is paired with a reference solution specifying the intended outcome. The reference may take the form of an image or a textual description. Image references illustrate an acceptable solution, such as highlighting a valid path in \textsc{maze} task. For tasks whose solutions are better specified semantically, we provide a description for the desired final state, such as task \textsc{untie knot}. The references define the target outcome and are used to guide both task proposal and the criteria construction in Section~\ref{3_eval_criteria}.

\vspace{-0.2em}
\subsection{Quality Control}
\vspace{-0.1em}
\label{3_quality_control}

\textbf{Pre-Generation.\quad}
To calibrate task difficulty, we add a pre-generation stage. For each proposed task, we sample two easiest-level instances and test them on several sota video generation models, such as Sora2, Veo3.1, Kling3.0, etc. A task is accepted only if the sampled instance is solved by at least one model and failed by at least one model; otherwise, we revise its design and input materials. This procedure filters out tasks that are trivially solvable or entirely infeasible for current models, making them \textit{challenging yet partly feasible}. We emphasize this stage serves as a sanity filter: it checks whether a task can plausibly be expressed as a video process, rather than certifying full solvability.

\noindent
\textbf{Manual Review.\quad}
We manually review each task instance: whether it \ding{172} remains faithful to the intended goal, \ding{173} fits within the typical video duration limit like 5-10s, \ding{174} is described by a clear prompt. Unqualified instances are revised or discarded.

% \paragraph{Pre-Generation.}
% To calibrate task difficulty, we introduce a pre-generation stage. For each proposed task, we sample two easiest-level instances and test them with several state-of-the-art video generation models like Sora2, Veo3.1, etc. A task is accepted only if each sampled instance is solved by at least one model and failed by at least one model; otherwise, we revise its design, difficulty, prompt, or input image. This procedure places tasks near the capability boundary of current video models, making them \textbf{challenging yet (partly) feasible}. We stress that this stage only verifies that a task \emph{can, or shows promise to, be expressed as a video} by video generation models; it is not a full solvability check, but a sanity filter that rules out tasks no current model can even begin to render as a coherent visual process.

% \paragraph{Manual Review.}
% Each task is then manually reviewed against four checks: (i) whether it stays faithful to the intended task goal rather than drifting from it; (ii) whether its solution fits within the typical duration limit of video generation (e.g.\ within $10$\,s); (iii) whether the input prompt describes the task clearly and unambiguously; and (iv) whether the evaluation criteria cover the key factors that determine success or failure on the task. Instances failing any check are revised or discarded.

\begin{table*}[ht]
\centering
\setlength{\tabcolsep}{8pt}
\renewcommand{\arraystretch}{1.1}
\resizebox{0.95\textwidth}{!}{%
% \newcolumntype{G}{>{\columncolor{gray!8}[2pt][2pt]}c}  % alternating grey shading (disabled; restore this line to re-enable)
\newcolumntype{G}{c}
% Highlight: pill-shaped \colorbox painted ONLY around the number, not the
% whole cell. Tight \fboxsep so the pill hugs the digits.
\newcommand{\hl}[2]{{\setlength{\fboxsep}{2pt}\colorbox{#1}{#2}}}
\begin{tabular}{ll c G c G c G c G c}
\toprule
\multirow{2}{*}{\textbf{Category}} & \multirow{2}{*}{\textbf{Level}}
 & \multicolumn{6}{c}{\textit{Commercial}}
 & \multicolumn{3}{c}{\textit{Open Source}} \\
\cmidrule(lr){3-8} \cmidrule(lr){9-11}
 &
 & \textbf{Sdce2.0} & \textbf{Sora2} & \textbf{Veo3.1} & \textbf{Kling3.0} & \textbf{Wan2.7} & \textbf{Gen4.5}
 & \textbf{Mnx-H3} & \textbf{HY1.5} & \textbf{Wan2.2} \\
\arrayrulecolor{gray!50}\midrule\midrule\arrayrulecolor{black}
\multirow{4}{*}{\makecell{Visual Org-\\anization}} & Easy & \hl{hlbestb}{72.6} & 55.6 & 50.3 & \hl{hlsecb}{67.6} & 37.0 & 60.5 & 42.8 & 26.6 & 30.4 \\
 & Mid & \hl{hlbestb}{56.4} & 37.0 & 43.3 & \hl{hlsecb}{47.5} & 32.2 & 38.3 & 42.3 & 17.9 & 15.3 \\
 & Hard & \hl{hlbestb}{53.5} & 39.1 & \hl{hlsecb}{44.0} & 43.7 & 32.5 & 41.9 & 39.8 & 24.7 & 18.5 \\
\arrayrulecolor{gray!50}\cline{3-8}\cline{9-11}\arrayrulecolor{black}
 & Avg. & \hl{hlbestb}{60.8} & 43.9 & 45.9 & \hl{hlsecb}{52.9} & 33.9 & 46.9 & 41.6 & 23.1 & 21.4 \\
\arrayrulecolor{gray!50}\midrule\arrayrulecolor{black}
\multirow{4}{*}{\makecell{Spatiotemporal\\Dynamics}} & Easy & \hl{hlbestb}{54.6} & \hl{hlsecb}{45.3} & 28.1 & 45.2 & 38.9 & 39.6 & 44.3 & 34.7 & 38.7 \\
 & Mid & \hl{hlbestb}{47.4} & 34.6 & 21.2 & 35.2 & \hl{hlsecb}{42.6} & 34.6 & 40.7 & 27.2 & 32.0 \\
 & Hard & \hl{hlsecb}{33.4} & 24.4 & 16.9 & 28.2 & 28.9 & 29.5 & \hl{hlbestb}{35.9} & 23.9 & 19.9 \\
\arrayrulecolor{gray!50}\cline{3-8}\cline{9-11}\arrayrulecolor{black}
 & Avg. & \hl{hlbestb}{45.3} & 34.8 & 22.3 & 36.5 & 36.8 & 34.8 & \hl{hlsecb}{40.3} & 28.7 & 30.2 \\
\arrayrulecolor{gray!50}\midrule\arrayrulecolor{black}
\multirow{4}{*}{\makecell{Structured\\Puzzles}} & Easy & \hl{hlsecb}{46.9} & 40.5 & 31.8 & 45.3 & 25.6 & 29.2 & \hl{hlbestb}{51.9} & 9.9 & 17.1 \\
 & Mid & \hl{hlsecb}{45.1} & 25.8 & 21.8 & 38.3 & 23.3 & 20.7 & \hl{hlbestb}{52.3} & 8.8 & 8.4 \\
 & Hard & \hl{hlsecb}{41.8} & 22.4 & 13.5 & 28.9 & 24.7 & 18.7 & \hl{hlbestb}{47.6} & 7.9 & 5.6 \\
\arrayrulecolor{gray!50}\cline{3-8}\cline{9-11}\arrayrulecolor{black}
 & Avg. & \hl{hlsecb}{44.6} & 29.5 & 22.4 & 37.5 & 24.5 & 22.9 & \hl{hlbestb}{50.6} & 8.9 & 10.4 \\
\arrayrulecolor{gray!50}\midrule\arrayrulecolor{black}
\multirow{4}{*}{\makecell{Physical\\Manipulation}} & Easy & \hl{hlbestb}{64.4} & 47.9 & 49.6 & \hl{hlsecb}{60.6} & 55.8 & 52.1 & 58.6 & 22.5 & 33.7 \\
 & Mid & \hl{hlbestb}{59.2} & 41.8 & 40.4 & \hl{hlsecb}{51.2} & 48.2 & 43.4 & 41.5 & 16.7 & 22.1 \\
 & Hard & \hl{hlsecb}{44.4} & 32.6 & 37.3 & \hl{hlbestb}{45.6} & 37.1 & 39.5 & 33.0 & 11.9 & 17.5 \\
\arrayrulecolor{gray!50}\cline{3-8}\cline{9-11}\arrayrulecolor{black}
 & Avg. & \hl{hlbestb}{56.0} & 40.8 & 42.5 & \hl{hlsecb}{52.5} & 47.1 & 45.0 & 44.4 & 17.0 & 24.4 \\
\arrayrulecolor{gray!50}\midrule\midrule\arrayrulecolor{black}
\multicolumn{2}{c}{\textbf{Overall}} & \hl{hlbestb}{51.0} & 36.7 & 32.0 & 44.0 & 35.7 & 36.6 & \hl{hlsecb}{44.4} & 19.1 & 21.6 \\
\bottomrule
\end{tabular}%
}
\vspace{-0.3em}
\caption{Evaluation results of video generation models, the \colorbox{hlbestb}{\strut best} and \colorbox{hlsecb}{\strut second-best} scores are highlighted. HY1.5 for \texttt{HunyuanVideo-1.5}, Sdce2.0 for \texttt{Seedance2.0}, Mnx-H3 for \texttt{MiniMax-H3}.}
\label{tab:main_res}
\end{table*}

\vspace{-0.2em}
\subsection{Evaluation Criteria}
\label{3_eval_criteria}
\vspace{-0.2em}

Since our tasks are process-sensitive, evaluation must assess both \textbf{goal completion} and \textbf{process validity}. A correct-looking final state is insufficient if the trajectory violates task rules, while a locally plausible video may still fail by making little progress toward the goal. We therefore 
%adopt
propose two complementary metrics.

\noindent
\textbf{Completeness (Comp.)\quad}
This metric captures the global progress toward the task goal. Since different tasks have different goal states, we define a task-specific tiered standard and map the video to one of three levels: \texttt{<complete>}, \texttt{<partial>}, or \texttt{<failed>}. The VLM-judge receives a set of uniformly sampled frames (2fps) together with the standard and returns the corresponding tier. 

\noindent
\textbf{Rubric Score (Rub.)\quad}
This metric measures local process validity throughout the video. We design a fine-grained checklist for each task that covers explicit rules and other constraints. Since some critical violations may occur briefly, we use a coarse-to-fine adaptive frame sampling strategy. The VLM-judge starts from coarse sampling (4fps), inspecting the video against the checklist and flagging intervals that require closer inspection (the per-window calls only localize evidence; the final violation counts are assigned by a later aggregation step). These intervals are then resampled at a finer rate (8fps) and re-evaluated, allowing the judge to capture transient violations without densely sampling the entire video. To keep each judgment focused, we further adopt a sliding focus window (10-frame) with edge frame overlapping, so that only a small local segment is inspected at a time rather than all the sampled frames. 

Each rubric item is scored by an inverse decay penalty, $1/(x+1)$, where $x$ is the number of violations of this item. The inverse decay reflects the intuition that once a violation occurs repeatedly, additional occurrence should have diminishing marginal effect. The Rubric Score is the average over all item-level scores. Other monotonic decay functions like exponential decay are also applicable and preserve the same qualitative trend.

\noindent
\textbf{Aggregation.\quad} 
The \textbf{Final Score} combines global completion and local process validity by multiplying Comp. with Rub.. This multiplicative design penalizes either type of failure. An almost static video preserves most local constraints (high Rub.) but make little progress towards task goal(low Comp.); Conversely, a video may appear to reach the target state (high Comp.) while violating the rules (low Rub.). The aggregation therefore treats both as jointly necessary conditions, preventing either aspect alone from dominating the evaluation.

Appendix~\ref{apdx:eval_criteria} shows examples of criteria for Comp. and Rub., while 
Appendix~\ref{apdx:vlm_as_judge} provides implementation details of the VLM-based evaluator.

\vspace{-0.2em}
\subsection{Data Augmentation}
\vspace{-0.2em}

\label{3_augmentation}
\noindent
\textbf{Image Output Adaptation.\quad}
In order to extend \bench as a testbed for reasoning in image generative models, we repurpose some tasks into single-image output format while preserving task goal. This adaptation does not contradict the  process-sensitive task design: the original video tasks evaluate whether a video model can express the procedure through temporal state evolution, while the image version asks whether an image model can infer and render the target state or visual solution. For example, \textsc{maze} is adapted into drawing a valid path from start to the goal, and \textsc{recover 2d net} is adapted into rendering the completed 3D structure. This branch enables comparison of static and procedural reasoning under related task goals. More details are in Appendix~\ref{apdx:bench_aug}.

\vspace{-0.2em}
\section{Experiments}
\label{4_experiments}

\vspace{-0.2em}
\subsection{Setup}
\label{4_setup}
\vspace{-0.1em}

Our evaluation covers a broad suite of closed-source and open-source video models, together with image models on the adapted subset from Section~\ref{3_augmentation}. 
Due to the evaluation cost, we evaluate half of the instances for each task using a fixed random seed. Appendix~\ref{apdx:stable_eval_subset} further analyzes stability of the half-instances protocol compared to full-set evaluation.
Following the criteria in Section~\ref{3_eval_criteria}, We use Gemini-3-Flash~\citep{google2025gemini3flash} for our automatic evaluator. The full model list and generation configurations are in Appendix~\ref{apdx:model_gen_config}. 

\vspace{-0.2em}
\subsection{Results}
\label{4_main_results}
\vspace{-0.2em}

\begin{figure}[ht]
  \centering
  \includegraphics[width=0.9\linewidth]{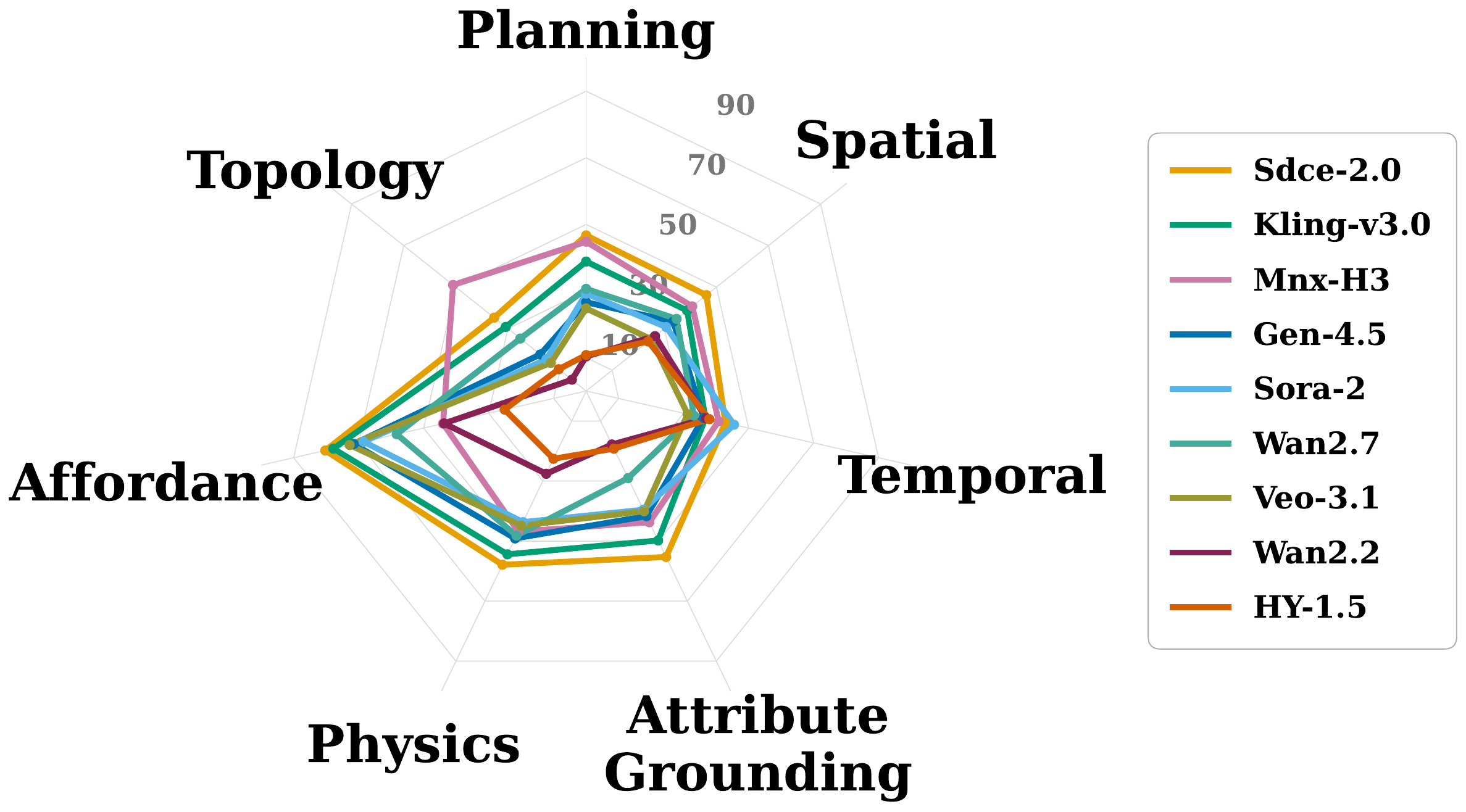}
  \caption{Per-model performance across the skill tags.}
  \vspace{-0.5em}
  \label{fig:radar_skill}
\end{figure}

\paragraph{Commercial video models lead, but all remain far from solved.}
Table~\ref{tab:main_res} reports the video generation model results. Commercial models consistently outperform open-source models, with \texttt{Seedance-2.0} achieving the best overall score (51.0); nevertheless, all 
%models
tasks remain far from solved. \textbf{Structured Puzzles} is the most challenging domain, exposing failures in multi-step rules and state tracking. Figure~\ref{fig:radar_skill} further shows that \textbf{Topology} and \textbf{Temporal} are the weakest skill dimensions (one task may have several skill tags), reflecting failures in connectivity preservation and multi-step state tracking.  
To complement the main metrics, we report a strict success rate (S.R.), where an instance is counted as successful only if both completeness and rubric scores are perfect. Besides, we also conduct a human study for estimating the benchmark ceiling to demonstrate the gap between generative models and average humans. Detailed results are in Appendix~\ref{apdx:full_eval_res} and \ref{apdx:human_ceiling}.

\paragraph{Image generation serves as a static goal-state diagnostic.}
We evaluate image models on an output-adapted subset, measuring how well the image models can realize the static goal-state when temporal process validity is removed. Table~\ref{tab:main_img_gen_res} shows a similar commercial--open-source gap, led by \texttt{Nano-Banana-Pro} (Avg. 55.0) and \texttt{Seedream-5.0-Pro} (Avg. 52.1). The consistent drop from easy to hard suggests meaningful difficulty gradients of our tasks. However, this setting only tests target-state inference and rendering; it does not evaluate valid intermediate process, which remains the focus of our video benchmark.

\vspace{-0.2em}
\subsection{Evaluation Reliability}
\label{4_eval_reliability}
\vspace{-0.1em}

The reliability of our VLM-based evaluator is assessed by comparing it against human annotations and ablating the key components: adaptive frame sampling and the sliding focus window. Also, we report the ablation results of base model with comparable pricing level. Table~\ref{tab:judge_pairwise_auc} reports AUC and pairwise accuracy against human preferences for the full method and its two ablations. The full evaluator achieves the strongest agreement with human judgments, while removing either component leads to degradation, confirming the necessity of both. Details of correlation analysis are in Appendix~\ref{apdx:judge_human_corr}.

\begin{table}[ht]
\centering
\setlength{\tabcolsep}{8pt}
\renewcommand{\arraystretch}{1.0}
\resizebox{0.9\linewidth}{!}{%
\begin{tabular}{l cc}
\toprule
\textbf{Judge Design} & \textbf{AUC} & \textbf{Pairwise Acc.} \\
\midrule
Main (w/\; \texttt{Gemini-3-Flash})        & \textbf{0.803} & \textbf{73.2\%} \\
\midrule
\;\;w/o adaptive fps  & 0.772          & 69.5\%          \\
\;\;w/o focus window  & 0.753          & 68.3\%          \\
\midrule
\;\;w/\; \texttt{GPT-5-mini}  & 0.690          & 64.3\%          \\
\;\;w/\; \texttt{Claude-Haiku-4.5}  & 0.478          & 47.9\%          \\
\;\;w/\; \texttt{Qwen3.6-Plus}  & 0.624          & 59.1\%          \\
% \;\;w/\; \texttt{Qwen3-VL-235B-A22B}  & 0.634          & 59.0\%          \\

\bottomrule
\end{tabular}
}
\vspace{-0.4em}
\caption{
Reliability of our VLM-based evaluator compared with human annotations, including key component and base model ablations.
% \textbf{AUC} for the ROC-AUC over the strict-preference pairs (label $\in\{+1,-1\}$), using the raw judge-score difference as the ranking scalar.
% \textbf{Pairwise Acc.} is judge-vs-human ternary agreement (\textgreater, \textless, $=$).
}
\label{tab:judge_pairwise_auc}
\end{table}

% low-saturation sage greens for best/2nd-best (defined outside \resizebox
% so the caption's \colorbox sample can also see them — \definecolor is
% local to the current group, and \resizebox introduces one).
\definecolor{hlbest}{HTML}{C5D9B5}
\definecolor{hlsec}{HTML}{E3EDDA}

\begin{table*}[ht]
\centering
\setlength{\tabcolsep}{8pt}
\renewcommand{\arraystretch}{1.1}
\resizebox{\textwidth}{!}{%
% \newcolumntype{G}{>{\columncolor{gray!8}[2pt][2pt]}c}  % alternating grey shading (disabled; restore this line to re-enable)
\newcolumntype{G}{c}
% Highlight: pill-shaped \colorbox painted ONLY around the number, not the
% whole cell. Tight \fboxsep so the pill hugs the digits.
\newcommand{\hl}[2]{{\setlength{\fboxsep}{2pt}\colorbox{#1}{#2}}}
\begin{tabular}{l c G c G c G c G c G c}
\toprule
 & \multicolumn{6}{c}{\textit{Commercial}}
 & \multicolumn{5}{c}{\textit{Open Source}} \\
\cmidrule(lr){2-7} \cmidrule(lr){8-12}
 & \textbf{\makecell{Nano-\\Banana-Pro}}
 & \textbf{\makecell{Qwen-\\Image-3-Pro}}
 & \textbf{\makecell{GPT-\\Image-2}}
 & \textbf{\makecell{MAI-Image\\-2.5-Pro}}
 & \textbf{\makecell{Seedream\\4.5}}
 & \textbf{\makecell{Flux.2\\Max Edit}}
 & \textbf{\makecell{Sense-\\Nova-U1}}
 & \textbf{\makecell{JoyAI-\\Image}}
 & \textbf{\makecell{Qwen-\\Image-Edit}}
 & \textbf{\makecell{BAGEL}}
 & \textbf{\makecell{Step1X-\\Edit}} \\
\arrayrulecolor{gray!50}\midrule\midrule\arrayrulecolor{black}
Easy    & \hl{hlbest}{62.5} & \hl{hlsec}{51.9} & 48.8 & 44.9 & 27.5 & 23.8 & 22.2 & 11.2 & 11.2 & 1.2 & 1.2 \\
Mid     & \hl{hlbest}{50.0} & \hl{hlsec}{43.8} & 38.8 & 35.1 & 21.2 & 20.3 & 14.4 & 10.0 & 5.0 & 1.2 & 0.0 \\
Hard    & \hl{hlbest}{52.5} & 32.5 & \hl{hlsec}{36.2} & 30.3 & 22.5 & 17.7 & 12.2 & 10.0 & 6.2 & 0.0 & 1.2 \\
\arrayrulecolor{gray!50}\midrule\arrayrulecolor{black}
Avg. & \hl{hlbest}{55.0} & \hl{hlsec}{42.7} & 41.2 & 36.8 & 23.8 & 20.6 & 16.3 & 10.4 & 7.5 & 0.8 & 0.8 \\
\bottomrule
\end{tabular}%
}
\vspace{-0.4em}
\caption{Evaluation results of image models on the adapted subset. Success Rate (\%) is measured and reported (different from the video branch), against the ground truth image, the \colorbox{hlbest}{\strut best} and \colorbox{hlsec}{\strut second-best} scores are highlighted.}
\vspace{-0.5em}
\label{tab:main_img_gen_res}
\end{table*}

\vspace{-0.2em}
\section{Discussion}
\label{5_discussion}

\vspace{-0.2em}
% --- shared takeaway-box style: rounded rectangle, black frame, grey fill ---
\newtcolorbox{takeaway}[1][]{%
  enhanced,
  colback=black!6,
  colframe=black,
  boxrule=0.6pt,
  arc=6pt,
  left=10pt, right=10pt, top=7pt, bottom=7pt,
  #1
}

Beyond aggregated scores, we further diagnose video reasoning across four stages: output failures, input conditions, training-time transfer, and internal denoising dynamics. This progression examines not only where the model fails, but also how their behavior changes across multiple aspects.

\vspace{-0.1em}
\subsection{Failure Modes}
\label{5_failure_modes}
\vspace{-0.1em}

\begin{figure}[ht]
  \centering
  \begin{subfigure}[t]{\linewidth}
    \includegraphics[width=\linewidth]{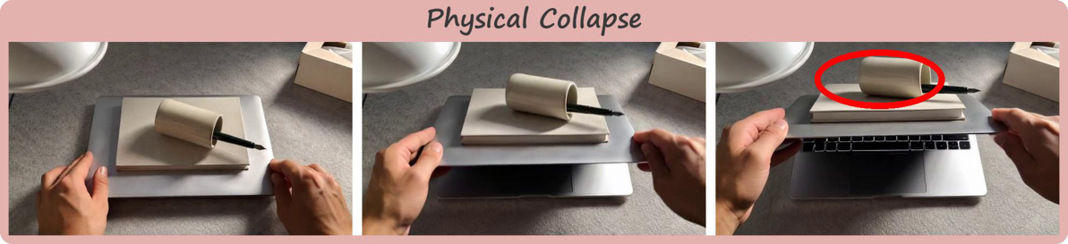}
    \caption{\textbf{Physical collapse.} The laptop is tilted up, yet \textbf{the cup lying on its lid does not roll down}.}
  \end{subfigure}
  \par\medskip
  \begin{subfigure}[t]{\linewidth}
    \includegraphics[width=\linewidth]{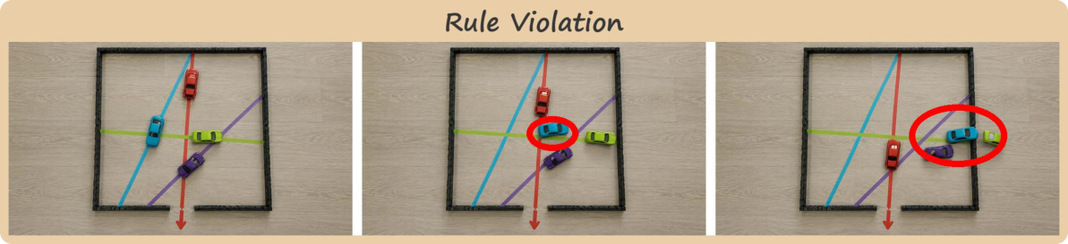}
    \caption{\textbf{Rule violation.} Each car must stay on its own colored track and may not cross walls; \textbf{the video violates both rules}.}
  \end{subfigure}
  \par\medskip
  \begin{subfigure}[t]{\linewidth}
    \includegraphics[width=\linewidth]{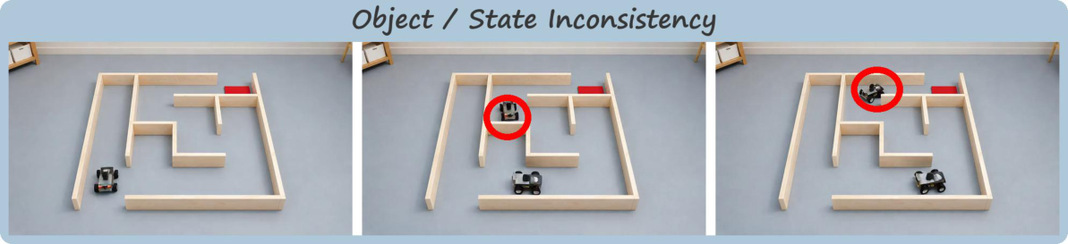}
    \caption{\textbf{Object/state inconsistency.} While the car drives through the maze, \textbf{a spurious second car appears partway through}.}
  \end{subfigure}
  \caption{Representative cases in the failure modes.}
  \label{fig:failure_modes_main}
\end{figure}

We firstly examine representative failure cases and summarize several recurring failure modes, as shown in Fig ~\ref{fig:failure_modes_main}.
\boxednum{1} \textbf{Physical collapse.}
Generated videos sometimes contain unrealistic deformation, object penetration, or sudden disappearance. Under strong task-goal or constraint pressure, physical causality becomes fragile and may be sacrificed for a goal-like visual outcome, suggesting current models may encode local visual dynamics, but still struggle to maintain physically coherent processes under goal-directed generation.
\boxednum{2} \textbf{Rule violation.}
Models may generate videos that remain physically plausible and coherent, yet violate the explicit rules. For example, they may perform prohibited actions or skip essential intermediate steps. In some cases, the model reaches a visually plausible final state by directly altering the target scene, rather than following the expected procedure.
\boxednum{3} \textbf{Object/state inconsistency.}
Models often fail to preserve object identities, positions, or intermediate states over time. Objects may disappear, transform, or reset to earlier states, even when the overall motion appears smooth. This reveals weak temporal state tracking, which is critical for goal-directed process generation.
See more examples in Appendix~\ref{apdx:failure_modes}.

\vspace{-0.2em}
\subsection{Input Condition Sensitivity}
\label{5_sensitivity}
\vspace{-0.1em}

We probe input robustness of video reasoning along two axes: \textbf{text prompt} with varying levels of detail and \textbf{input image} in different visual styles. For both aspects, we evaluate a fixed subset of tasks. 

% Prompt optimization is widely used to improve quality and controllability of visual generation. We test its potential ceiling for video reasoning by building an \emph{oracle prompt} describing the intended solution as explicitly as possible. This reduces the reasoning burden and probes whether the model can follow a goal-directed solution description and render a valid visual process. Figure~\ref{fig:sens_combined} shows oracle prompting generally improves performance, with the effect being more pronounced for closed-source models. Details on task and model selection are in Appendix~\ref{apdx:sens_analysis}. 

\begin{figure}[ht]
    \centering
    \includegraphics[width=\linewidth]{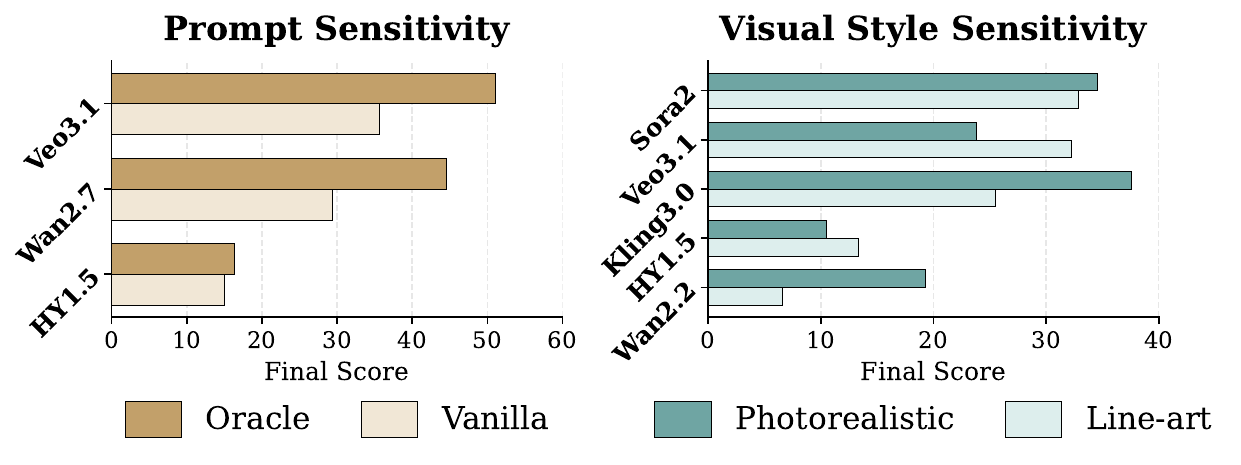}
    \caption{Performance comparison in oracle prompting and in varying input visual styles.}
    \label{fig:sens_combined}
    \vspace{-0.4em}
\end{figure}

\noindent
\textbf{Oracle Prompting.\quad}
Prompt optimization is widely used to improve the quality and controllability of visual generation. We test its potential ceiling for video reasoning by building an \emph{oracle prompt} describing the intended solution as explicitly as possible. This reduces the high-level reasoning burden and probes whether the model can follow a goal-directed solution description and render the corresponding visual process. As shown in Figure~\ref{fig:sens_combined}, oracle prompting improves performance for some models, especially closed-source ones, but the gains remain limited. Even with the full solution provided, video models often fail to render the complete solution trajectory. This reflects two bottlenecks: some solutions are intrinsically difficult to specify precisely in language, especially when involving fine-grained spatiotemporal transitions; and current video models still struggle with instruction following and physical simulation under strong rule constraints. The performance gain is particularly weak for \texttt{HunyuanVideo 1.5}, whose lower base reasoning capability leaves little room for oracle prompts to help. Details on task and model selection are provided in Appendix~\ref{apdx:sens_analysis}.

\noindent
\textbf{Visual Style. \quad}
Using the metadata of selected tasks, we generate line-art variants that preserve the original task structure, and compare performance against the realistic-style setting. As shown in Figure~\ref{fig:sens_combined}, model rankings under line-art inputs differ noticeably from realistic-style. 
This discrepancy is especially pronounced for open-source models, indicating stronger sensitivity to input visual appearance. These results also support one of our motivations: visual style can substantially affect the measured performance, so evaluations dominated by abstract inputs may conflate reasoning limitations with visual-domain mismatch. See more details in Appendix~\ref{apdx:sens_analysis}.

\begin{figure}[ht]
    \centering
    \includegraphics[width=\linewidth]{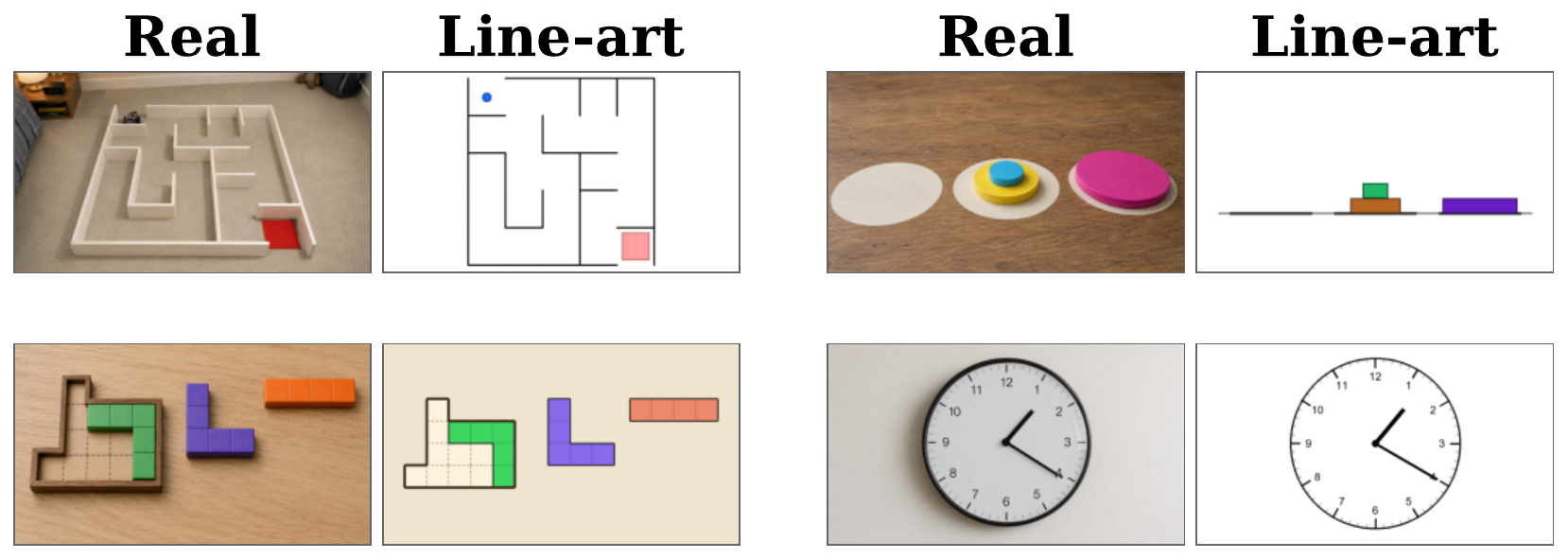}
    \caption{Style variants (Real / Line-art) from 4 tasks: {maze}, {hanoi tower}, {polyform tiling}, {clock running}.}
    \label{fig:style_eg}
\end{figure}

\vspace{-0.2em}
\subsection{How Well does Scaling Tuning Transfer?}
\label{5_sft_transfer}
\vspace{-0.1em}
Given that pretrained video models already encode useful visual priors, an interesting question is how much downstream fine-tuning can further improve the reasoning capability ~\citep{vbvr2026, zhu2026videomodelsreasonverifiable, chen2026opencoflearningreasonvideo}. This question becomes more practical when considering data scalability: curating large-scale real videos with explicit reasoning goals is expensive, and synthetic data offers a more controllable and scalable alternative, but it remains unclear whether such scaling can systematically improve video reasoning in realistic scenarios. We therefore study it through a synthetic-data scaling setting, by comparing the released VBVR models~\citep{vbvr2026}, fine-tuned on 1M-sample abstract-style dataset, with its respective base models \texttt{Wan2.2-I2V-A14B}, \texttt{Wan2.1-I2V-14B}, and \texttt{LTX-2.3}.

\begin{table}[ht]
\centering
\setlength{\tabcolsep}{8pt}
\renewcommand{\arraystretch}{1.15}
\resizebox{\columnwidth}{!}{%
\begin{tabular}{l c c c}
\toprule
\textbf{Model}
 & \textbf{Overlap}
 & \textbf{Semi-overlap}
 & \textbf{Non-overlap} \\
\midrule
\texttt{Wan2.2-I2V} (base) & 15.3 & 17.8 & 29.2 \\
\texttt{VBVR-Wan2.2}       & \cellcolor{hlgreen!40}\textbf{55.8}\,{\footnotesize\textcolor{Green}{$\uparrow$40.5}}
                           & \cellcolor{hlgreen!17}\textbf{34.9}\,{\footnotesize\textcolor{Green}{$\uparrow$17.1}}
                           & \cellcolor{hlgreen!6}\textbf{35.4}\,{\footnotesize\textcolor{Green}{$\uparrow$\phantom{0}6.2}} \\
\midrule
\texttt{Wan2.1-I2V} (base) & 11.3 & 17.0 & 25.6 \\
\texttt{VBVR-Wan2.1}       & \cellcolor{hlgreen!14}\textbf{25.1}\,{\footnotesize\textcolor{Green}{$\uparrow$13.8}}
                           & \cellcolor{hlgreen!3}\textbf{20.2}\,{\footnotesize\textcolor{Green}{$\uparrow$\phantom{0}3.2}}
                           & \cellcolor{hlred!4}22.0\,{\footnotesize\textcolor{Red}{$\downarrow$\phantom{0}3.6}} \\
\midrule
\texttt{LTX-2.3} (base)    & 15.0 & 18.2 & 15.5 \\
\texttt{VBVR-LTX2.3}       & \cellcolor{hlgreen!9}\textbf{24.1}\,{\footnotesize\textcolor{Green}{$\uparrow$\phantom{0}9.1}}
                           & \cellcolor{hlgreen!5}\textbf{23.6}\,{\footnotesize\textcolor{Green}{$\uparrow$\phantom{0}5.4}}
                           & \cellcolor{hlgreen!4}\textbf{19.2}\,{\footnotesize\textcolor{Green}{$\uparrow$\phantom{0}3.7}} \\
\bottomrule
\end{tabular}%
}
\vspace{-0.4em}
\caption{Performance of the three released VBVR models and their respective base models, grouped by structural overlap with training distribution. Per-cell values are mean final score across tasks in the group.}
\label{tab:vbvr_sft_res}
\end{table}

Table~\ref{tab:vbvr_sft_res} groups our tasks by the structural overlap with the VBVR training data distribution, with details and examples in Appendix~\ref{apdx:vbvr_sft}. At the group level, performance gains decrease with structural overlap for all three base models, indicating that synthetic fine-tuning transfers more effectively across aligned task structures. However, this trend is not uniform at the task level: overlap does not guarantee reasoning improvement, while some non-overlap tasks still benefit. For example, the task \textsc{untie\_knot} shows little change in overall success rate, yet its rubric score rises substantially from 0.02 to 0.34. We attribute this to more controlled behavior in the video and fewer rule violations after the fine-tuning, together with limited transfer of basic spatial or logical capabilities. More results and details are provided in Appendix~\ref{apdx:vbvr_sft}.

Table~\ref{tab:vbvr_sft_skill} further shows the domain- and skill-level breakdown and the uneven gains. Improvements are concentrated on skills well represented in the synthetic training set, such as planning and spatial reasoning, while capabilities like physical interaction or strong temporal dependency remain hard to improve and may even degrade. 

% 3-lv canonical view (build_release stripe), TASK_LEVEL_EXCLUDE/TASK_EXCLUDE honored.
% Combined table: per-domain (top) + per-skill (bottom, via \input), one shared caption.
\begin{table}[ht]
\centering
\setlength{\tabcolsep}{4pt}
\renewcommand{\arraystretch}{1.15}
\resizebox{\columnwidth}{!}{%
\begin{tabular}{l c c c c c}
\toprule
\textbf{Model}
 & \makecell{Structured\\Puzzles}
 & \makecell{Visual\\Organization}
 & \makecell{Spatiotemporal\\Dynamics}
 & \makecell{Physical\\Manipulation}
 & \textbf{Overall} \\
\midrule
\texttt{Wan2.2-I2V}  & 11.6 & 21.4 & 30.2 & 24.4 & 21.5 \\
\texttt{VBVR-Wan2.2} & \cellcolor{hlgreen!42}53.2\,{\footnotesize\textcolor{Green}{$\uparrow$41.6}}
                     & \cellcolor{hlgreen!16}37.6\,{\footnotesize\textcolor{Green}{$\uparrow$16.2}}
                     & \cellcolor{hlred!5}29.0\,{\footnotesize\textcolor{Red}{$\downarrow$1.2}}
                     & \cellcolor{hlgreen!18}42.6\,{\footnotesize\textcolor{Green}{$\uparrow$18.1}}
                     & \cellcolor{hlgreen!20}41.2\,{\footnotesize\textcolor{Green}{$\uparrow$19.7}} \\
\bottomrule
\end{tabular}%
}
\\[0.9em]
% Per-skill tabular fragment; included by Tables/5_vbvr_sft_per_domain.tex
% (the combined per-domain + per-skill table), which owns the table env,
% caption, and labels.
% inline skill-icon helper: glyph + bold name
\providecommand{\skicon}[2]{%
  \raisebox{-0.2em}{\includegraphics[height=1.1em]{Figures/skill_icons/skill_#1.png}}\,\textbf{#2}%
}
\resizebox{\columnwidth}{!}{%
\begin{tabular}{l c c c c}
\toprule
\textbf{Model}
 & \skicon{planning}{Planning}
 & \skicon{spatial}{Spatial}
 & \skicon{temporal}{Temporal}
 & \skicon{attribute_grounding}{\makecell{Attribute\\Grounding}} \\
\midrule
\texttt{Wan2.2-I2V}  & 10.4 & 26.4 & 37.0 & 17.8 \\
\texttt{VBVR-Wan2.2} & \cellcolor{hlgreen!34}44.1\,{\footnotesize\textcolor{Green}{$\uparrow$33.7}}
                     & \cellcolor{hlgreen!11}37.3\,{\footnotesize\textcolor{Green}{$\uparrow$10.9}}
                     & \cellcolor{hlred!11}25.8\,{\footnotesize\textcolor{Red}{$\downarrow$11.2}}
                     & \cellcolor{hlgreen!20}37.9\,{\footnotesize\textcolor{Green}{$\uparrow$20.1}} \\
\midrule
\textbf{Model}
 & \skicon{spring}{Physics}
 & \skicon{affordance}{Affordance}
 & \skicon{topology}{Topology}
 & \\
\midrule
\texttt{Wan2.2-I2V}  & 27.6 & 43.7 &  5.4 & \\
\texttt{VBVR-Wan2.2} & \cellcolor{hlgreen!12}39.9\,{\footnotesize\textcolor{Green}{$\uparrow$12.3}}
                     & \cellcolor{hlgreen!12}55.2\,{\footnotesize\textcolor{Green}{$\uparrow$11.5}}
                     & \cellcolor{hlgreen!21}26.3\,{\footnotesize\textcolor{Green}{$\uparrow$20.9}}
                     & \\
\bottomrule
\end{tabular}%
}

\vspace{-0.4em}
\caption{Per-domain (top) and per-skill (bottom) performance (final score) breakdown for \texttt{VBVR-Wan2.2} vs.\ base \texttt{Wan2.2-I2V}.}
\label{tab:vbvr_sft_skill}
\label{tab:vbvr_sft_domain}
\end{table}

These results demonstrate the promise of synthetic data as a scalable source of supervision, while also revealing that its transfer is bounded by the structural coverage of the training distribution, leaving more effective scaling strategies and the transfer limits of video reasoning for future exploration.

\vspace{-0.2em}
\subsection{When Is Visual Reasoning Decided?}
\label{5_reasoning_denoising}

In the section, we go beyond the performance and examine reasoning behavior along the denoising trajectory. Recent work on diffusion large language models (dLLMs)~\citep{ye2024diffusion, zhao2026d1, nie2026large} connects iterative denoising with reasoning behaviors such as ``self-consistency" and ``self-correction", where the latter refers to revising incorrect intermediate states in the later denoising steps. This perspective has also been extended to video generative reasoning~\citep{wang2026demystifing}, suggesting that the reasoning may unfold along denoising steps and that self-correction may emerge during the generation.

% Distribution of solution-state transitions between decoded denoising checkpoints.
% Source: stats_fig_tab/5_4_denoising_intermediates/label_dist.txt
\begin{table}[ht]
\centering
\setlength{\tabcolsep}{3.5pt}
\renewcommand{\arraystretch}{1.12}
\resizebox{\columnwidth}{!}{%
\begin{tabular}{l cccccc}
\toprule
\textbf{Transition}
 & \textbf{1$\to$2} & \textbf{2$\to$3} & \textbf{3$\to$4}
 & \textbf{4$\to$10} & \textbf{10$\to$20} & \textbf{20$\to$40} \\
\midrule
\catOne\;\;Unrecognizable        & 69.2 & 41.9 & 22.2 & \phantom{0}6.8 & \phantom{0}0.0 & \phantom{0}0.0 \\
\catTwo\;\;Stable                & 17.9 & 49.6 & 70.1 & 69.2 & 75.2 & 90.6 \\
\midrule
\catThree \  (a)\;\;Correct\,$\to$\,Wrong   & \phantom{0}2.6 & \phantom{0}0.9 & \phantom{0}0.0 & \phantom{0}0.9 & \phantom{0}0.0 & \phantom{0}0.0 \\
\rowcolor{hlgreen!12}
\catThree \  (b)\;\;Wrong\,$\to$\,Correct   & \phantom{0}0.9 & \phantom{0}0.0 & \phantom{0}0.9 & \phantom{0}0.0 & \phantom{0}0.0 & \phantom{0}0.0 \\
\catThree \ (c)\;\;Wrong\,$\to$\,Wrong$'$  & \phantom{0}9.4 & \phantom{0}7.7 & \phantom{0}6.8 & 23.1 & 24.8 & \phantom{0}9.4 \\
\bottomrule
\end{tabular}%
}
\vspace{-0.4em}
\caption{Distribution (\%) of solution-state transitions between consecutive decoded denoising steps. Row \catThree \  (b) (shaded) is self-correction. Protocols are in Appendix~\ref{apdx:denoising}.}
\label{tab:denoise_transition}
\end{table}

\begin{figure}[ht]
  \centering
  \begin{tcolorbox}[enhanced, 
  width=0.92\linewidth,
  colback=white, colframe=orange!75!black, boxrule=0.8pt, arc=4pt, left=4pt, right=4pt, top=4pt, bottom=4pt]
    \includegraphics[width=\linewidth]{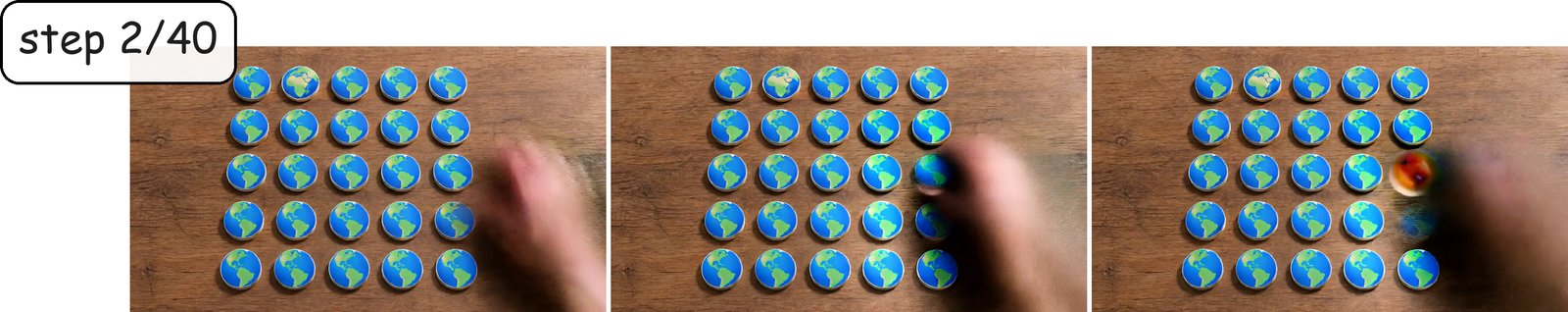}
    \par\vspace{-2pt}
    \includegraphics[width=\linewidth]{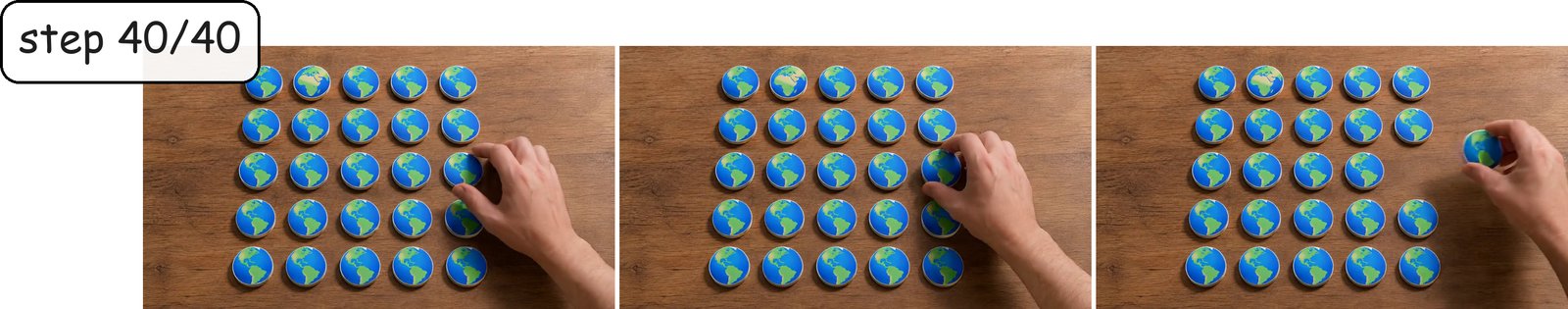}
  \end{tcolorbox}
  \vspace{-6pt}

  \begin{tcolorbox}[enhanced, 
  width=0.92\linewidth, colback=white, colframe=blue!55!black, boxrule=0.8pt, arc=4pt, left=4pt, right=4pt, top=4pt, bottom=4pt]
    \includegraphics[width=\linewidth]{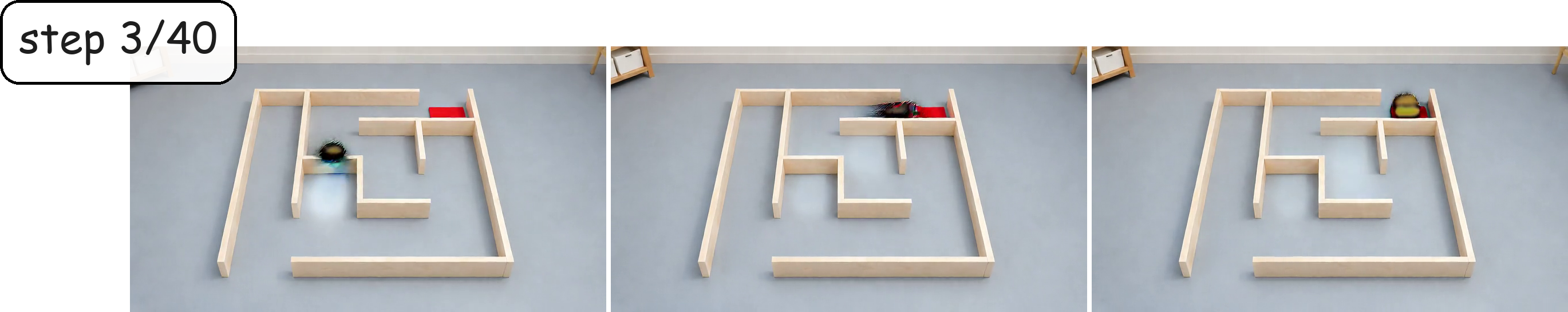}
    \par\vspace{-2pt}
    \includegraphics[width=\linewidth]{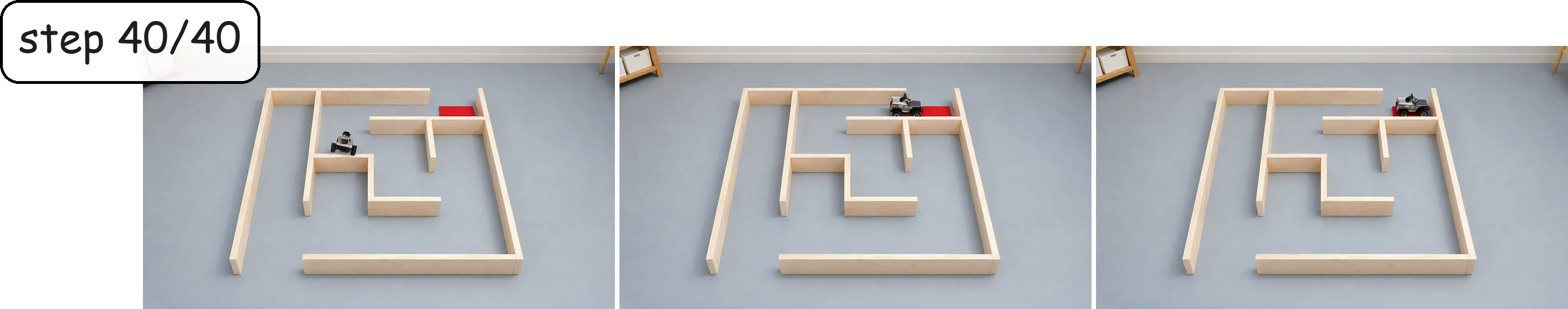}
  \end{tcolorbox}
  \caption{Decoded frames at different denoising stages.}
  \label{fig:reasoning_denoising}
\end{figure}

Our empirical analysis offers a more cautious view. We decode intermediate denoising states from four open-source models and label how the visible solution state changes between consecutive decoded checkpoints: \catOne~unrecognizable, \catTwo~stable, or \catThree~changed, with changes split into correct$\to$wrong, wrong$\to$correct (i.e.\ self-correction), and wrong$\to$wrong$'$ (one incorrect hypothesis replaced by another). More details about the protocol are detailed in Appendix~\ref{apdx:denoising}.

As shown in Table~\ref{tab:denoise_transition}, the solution state does change during denoising, but almost never toward a correct one. Self-correction stays below $1\%$ everywhere and stops occurring in later steps, while wrong$\to$wrong$'$ is an order of magnitude more frequent ($23.1\%$ at $4\to10$ and $24.8\%$ at $10\to20$). Once the state is readable at all, it mostly stays put, with stability rising to $90.6\%$ over the second half of denoising. Revision therefore moves between wrong solutions rather than toward the right one; later steps mostly lock in and refine the early hypothesis~\citep{newman2026video, zhu2026videomodelsreasonverifiable}, even when it already violates task rules, as shown in Figure~\ref{fig:reasoning_denoising}. Both these quantitative and qualitative results provide a more grounded view of reasoning behavior in video generation.

% This may stem from the representation gap between dLLMs and video diffusion models: dLLMs operate over \textbf{discrete and semantically interpretable tokens}, while video models denoise \textbf{high-dimensional continuous spatio-temporal latents} where identity, layout, and motion are deeply entangled. In contrast, sampling-based self-consistency appears more practical, as multiple generations can explore different trajectories and selection can improve performance. These results provide a more grounded view of reasoning behavior in video generation.

\vspace{-0.2em}
\section{Conclusion}
\label{6_conclusion}
\vspace{-0.2em}

We introduce \bench, a benchmark for evaluating visual intelligence in video generation models. With realistic-style inputs, process-sensitive task design, and calibrated difficulty levels, \bench provides a diagnostic testbed for assessing whether video models can solve tasks through valid visual rollouts. Extensive evaluation shows current models exhibit emerging reasoning ability, but still far from general visual intelligence. Further analyses reveal sensitivity to input conditions, bounded transfer from scaling synthetic fine-tuning, and limited self-correction during denoising process. We hope \bench can support more systematic diagnosis and development of future video and multimodal foundation models.

% Bibliography entries for the entire Anthology, followed by custom entries
%\bibliography{anthology,custom}
% Custom bibliography entries only
\section*{Limitations}
\label{limitations}

Our benchmark has several scope-related limitations we want to make explicit. First, all tasks are designed around the typical generation length of current video models (roughly 5--10s); longer-horizon procedural reasoning, such as multi-minute multi-step assembly or long-trajectory planning, is therefore out of scope. Second, the benchmark covers only the image-to-video (i2v) setting with a fixed 16:9 aspect ratio; text-to-video, multi-image conditioning, and audio-conditioned generation are left to future work. Third, all task prompts and rubrics are written in English, so multilingual or cross-lingual evaluation is not addressed. Finally, the task suite is intentionally \emph{representative rather than exhaustive}: it captures a focused slice of visual reasoning domains and difficulty levels, but does not aim to enumerate every conceivable visual reasoning scenario, and may need to be extended as model capabilities improve.

\bibliography{custom}

\clearpage
\appendix

% The class sets \flushbottom (acl.sty), which is right for running text but
% wrong here: the appendix is mostly tall tcolorboxes and figures, so on pages
% that cannot be filled LaTeX stretched whatever glue it could reach and tore
% boxes away from their captions. \raggedbottom lets the leftover space collect
% at the foot of the page instead.
\raggedbottom

% ---- appendix-only table of contents (hyperlinked, down to the A.1 level) ----
% The .toc file is read line by line by \tableofcontents, so a \setcounter
% smuggled into it via \addtocontents takes effect *mid-listing*: everything
% written before the marker (the main-body sections) is suppressed by
% tocdepth=-2, and the appendix entries that follow print at tocdepth=2.
% Nothing else in the document needs to change, and entries stay hyperlinked.
\setcounter{tocdepth}{-2}
\addtocontents{toc}{\protect\setcounter{tocdepth}{2}}
\begingroup
  \renewcommand{\contentsname}{Appendix Contents}
  \setlength{\parskip}{0pt}
  \tableofcontents
\endgroup
\clearpage

\section{Benchmark Construction Details}
\label{apdx:bench_details}

\subsection{Task Material Collection}
\label{apdx:material_collection}
Complementing Section~\ref{3_task_collection}, we further provide more details about the task material collection.

\paragraph{Input Image Construction.}
Our input images come from two sources. The first source consists of web images or existing visual datasets, selected and adapted when they naturally match the task setting. The second source consists of images generated by image generation models such as GPT-Image-2~\citep{openai_gpt_image_2} and Nano Banana Pro~\citep{google2025nanobananapro}. For generated inputs, we use two construction pipelines. Some images are generated directly from textual scene descriptions when the target scene can be clearly specified in language. For tasks requiring precise spatial layouts or structured configurations, we first create an intermediate sketch or schematic using scripts or rendering engines, and then use it as a visual reference for photorealistic-style image generation. In both cases, we aim to produce inputs with natural lighting, plausible object appearance, realistic backgrounds, and stable spatial layouts.

Since generated images are not always faithful to the intended task design, we use a human-in-the-loop refinement process. Each generated image is manually checked for object and attribute correctness, spatial layout, and visual artifacts. Unqualified images are regenerated or edited through multiple iterations, until the input image supports the intended visual procedure. All input images are standardized to a 16:9 aspect ratio for compatibility across video generation models.

% The syn-to-real conversion examples (rendered schematic -> realistic input
% image, generated with GPT-Image-2 / Nano Banana Pro) are commented out below;
% the conversion process is summarized in the ``Input Image'' paragraph above.
% To restore the example boxes, remove the surrounding \iffalse ... \fi.
\iftrue
\phantomsection
\label{lst:apdx_convert_prompt_maze}
\begin{tcolorbox}[
  title=Input Image Construction Prompt: \textsc{maze},
  colback=YellowLight,
  colframe=Yellow,
  breakable, contbreak
]
\linespread{1.18}\scriptsize\selectfont
Convert this line-art schematic of a square grid maze into a realistic indoor photo, viewed top-down at $\sim$75$^\circ$. \\
- \textbf{Walls}: where the input shows thin black line segments, render low white wooden-plank walls (painted or light natural wood, a single material); keep each wall segment's position and length, but make the walls clearly low so the corridors stay visible from above. \\
- \textbf{Floor}: one piece of light short-pile carpet (solid or faint pattern, e.g.\ off-white, light grey, or pale khaki) covering the whole maze; its pattern must not blur the wall boundaries. \\
- \textbf{Robot car}: the blue circle in the input marks the car's starting cell. Replace it with a small realistic engineering-style robot car (wheels, visible sensors / camera, metal or plastic shell), placed in the same cell and facing the adjacent open corridor; its length must be under 2/3 of the corridor width and it must not touch any wall. \\
- \textbf{Goal}: the highlighted pink-red filled square is the unique exit. Place a distinct rectangular red mat on that cell (clearly separate from the carpet, vivid colour); no other coloured markers anywhere. \\
- \textbf{Background}: a simple indoor room (faint furniture edges, baseboards), no outdoor elements; edges and corners may faintly show a lived-in bedroom corner (bed sheet, nightstand, lamp), soft and slightly blurred, never a plain white or grey studio backdrop. \\
- Keep the viewpoint and the maze geometry (every cell, every wall segment, the exit gap) strictly unchanged; no artifacts, no distortion.

\vspace{4pt}
\centering
\includegraphics[width=0.44\linewidth]{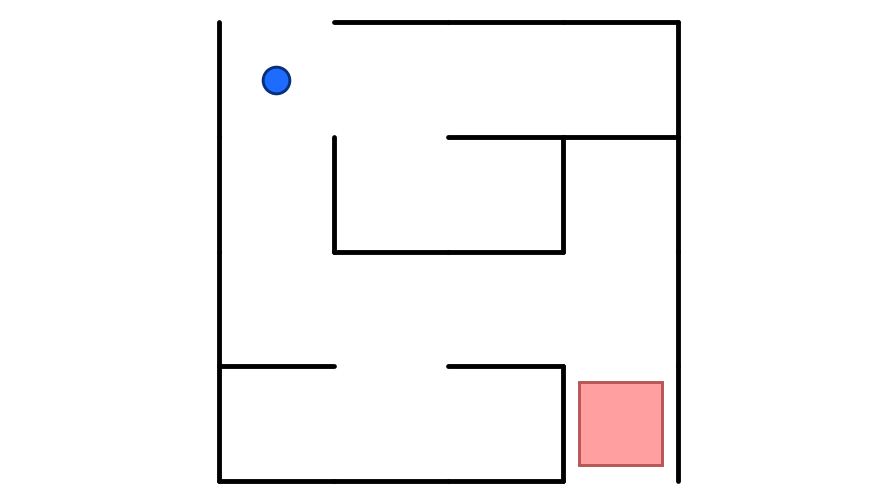}\;$\Rightarrow$\;%
\includegraphics[width=0.44\linewidth]{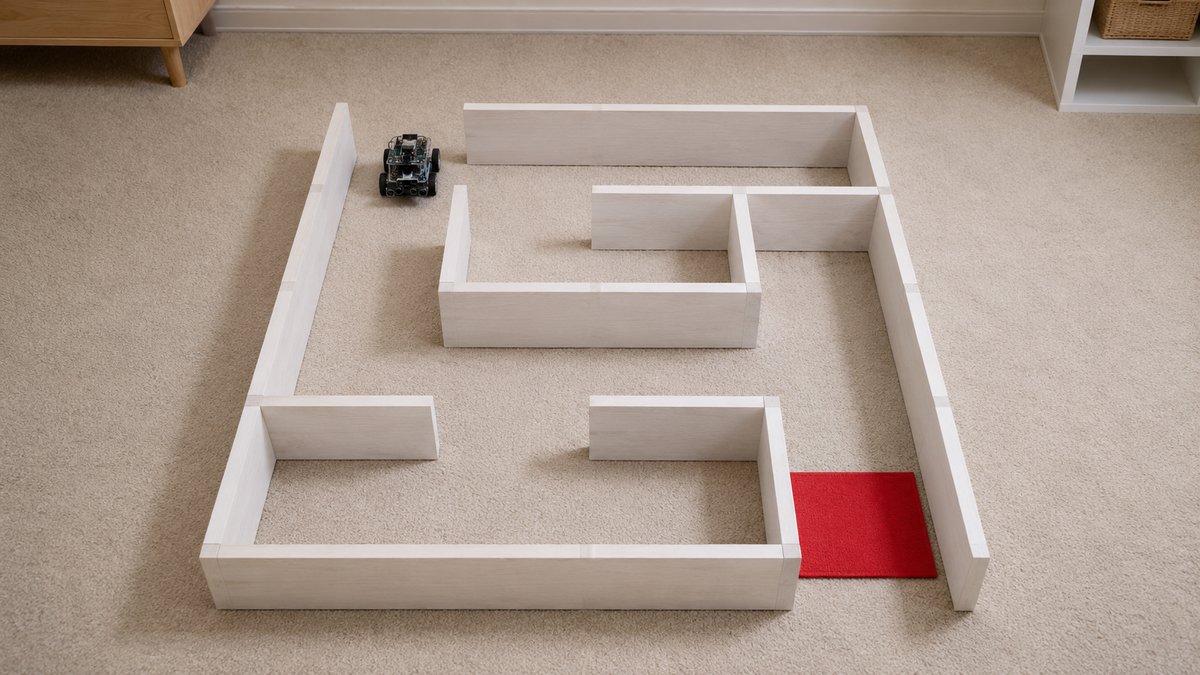}
\end{tcolorbox}

\phantomsection
\label{lst:apdx_convert_prompt_2d3d}
\begin{tcolorbox}[
  title=Input Image Construction Prompt: \textsc{recover\_2d\_net},
  colback=YellowLight,
  colframe=Yellow,
  breakable, contbreak
]
\linespread{1.18}\scriptsize\selectfont
Convert this polygon-net schematic into a realistic $\sim$25$^\circ$--30$^\circ$ top-down photo so each tile's thickness is visible. \\
- \textbf{Tiles}: each polygon is an independent semi-transparent matte Magna-Tiles-style plate ($\sim$3--4\,mm thick); per-tile shape, colour, and count match the input 1-to-1. \\
- \textbf{Rigid pieces}: visible plastic rim, silver magnetic strips along the sides, faint seams between adjacent tiles; never a single printed sheet. \\
- \textbf{Hatched face}: keep the $45^\circ$ diagonal lines as a silk-screened pattern on that tile's top face. \\
- \textbf{Lift / shadow}: tiles sit slightly above the table and cast shape-aligned shadows. \\
- \textbf{Table}: light wood, soft directional lighting; edges may faintly show a desk corner. \\
- No hands, robot arms, clamps or external objects; every tile stays in its input position with magnetic edges snapped.

\vspace{4pt}
\centering
\includegraphics[width=0.44\linewidth]{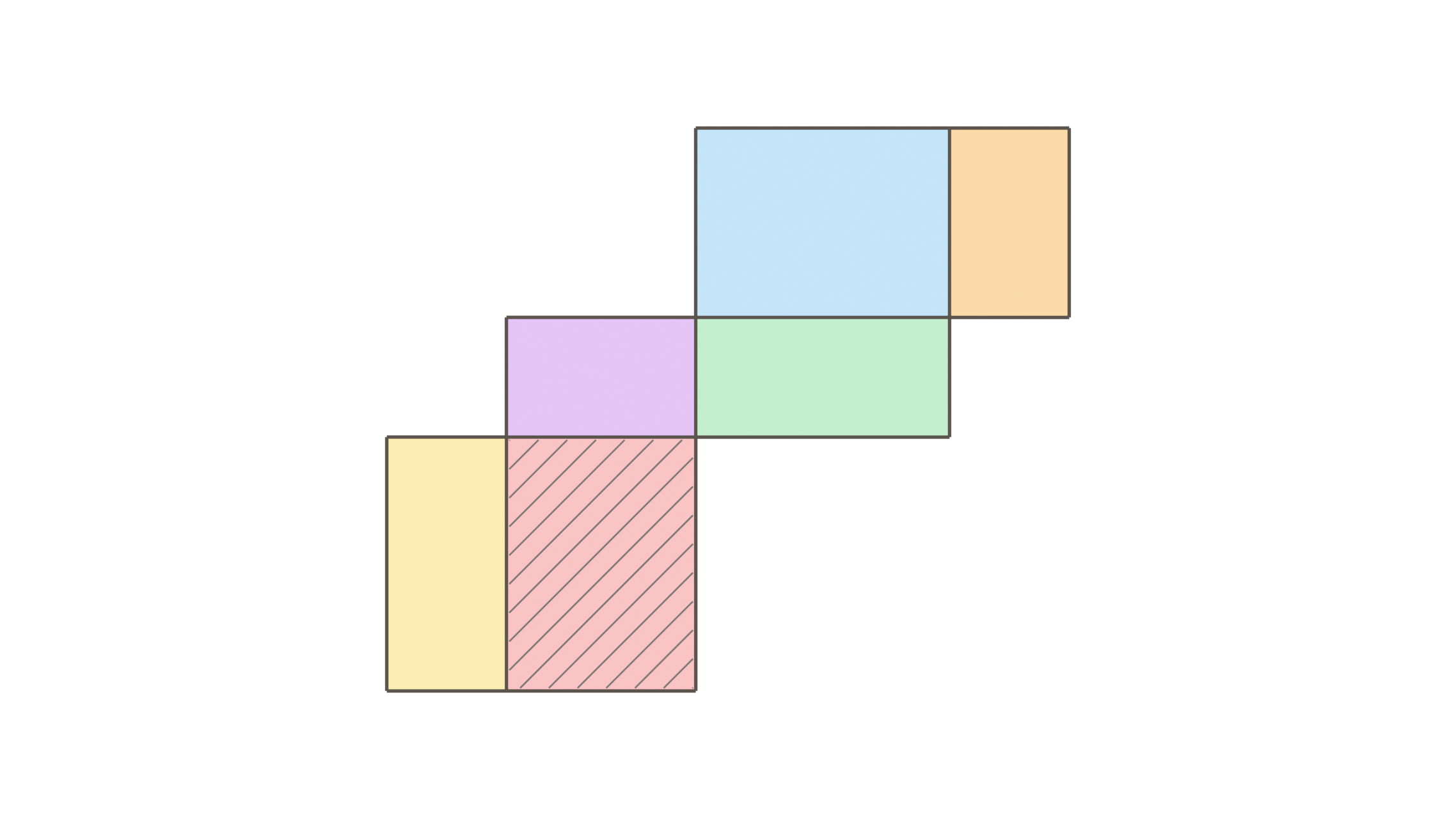}\;$\Rightarrow$\;%
\includegraphics[width=0.44\linewidth]{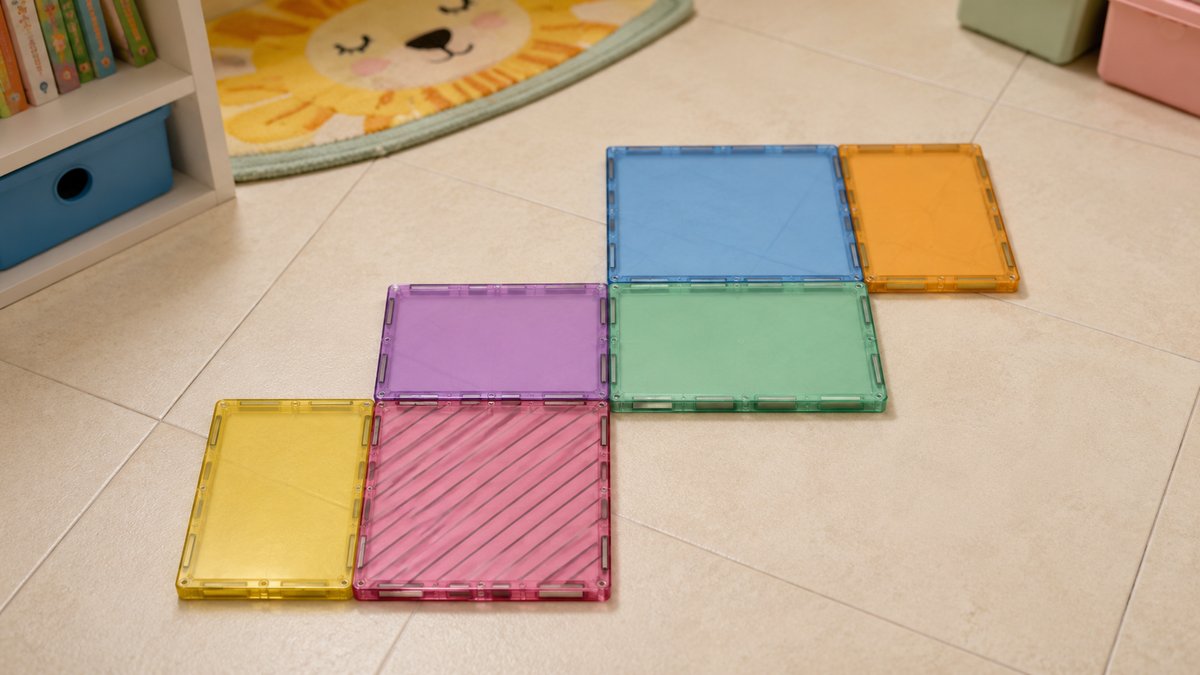}
\end{tcolorbox}

\phantomsection
\label{lst:apdx_convert_prompt_object_packing}
\begin{tcolorbox}[
  title=Input Image Construction Prompt: \textsc{object\_packing},
  colback=YellowLight,
  colframe=Yellow,
  breakable, contbreak
]
\linespread{1.18}\scriptsize\selectfont
A casual phone-style real-scene photo, no intermediate schematic. \\
- \textbf{Container}: in the centre of a light wooden table, place an open beige canvas tote bag (~35\,cm tall) with its top fully open so the empty interior cavity is clearly visible. The interior is plainly roomy: large enough to fit every sensibly-sized item with margin to spare. \\
- \textbf{Surrounding items}: arrange 4 items on the tabletop around the container, evenly spaced, none overlapping or occluding another: a small white folded parasol, an orange-capped sunscreen bottle, a rolled blue-and-white striped beach towel, and a clear glass of iced lemonade. \\
- \textbf{Lighting}: natural cool-white window light from the upper left; every item casts a crisp, real-world drop shadow and shows side highlights, with the contrast and sharpness of a casual snapshot: \textbf{not} a soft studio-lit look. \\
- \textbf{Camera}: ~70$^\circ$ top-down at 16:9 aspect ratio. \\
- No people, hands, text, labels, stickers, logos, arrows, numbers, tools, or extra props anywhere in the frame.

\vspace{4pt}
\centering
\includegraphics[width=0.72\linewidth]{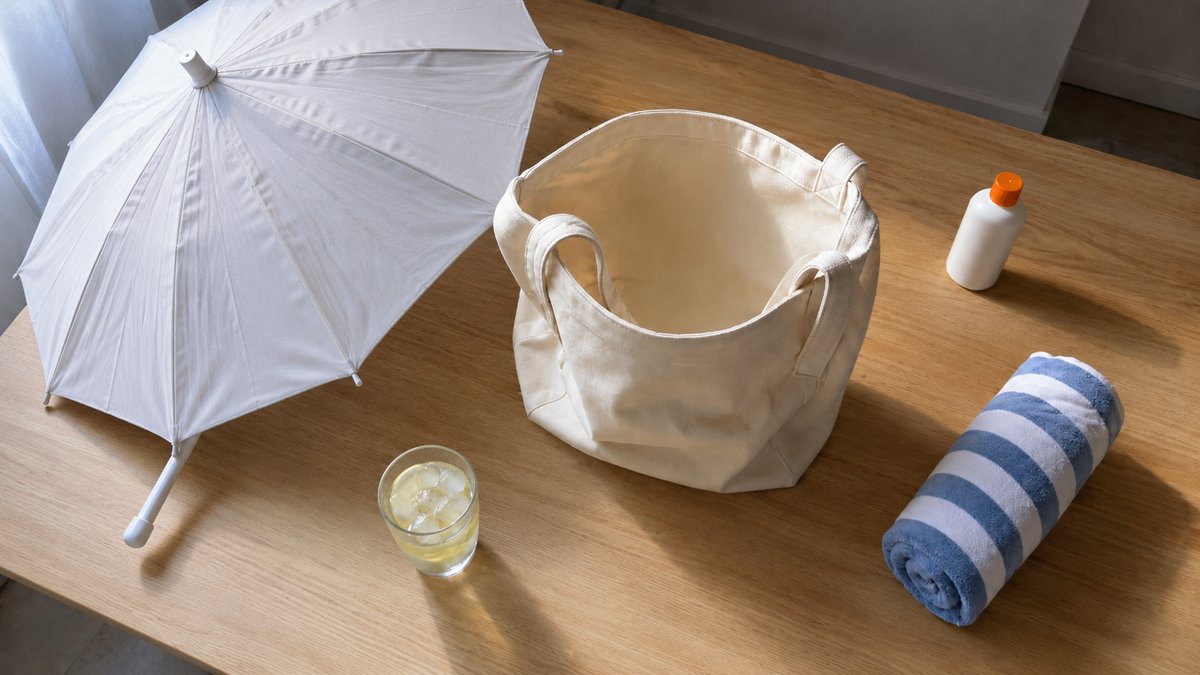}
\end{tcolorbox}
\fi

\paragraph{Text Prompt Construction.}
Each prompt is structured to describe both the task objective and the constraints under which the video should be generated. It first states the goal of the task, then specifies task-specific rules over relevant objects, visual attributes, allowed actions, prohibited actions, and required state changes. Following \cite{wiedemer2025video}, we also include generation-control instructions on the scene background, spatial layout, camera motion, etc. These constraints are intended to reduce irrelevant variation and keep the generated video focused on the intended visual procedure.

Below we show the video prompts used for three representative tasks: \textsc{maze}, \textsc{recover\_2d\_net}, and \textsc{object\_packing}.

\phantomsection
\label{lst:apdx_video_prompt_maze}
\begin{tcolorbox}[
  title=Video Prompt: \textsc{maze},
  colback=GreenLight,
  colframe=Green,
  breakable, contbreak
]
\linespread{1.18}\scriptsize\selectfont
\textbf{Task Goal:}
\\
Create a smooth video of the toy car driving forward through the open corridors of the maze, turning at junctions, and stopping on the red goal square.

\textbf{Constraints:}
\\
- The car must not climb, jump, fly over, or clip through any wall; it stays on the corridor floor at all times. \\
- Use the input layout exactly: no walls, corridors, start cell, or goal are altered. \\
- Motion is continuous along a single valid corridor path from start to goal; no teleport and no sudden cut. \\
- Keep the walls, floor, lighting, and viewing angle fixed throughout; no glitches or artifacts.
\end{tcolorbox}

\phantomsection
\label{lst:apdx_video_prompt_2d3d}
\begin{tcolorbox}[
  title=Video Prompt: \textsc{recover\_2d\_net},
  colback=GreenLight,
  colframe=Green,
  breakable, contbreak
]
\linespread{1.18}\scriptsize\selectfont
\textbf{Task Goal:}
\\
The translucent plastic tiles in the input complete the assembly of a 3D solid entirely on their own: no hands, tools, or external props appear. Each remaining flat face rotates upward about its hinge edge until the net closes into a single closed 3D solid resting on the hatched bottom face.

\textbf{Constraints:}
\\
- \textbf{Polygon set preserved}: same count, shapes, and colours as the input; no face is added, removed, replaced, or recoloured. \\
- \textbf{Rigid faces}: each face keeps its exact polygonal shape (a triangle stays the same triangle, a 2:1 rectangle stays 2:1); no stretching, no edge-length or angle change. \\
- \textbf{Adjacency}: two polygons sharing a creased edge in the input share that same edge in the 3D solid; adjacent faces remain joined along their hinge edge throughout; no two faces overlap in 3D, no face is missing on the surface. \\
- \textbf{Bottom face anchored}: the hatched face stays in contact with the table throughout: it does not lift, rotate, or change shape; faces already in their 3D position at frame 0 stay there. \\
- \textbf{Continuous fold}: faces rotate gradually and continuously, no sudden cut from flat to solid, and stay perfectly rigid throughout. \\
- \textbf{Scene fidelity}: no hands, fingers, arms, sleeves, or other objects enter the frame; background, lighting, and table stay consistent with the input; the camera may tilt from top-down up to a $3/4$ view as the solid forms.
\end{tcolorbox}

\phantomsection
\label{lst:apdx_video_prompt_object_packing}
\begin{tcolorbox}[
  title=Video Prompt: \textsc{object\_packing},
  colback=GreenLight,
  colframe=Green,
  breakable, contbreak
]
\linespread{1.18}\scriptsize\selectfont
\textbf{Task Goal:}
\\
Pack every item that is BOTH size-appropriate AND semantically appropriate for the open container in the centre of the frame into that container; items that do not fit, would spill, contaminate, or damage the container or its contents must stay still on the table. By the end of the video every appropriate item is inside the container and every inappropriate item is still in its original position.

\textbf{Constraints:}
\\
- \textbf{Sole container}: the open container in the centre of the frame is the ONLY container: any box, carton, bag, packaging, plate, or other container-like object among the scattered candidate items is itself a candidate item to be judged, never the container. \\
- \textbf{Capacity is not the constraint}: the container is generously sized; rejection comes only from per-item size mismatch or semantic appropriateness, not from running out of room. \\
- \textbf{Container stays put}: the container itself stays in its starting position and orientation, with its top opening visibly upward so packed items can be seen inside; only its top is open. \\
- \textbf{Continuous physical motion}: one continuous take that ends with the correct items deposited inside the container interior; no jump cuts, teleporting, morphing, fades, or glitches. \\
- \textbf{Scene fidelity}: background, table surface, lighting, and camera viewpoint stay fixed throughout; no labels, text, arrows, debugging overlays, or captions appear.
\end{tcolorbox}

% \subsection{Quality Control}
% \label{apdx:quality_control}
% \paragraph{Pre-Generation} As part of our quality control process, we introduce a pre-generation stage. After a task is proposed, we randomly sample two instances from its easiest difficulty level and test them with three representative video generation models: Sora2, Veo3.1, and Wan2.5. For each sampled instance, at least one model must successfully complete the task, while at least one model must fail. Tasks that do not meet this criterion are sent back for revision, which may involve adjusting the task design or difficulty, clarifying the prompt, or recollecting/generating higher-quality input images.

% This stage helps calibrate each task to lie near the boundary of current video models’ reasoning abilities. To the best of our knowledge, such a pre-generation procedure has rarely been used in existing video reasoning benchmarks. It allows us to construct tasks that are both challenging and feasible, making the benchmark useful for evaluating current models and for testing future, more capable video generation systems.

\subsection{Evaluation Criteria}
\label{apdx:eval_criteria}

\paragraph{Completeness (Comp.)}
Completeness is the \emph{global} metric of our two-metric design. As defined in Section~\ref{3_eval_criteria}, it captures how far the model carried the task goal across the whole clip: a task-specific tiered standard maps each run to one of three levels: \texttt{<complete>}, \texttt{<partial>}, or \texttt{<failed>}. The VLM-based judge, shown a small set of uniformly sampled frames together with that standard, returns the corresponding tier, which is then mapped to $\{0, 0.5, 1\}$. The tiered standard is short by design, so the judge only has to discriminate between the three tiers; the per-task tier definitions used in our experiments are listed in the boxes below.

\phantomsection
\label{lst:apdx_completeness_maze}
\begin{tcolorbox}[
  title=Completeness Standard: \textsc{maze},
  colback=BlueLight,
  colframe=Blue,
  breakable, contbreak
]
\linespread{1.18}\scriptsize\selectfont
\textbf{0}: The car doesn't move, or it immediately clips through / climbs / flies over walls, or it heads the wrong way. \\
\textbf{1}: The car drives correctly through part of the maze (staying in corridors) but does not reach the red goal square. \\
\textbf{2}: The car reaches and stops on the red goal square, staying on the corridor floor throughout (a tiny clip is tolerable if it clearly solves the maze).
\end{tcolorbox}

\phantomsection
\label{lst:apdx_completeness_2d3d}
\begin{tcolorbox}[
  title=Completeness Standard: \textsc{recover 2d net},
  colback=BlueLight,
  colframe=Blue,
  breakable, contbreak
]
\linespread{1.18}\scriptsize\selectfont
\textbf{0}: Stays flat / faces don't fold, or faces are added / removed, or it folds into a wrong shape. \\
\textbf{1}: Some faces rotate up about their hinges but the solid is not closed / only partially assembled. \\
\textbf{2}: The net closes into the complete 3D solid resting on its bottom face. One face not perfectly seated is fine if the solid is essentially closed.
\end{tcolorbox}

\phantomsection
\label{lst:apdx_completeness_object_packing}
\begin{tcolorbox}[
  title=Completeness Standard: \textsc{object packing},
  colback=BlueLight,
  colframe=Blue,
  breakable, contbreak
]
\linespread{1.18}\scriptsize\selectfont
\textbf{0}: No item is put into the container at all (packing not attempted), or the scene deviates drastically from the input. \\
\textbf{1}: Items are actually placed into the container (procedure carried out) but the selection is wrong: appropriate items missed and / or inappropriate items packed. \\
\textbf{2}: Essentially all size- and semantically-appropriate items end up inside the container and the inappropriate ones stay on the table (at most one mistake tolerated).
\end{tcolorbox}

\paragraph{Rubric Score (Rub.)}
Rubric Score is the \emph{local} metric of our two-metric design. As defined in Section~\ref{3_eval_criteria}, it scores a generated video against a fine-grained per-task checklist of process and final-state constraints. The judge runs an adaptive coarse-to-fine pass over the clip (sample at $2$\,fps, refine flagged intervals up to $8$\,fps) with $8$-frame sliding windows so each call stays focused; per rubric item we convert the violation count $x$ across all windows to an item-level score $1/(x+1)$ and average over items. Unlike Completeness, this gives credit for partial process adherence and penalises localised constraint violations even when the global outcome looks correct. The per-task checklists used in our experiments are listed in the boxes below.

\phantomsection
\label{lst:apdx_rubric_maze}
\begin{tcolorbox}[
  title=Evaluation Rubric: \textsc{maze},
  colback=BlueLight,
  colframe=Blue,
  breakable, contbreak
]
\linespread{1.18}\scriptsize\selectfont
- The maze layout: every wall, corridor, the start cell, and the red goal square: is preserved exactly as in the input image; no walls are added, removed, moved, recoloured, or distorted at any point. \\
- Throughout the video the toy car stays entirely inside the corridors; it never clips through, drives over, climbs on top of, or flies above a wall. A violation requires the car to visibly pass THROUGH a wall to reach the other side (a wall-crossing event); merely touching, grazing, or briefly overlapping a wall outline without crossing to the other side does NOT count as a violation. \\
- The car's motion is continuous along a single valid corridor path: no teleportation, no sudden jumps, no extra duplicate cars spawning. \\
- The car visibly turns at junctions to navigate the maze (rather than going straight through walls or taking an impossible shortcut). \\
- By the end of the video the car has come to a full stop fully on the red goal square, having actually travelled there from its starting cell during the clip.
\end{tcolorbox}

\phantomsection
\label{lst:apdx_rubric_2d3d}
\begin{tcolorbox}[
  title=Evaluation Rubric: \textsc{recover 2d net},
  colback=BlueLight,
  colframe=Blue,
  breakable, contbreak
]
\linespread{1.18}\scriptsize\selectfont
- \textbf{Overall task}: starting from the input (a flat 2D net or a partially-folded shape), the remaining flat faces hinge upward about their fold edges on their own until the net closes into a single closed 3D solid resting on the hatched bottom face. The completed solid is exactly: 2 regular triangle bases $+$ 3 rectangles (right regular 3-gonal prism). No other solid or unrelated object besides the target solid appears anywhere in the scene. \\
- \textbf{Bottom face anchored}: the hatched face stays flat on the table throughout and is the only face in contact with the table; every other face has rotated upward so the solid is closed. \\
- \textbf{Face inventory and identity preserved}: same number of faces with their original shapes, sizes, proportions, and pastel colours: no face added, removed, recoloured, stretched, or scaled. \\
- \textbf{Continuous physical fold}: the unfolding-to-3D plays out as a continuous physical fold: faces hinge upward gradually, stay rigid, and stay joined at their fold edges throughout; no sudden teleport from flat to solid.
\end{tcolorbox}

\phantomsection
\label{lst:apdx_rubric_object_packing}
\begin{tcolorbox}[
  title=Evaluation Rubric: \textsc{object packing},
  colback=BlueLight,
  colframe=Blue,
  breakable, contbreak
]
\linespread{1.18}\scriptsize\selectfont
- All starting elements from the input remain present and visually identifiable throughout the output: the beige canvas cloth tote bag (about 30\,cm tall, top open, interior visible) and every candidate item (open white parasol / umbrella; glass of lemonade (full, open); bottle of sunscreen; rolled-up striped beach towel). \\
- By the final frame the container holds exactly the items that SHOULD be packed: bottle of sunscreen, rolled-up striped beach towel. \\
- The items that should NOT be packed remain on the table in their original positions and orientations and are never circled, never moved, never tilted, never picked up: open white parasol / umbrella (reason: an opened parasol is far too large to fit in the bag); glass of lemonade (full, open) (reason: open liquid would spill in a soft bag). \\
- Only the candidate items end up moved; the container itself stays in its starting position and orientation, with its top opening visibly oriented upward so packed items can be seen inside. \\
- Motion is one continuous physical take that ends with the correct items deposited inside the container interior. No jump cuts, teleporting, morphing, fades, or glitches. \\
- Background, table surface, lighting, and camera viewpoint are unchanged from the input throughout the output. No labels, text, arrows, debugging overlays, or captions. \\
- Total count of items inside the container at the final frame equals the number of correct items (2); total count of items still on the table equals the number of wrong items (2).
\end{tcolorbox}

\subsection{Benchmark Augmentation}
\label{apdx:bench_aug}
\paragraph{Image Output Subset}

We adapt a subset (totally 16 tasks) of \bench tasks into single-image output format to support auxiliary evaluation of image generation and unified multimodal models. A task is adapted only when its goal can be meaningfully represented by a single image, such as a final configuration, a selected object, a completed structure, or a visual annotation. Tasks whose correctness inherently depends on continuous temporal interaction, such as hand-object manipulation or contact-rich physical processes, are not included in this branch.

The adaptation preserves the original task goal but changes the expected output form and is used only as a supplementary evaluation. It allows us to compare static target-state inference with procedural video generation under related goals, and helps separate failures caused by goal-state inference from those caused by maintaining a valid process over time. Figure~\ref{fig:apdx_img_io} shows representative input--output examples on three adapted tasks.

% inline helper: 1x3 (input | Nano-Banana-Pro | GPT-Image-2) row inside a
% rounded-dashed border, no background fill. `\imgio{stem}` expands to one
% box; the stem must match Figures/apdx_img_io/<stem>_{in,nbp,gpt}.png.
\providecommand{\imgio}[1]{%
  \begin{tcolorbox}[
    blank, enhanced,
    borderline={0.8pt}{0pt}{black!55,dashed},
    arc=4pt,
    left=4pt, right=4pt, top=4pt, bottom=4pt,
    width=\linewidth,
  ]
    \begin{minipage}[t]{0.32\linewidth}\centering
      \scriptsize\textbf{Input}\par\vspace{2pt}
      \includegraphics[width=\linewidth]{Figures/apdx_img_io/#1_in.jpg}
    \end{minipage}\hfill
    \begin{minipage}[t]{0.32\linewidth}\centering
      \scriptsize\textbf{Nano-Banana-Pro}\par\vspace{2pt}
      \includegraphics[width=\linewidth]{Figures/apdx_img_io/#1_nbp.jpg}
    \end{minipage}\hfill
    \begin{minipage}[t]{0.32\linewidth}\centering
      \scriptsize\textbf{GPT-Image-2}\par\vspace{2pt}
      \includegraphics[width=\linewidth]{Figures/apdx_img_io/#1_gpt.jpg}
    \end{minipage}
  \end{tcolorbox}%
}

\subsection{More Details}
\label{apdx:data_stats}

\begin{table}[h]
\centering
\setlength{\tabcolsep}{6pt}
\renewcommand{\arraystretch}{1.2}
\resizebox{\columnwidth}{!}{%
\begin{tabular}{l c c c c c}
\toprule
& \dombox{domSP}{\makecell{Structured\\Puzzles}}
& \dombox{domVO}{\makecell{Visual\\Organization}}
& \dombox{domSD}{\makecell{Spatiotemporal\\Dynamics}}
& \dombox{domPM}{\makecell{Physical\\Manipulation}}
& \textbf{Total} \\
\midrule
\# tasks & 8 & 5 & 7 & 7 & 27 \\
\bottomrule
\end{tabular}%
}
\caption{Number of tasks per domain. Domains are mutually exclusive; each task belongs to exactly one domain.}
\label{tab:bench_tasks_per_domain}
\end{table}

\begin{table}[h]
\centering
\setlength{\tabcolsep}{6pt}
\renewcommand{\arraystretch}{1.2}
\resizebox{\columnwidth}{!}{%
\begin{tabular}{l c c c c c c c}
\toprule
& \makecell{\textbf{Planning}\\\skillicon{planning}}
& \makecell{\textbf{Physics}\\\skillicon{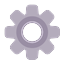}}
& \makecell{\textbf{Spatial}\\\skillicon{spatial}}
& \makecell{\textbf{Attribute}\\\textbf{Grounding}\;\skillicon{attribute_grounding}}
& \makecell{\textbf{Affordance}\\\skillicon{affordance}}
& \makecell{\textbf{Topology}\\\skillicon{topology}}
& \makecell{\textbf{Temporal}\\\skillicon{temporal}} \\
\midrule
\# tasks & 12 & 8 & 7 & 6 & 5 & 3 & 3 \\
\bottomrule
\end{tabular}%
}
\caption{Number of tasks per skill tag. Skill tags are non-exclusive: each task carries one to three tags, so column counts sum to more than the total number of tasks.}
\label{tab:bench_tasks_per_skill}
\end{table}

\begin{figure}[ht]
    \centering
    \includegraphics[width=0.75\linewidth]{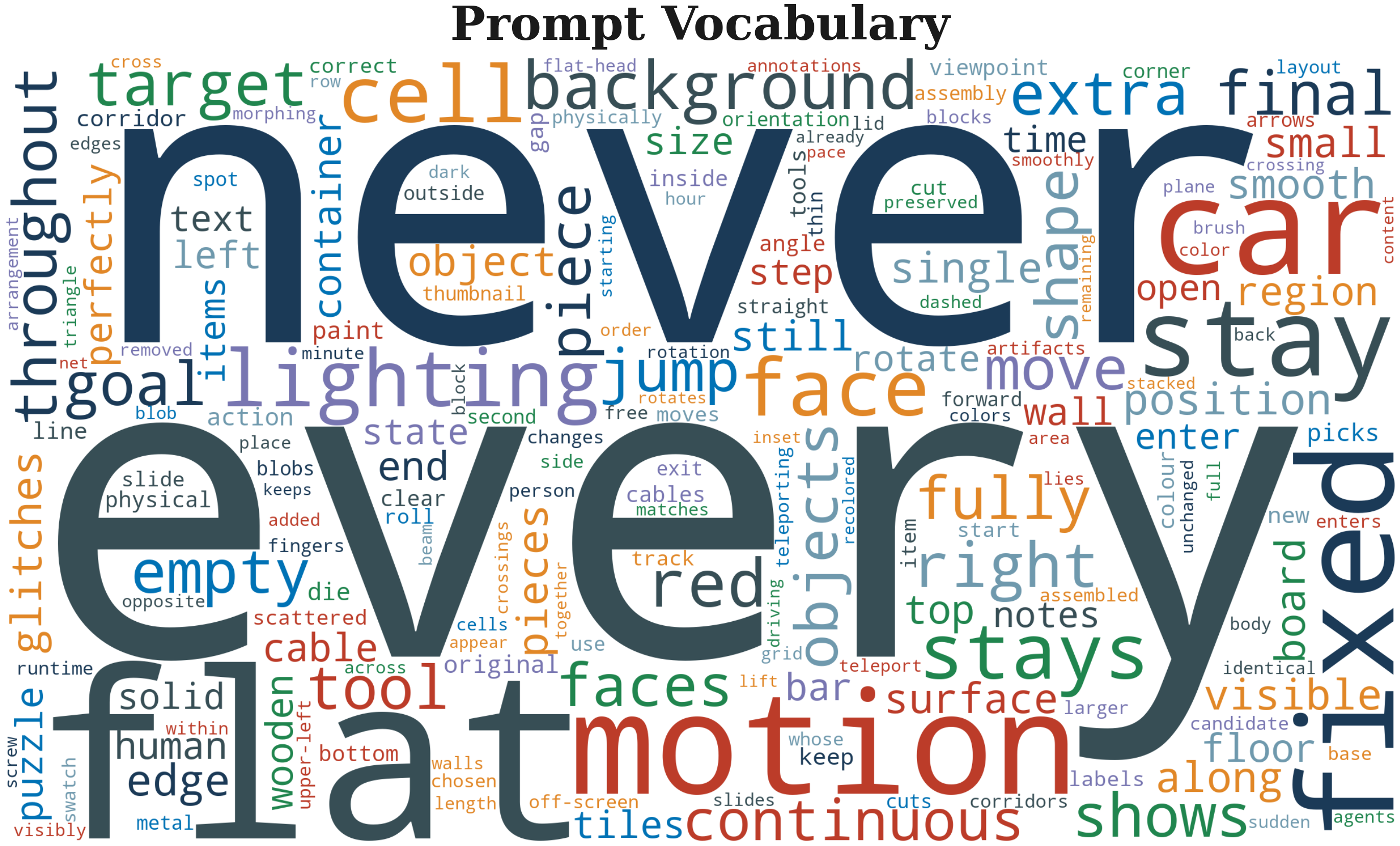}
    \caption{Wordcloud of Task Prompts}
    \label{fig:bench_stats_wordcloud}
\end{figure}

Table~\ref{tab:bench_tasks_per_domain} and Table~\ref{tab:bench_tasks_per_skill} report the number of tasks per domain and per skill tag, respectively. Figure~\ref{fig:bench_stats_wordcloud} shows the word cloud of task prompts. Frequent tokens reflect the process-sensitive and state-evolution nature of the benchmark (e.g., \textit{every}, \textit{flat}, \textit{fixed}, \textit{stays}, \textit{continuous}). 

% \begin{figure}[h]
%     \centering
%     \includegraphics[width=0.95\linewidth]{Figures/apdx_bench_stats_prompt_len_hist.pdf}
%     \caption{Per-instance video-prompt length distribution across all \bench{} instances.}
%     \label{fig:bench_stats_prompt_len_hist}
% \end{figure}

% \section{Task Gallery}

\begin{figure}[ht]
  \centering
  \imgio{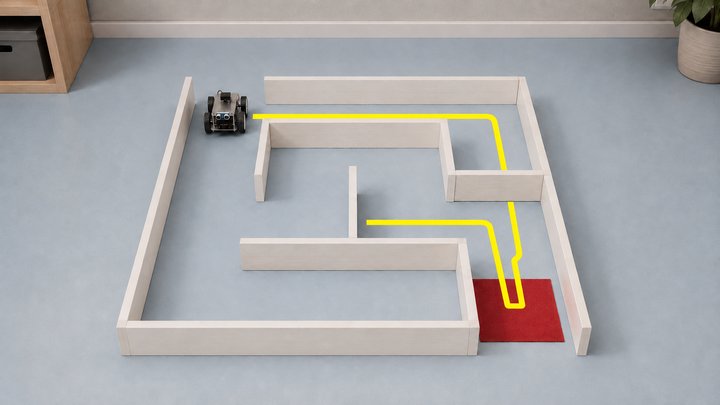}\par\vspace{4pt}
  \imgio{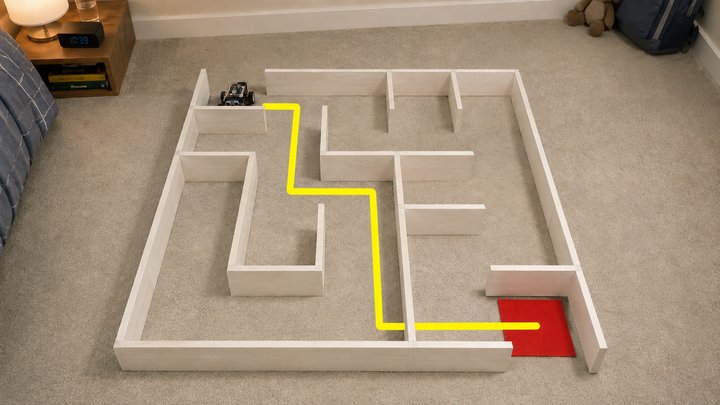}\par\vspace{4pt}
  \imgio{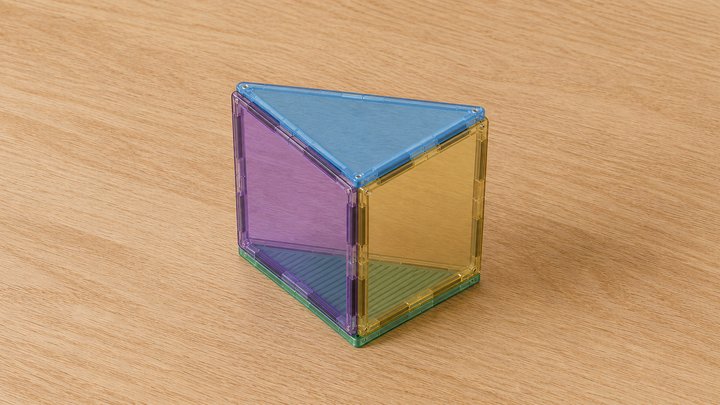}\par\vspace{4pt}
  \imgio{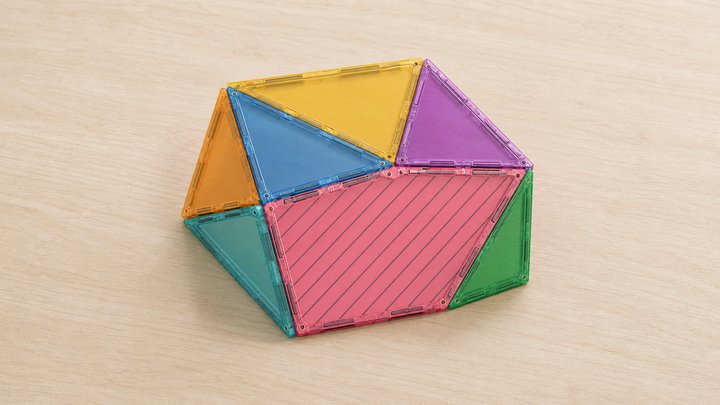}\par\vspace{4pt}
  \imgio{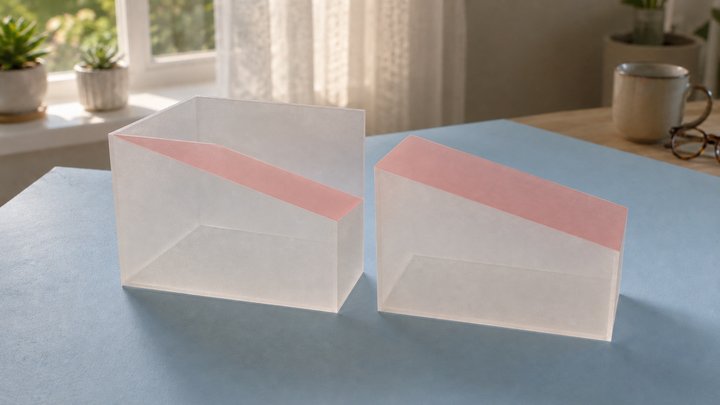}\par\vspace{4pt}
  \imgio{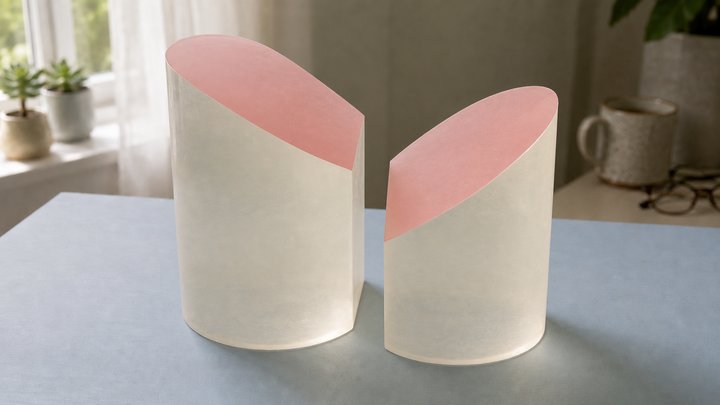}
  \caption{Image-output examples on the adapted single-image subset (top to bottom: \textsc{maze}, \textsc{recover\_2d\_to\_3d}, \textsc{section\_3d\_figure}). Each row shows the input image, the \texttt{Nano-Banana-Pro} output, and the \texttt{GPT-Image-2} output.}
  \label{fig:apdx_img_io}
\end{figure}

\paragraph{Acknowledgement on Task Inspirations}
% \label{apdx:acknowledgement}

While designing our tasks, we surveyed a wide range of benchmarks and studies on reasoning in generative models. A number of our tasks were informed by the task formats, visual settings, or evaluation perspectives introduced in prior efforts. We gratefully acknowledge the following works for inspiring our task collection:

RISE-Bench \citep{zhao2026envisioning},
MORSE-500 \citep{cai2025morse}, 
KRIS-Bench \citep{wu2026kris},
MIRA \citep{zhou2025visualizingstepreasoningmira}, 
Study on Veo3's emergent capability \citep{wiedemer2025video},
VideoThinkBench \citep{tong2025thinking}, 
Gen-ViRe \citep{liu2025can}, 
TiVi-Bench \citep{chen2025tivibench},
MMGR \citep{cai2025mmgrmultimodalgenerativereasoning},
BabyVision \citep{chen2026babyvisionvisualreasoninglanguage}, 
MentisOculi \citep{zeller2026mentisoculi},
VBVR-Bench \citep{vbvr2026}, 
MolmoAct2 \citep{fang2026molmoact2}, 
and others 
\citep{yang2025reasoning, 
luo2025v,
wei2025univideo, 
li2025viper, 
he2025ruler,
qi2026mme, 
wang2026demystifing, 
newman2026video, 
dai2026endocot, 
li2026thinking, 
he2025diffthinker, 
wu2026worldreasonbenchhumanalignedstresstesting}.
\clearpage
\section{Evaluation Details}
\label{apdx:eval_details}
\subsection{Model Generation Configuration}
\label{apdx:model_gen_config}

\paragraph{Video Evaluation (Primary).}
We evaluate our benchmark on a broad set of SOTA video generation models, including:

\texttt{Seedance2.0}~\citep{seedance2026seedance}, 

\texttt{MiniMax-H3}~\citep{minimax_h3_2026},

\texttt{Sora2}~\citep{openai2025sora2}, 

\texttt{Veo3.1}~\citep{google2025veo31}, 

\texttt{Kling~3.0}~\citep{kling2026video30},

\texttt{Wan2.7}~\citep{wan2026wan27}, 

\texttt{Gen~4.5}~\citep{runway2025gen45}, 

\texttt{Wan2.2-I2V-A14B}~\citep{wan2025wan}, 

\texttt{HunyuanVideo-1.5}~\citep{wu2025hunyuanvideo}. 

Due to evaluation cost, for each task we evaluate half of its instances. All tasks follow a unified input format, consisting of a text prompt and an input image. Table~\ref{tab:apdx_video_model_config} shows the specific generation configuration.

\begin{table}[ht]
\centering
\setlength{\tabcolsep}{6pt}
\renewcommand{\arraystretch}{1.1}
\resizebox{\columnwidth}{!}{%
\begin{tabular}{l|cccc}
\toprule
\textbf{Model} & \textbf{Resolution} & \textbf{Duration (s)} & \textbf{FPS} & \textbf{Price (USD / s)} \\
\midrule
\multicolumn{5}{c}{\textit{Open Source}} \\
\midrule
HY1.5       & 1280$\times$720 & 5/10 & 24 & -- \\
% LTX2.3      & 1280$\times$720 & 8 & 24 & -- \\
Wan2.2      & 1280$\times$720 & 5/10 & 16 & -- \\
Mnx-H3      & 960$\times$544 & 5/10 & 24 & -- \\
\midrule
\multicolumn{5}{c}{\textit{Commercial}} \\
\midrule
Sora2       & 1280$\times$720 & 8 & 30 & 0.10 \\
Veo3.1      & 1280$\times$720 & 8 & 24 & 0.20 \\
Sdce2.0     & 1280$\times$720 & 5/10 & 24 & 0.15 \\
Kling3.0    & 1280$\times$720 & 5/10 & 24 & 0.08 \\
Wan2.7      & 1280$\times$720 & 5/10 & 30 & 0.10 \\
Gen4.5      & 1280$\times$720 & 5/10 & 24 & 0.12 \\
\bottomrule
\end{tabular}%
}
\caption{Generation configuration of the video models we evaluate, resolution defaults to $1280\times720$. Price is the per-second generation cost. HY1.5 for \texttt{HunyuanVideo-1.5}, Sdce2.0 for \texttt{Seedance2.0},  Mnx-H3 for \texttt{MiniMax-H3}.}
\label{tab:apdx_video_model_config}
\end{table}

\vspace{-1em}

\paragraph{Image Evaluation (Auxiliary).}
In addition to video generation models, we adapt a subset of benchmark tasks that can be naturally reformulated as single-image output tasks, and evaluate several advanced image generation or unified multi-modal models, including 

\texttt{Nano Banana Pro}~\citep{google2025nanobananapro}, 

\texttt{Qwen-Image-3-Pro}~\citep{qwen_image_3_pro_2026},

\texttt{GPT-Image-2}~\citep{openai_gpt_image_2}, 

\texttt{MAI-Image-2.5-Pro}~\citep{mai_image_2_5_pro_2026},

% \texttt{Grok Imagine}~\citep{xai2026grokimaginequality}, 

\texttt{Seedream4.5}~\citep{bytedance2026seedream45},

\texttt{Flux.2-Max Edit}~\citep{blackforestlabs2025flux2max}, 

\texttt{SenseNova-U1}~\citep{diao2026sensenova}, 

\texttt{JoyAI-Image}~\citep{song2026awaking}, 

\texttt{Qwen-Image-Edit}~\citep{wu2025qwen}, 

\texttt{Step1X-Edit}~\citep{liu2025step1x},

\texttt{BAGEL}~\citep{deng2025emerging}. 

Unless otherwise specified, all image generation models are evaluated at a resolution of $1280 \times 720$.

\paragraph{Access and Budget.} Commercial models are accessed through their APIs and open-source models are run locally on NVIDIA H20 GPUs.

\subsection{Full Results of Evaluation}
\label{apdx:full_eval_res}

\paragraph{Per-metric full results} Table~\ref{tab:full_video_res} reports the full per-metric breakdown (Completeness, Checklist Score, and Final) for all video generation models, complementing the aggregated score summary (Table~\ref{tab:main_res}) in the main part. The three columns aggregate over instances independently: Comp.\ and Rub.\ are the per-cell means of Completeness and Rubric scores, while Final is the per-cell mean of the per-instance product $\textrm{Comp.}\times\textrm{Rub.}$. Because the mean of products is not equal to the product of means, $\textrm{Comp.}\times\textrm{Rub.}\neq\textrm{Final}$ within a cell in general; this difference reflects the alignment between completeness and rubric performance at the instance level rather than any inconsistency.

\paragraph{Difficulty-level fluctuations.}
The three difficulty levels are designed to reflect increasing task complexity, but per-domain and per-model scores need not be strictly monotonic. Current video models remain unstable on many tasks, and small changes in layout, object configuration, or required action pattern can interact differently with each model's generative priors. Fine-grained averages may show local inversions, such as a Mid score exceeding an Easy score. We therefore interpret difficulty trends mainly at the aggregate level rather than requiring monotonicity in every model--domain cell.

\begin{table*}[ht]
\centering
\setlength{\tabcolsep}{5pt}
\renewcommand{\arraystretch}{1.1}
\resizebox{\textwidth}{!}{%
\newcolumntype{G}{>{\columncolor{gray!8}[2pt][2pt]}c}
\newcommand{\hl}[2]{\multicolumn{1}{>{\columncolor{#1}[2pt][2pt]}c}{#2}}
\begin{tabular}{ll *{10}{c@{\hskip 3pt}c@{\hskip 3pt}G}}
\toprule
\multirow{3}{*}{\textbf{Category}} & \multirow{3}{*}{\textbf{Level}}
 & \multicolumn{18}{c}{\textit{Commercial}}
 & \multicolumn{12}{c}{\textit{Open Source}} \\
\cmidrule(lr){3-20} \cmidrule(lr){21-32}
 &
 & \multicolumn{3}{c}{\textbf{Sdce2.0}} & \multicolumn{3}{c}{\textbf{Sora2}} & \multicolumn{3}{c}{\textbf{Veo3.1}} & \multicolumn{3}{c}{\textbf{Kling3.0}} & \multicolumn{3}{c}{\textbf{Wan2.7}} & \multicolumn{3}{c}{\textbf{Gen4.5}}
 & \multicolumn{3}{c}{\textbf{Mnx-H3}} & \multicolumn{3}{c}{\textbf{HY1.5}} & \multicolumn{3}{c}{\textbf{Wan2.2}} & \multicolumn{3}{c}{\textbf{VBVR-Wan2.2}} \\
\cmidrule(lr){3-5} \cmidrule(lr){6-8} \cmidrule(lr){9-11} \cmidrule(lr){12-14} \cmidrule(lr){15-17} \cmidrule(lr){18-20} \cmidrule(lr){21-23} \cmidrule(lr){24-26} \cmidrule(lr){27-29} \cmidrule(lr){30-32}
 &
 & Comp. & Rub. & Final & Comp. & Rub. & Final & Comp. & Rub. & Final & Comp. & Rub. & Final & Comp. & Rub. & Final & Comp. & Rub. & Final & Comp. & Rub. & Final & Comp. & Rub. & Final & Comp. & Rub. & Final & Comp. & Rub. & Final \\
\arrayrulecolor{gray!50}\midrule\midrule\arrayrulecolor{black}
\multirow{4}{*}{\makecell{Visual Org-\\anization}} & Easy & 82.0 & 84.8 & \hl{hlbestb}{72.6} & 70.0 & 76.1 & 55.6 & 66.0 & 72.5 & 50.3 & 82.0 & 80.4 & \hl{hlsecb}{67.6} & 52.0 & 68.6 & 37.0 & 76.0 & 77.5 & 60.5 & 56.0 & 73.4 & 42.8 & 36.0 & 66.7 & 26.6 & 46.0 & 60.2 & 30.4 & 58.0 & 73.6 & 46.1 \\
 & Mid & 70.0 & 77.3 & \hl{hlbestb}{56.4} & 58.0 & 62.7 & 37.0 & 66.0 & 64.4 & 43.3 & 62.0 & 73.2 & \hl{hlsecb}{47.5} & 44.0 & 67.1 & 32.2 & 54.0 & 68.1 & 38.3 & 56.0 & 72.6 & 42.3 & 28.0 & 61.0 & 17.9 & 24.0 & 59.9 & 15.3 & 46.0 & 65.7 & 32.4 \\
 & Hard & 70.0 & 74.1 & \hl{hlbestb}{53.5} & 56.0 & 66.7 & 39.1 & 62.0 & 69.6 & \hl{hlsecb}{44.0} & 58.0 & 72.8 & 43.7 & 46.0 & 66.7 & 32.5 & 58.0 & 70.6 & 41.9 & 52.0 & 74.8 & 39.8 & 38.0 & 62.6 & 24.7 & 26.0 & 64.1 & 18.5 & 48.0 & 68.6 & 34.3 \\
\arrayrulecolor{gray!50}\cline{3-20}\cline{21-32}\arrayrulecolor{black}
 & Avg. & 74.0 & 78.8 & \hl{hlbestb}{60.8} & 61.3 & 68.5 & 43.9 & 64.7 & 68.8 & 45.9 & 67.3 & 75.4 & \hl{hlsecb}{52.9} & 47.3 & 67.5 & 33.9 & 62.7 & 72.0 & 46.9 & 54.7 & 73.6 & 41.6 & 34.0 & 63.4 & 23.1 & 32.0 & 61.4 & 21.4 & 50.7 & 69.3 & 37.6 \\
\arrayrulecolor{gray!50}\midrule\arrayrulecolor{black}
\multirow{4}{*}{\makecell{Spatiotemporal\\Dynamics}} & Easy & 81.1 & 67.3 & \hl{hlbestb}{54.6} & 62.9 & 63.9 & \hl{hlsecb}{45.3} & 46.2 & 58.7 & 28.1 & 62.5 & 70.2 & 45.2 & 57.1 & 64.3 & 38.9 & 56.2 & 66.9 & 39.6 & 52.9 & 76.0 & 44.3 & 50.0 & 64.7 & 34.7 & 60.0 & 62.1 & 38.7 & 57.1 & 66.6 & 37.0 \\
 & Mid & 68.6 & 66.6 & \hl{hlbestb}{47.4} & 52.9 & 60.6 & 34.6 & 37.2 & 54.7 & 21.2 & 56.4 & 62.2 & 35.2 & 61.4 & 66.0 & \hl{hlsecb}{42.6} & 54.3 & 60.0 & 34.6 & 51.4 & 74.3 & 40.7 & 41.4 & 61.7 & 27.2 & 52.9 & 56.9 & 32.0 & 30.0 & 71.0 & 21.6 \\
 & Hard & 50.0 & 62.3 & \hl{hlsecb}{33.4} & 41.4 & 57.3 & 24.4 & 28.6 & 52.4 & 16.9 & 47.2 & 58.2 & 28.2 & 42.9 & 61.1 & 28.9 & 50.0 & 54.6 & 29.5 & 47.1 & 71.8 & \hl{hlbestb}{35.9} & 38.2 & 58.2 & 23.9 & 31.4 & 55.5 & 19.9 & 40.0 & 68.2 & 28.2 \\
\arrayrulecolor{gray!50}\cline{3-20}\cline{21-32}\arrayrulecolor{black}
 & Avg. & 66.8 & 65.4 & \hl{hlbestb}{45.3} & 52.4 & 60.6 & 34.8 & 37.7 & 55.4 & 22.3 & 55.7 & 63.7 & 36.5 & 53.8 & 63.8 & 36.8 & 53.6 & 60.8 & 34.8 & 50.5 & 74.0 & \hl{hlsecb}{40.3} & 43.3 & 61.5 & 28.7 & 48.1 & 58.2 & 30.2 & 42.4 & 68.6 & 29.0 \\
\arrayrulecolor{gray!50}\midrule\arrayrulecolor{black}
\multirow{4}{*}{\makecell{Structured\\Puzzles}} & Easy & 64.3 & 67.1 & 46.9 & 51.4 & 72.1 & 40.5 & 41.4 & 64.3 & 31.8 & 62.9 & 68.4 & 45.3 & 30.0 & 66.8 & 25.6 & 41.4 & 59.6 & 29.2 & 58.6 & 80.5 & \hl{hlsecb}{51.9} & 14.3 & 64.5 & 9.9 & 22.9 & 61.2 & 17.1 & 68.6 & 82.0 & \hl{hlbestb}{57.9} \\
 & Mid & 70.0 & 63.8 & \hl{hlsecb}{45.1} & 34.3 & 63.8 & 25.8 & 32.9 & 59.7 & 21.8 & 61.4 & 61.7 & 38.3 & 31.4 & 60.9 & 23.3 & 35.7 & 51.7 & 20.7 & 60.0 & 81.8 & \hl{hlbestb}{52.3} & 11.4 & 67.4 & 8.8 & 12.9 & 55.7 & 8.4 & 54.3 & 77.0 & 45.0 \\
 & Hard & 60.0 & 63.6 & 41.8 & 34.3 & 60.5 & 22.4 & 22.9 & 53.4 & 13.5 & 48.6 & 57.1 & 28.9 & 34.3 & 61.0 & 24.7 & 34.3 & 54.5 & 18.7 & 58.6 & 80.2 & \hl{hlsecb}{47.6} & 10.0 & 63.4 & 7.9 & 10.0 & 51.9 & 5.6 & 60.0 & 79.6 & \hl{hlbestb}{50.0} \\
\arrayrulecolor{gray!50}\cline{3-20}\cline{21-32}\arrayrulecolor{black}
 & Avg. & 64.8 & 64.8 & 44.6 & 40.0 & 65.5 & 29.5 & 32.4 & 59.1 & 22.4 & 57.6 & 62.4 & 37.5 & 31.9 & 62.9 & 24.5 & 37.1 & 55.3 & 22.9 & 59.0 & 80.8 & \hl{hlsecb}{50.6} & 11.9 & 65.1 & 8.9 & 15.2 & 56.3 & 10.4 & 61.0 & 79.5 & \hl{hlbestb}{51.0} \\
\arrayrulecolor{gray!50}\midrule\arrayrulecolor{black}
\multirow{4}{*}{\makecell{Physical\\Manipulation}} & Easy & 78.6 & 76.8 & \hl{hlbestb}{64.4} & 61.4 & 66.8 & 47.9 & 62.9 & 67.2 & 49.6 & 78.6 & 71.9 & \hl{hlsecb}{60.6} & 71.4 & 70.6 & 55.8 & 67.1 & 69.3 & 52.1 & 72.9 & 76.7 & 58.6 & 35.7 & 58.5 & 22.5 & 44.3 & 66.2 & 33.7 & 55.7 & 78.6 & 46.7 \\
 & Mid & 81.4 & 69.8 & \hl{hlbestb}{59.2} & 60.0 & 62.7 & 41.8 & 57.4 & 59.2 & 40.4 & 68.6 & 68.2 & \hl{hlsecb}{51.2} & 75.7 & 61.7 & 48.2 & 67.1 & 58.9 & 43.4 & 58.6 & 67.3 & 41.5 & 28.6 & 57.0 & 16.7 & 32.9 & 64.8 & 22.1 & 57.1 & 72.9 & 43.1 \\
 & Hard & 72.9 & 58.0 & \hl{hlsecb}{44.4} & 51.4 & 58.9 & 32.6 & 55.7 & 58.8 & 37.3 & 72.9 & 58.9 & \hl{hlbestb}{45.6} & 60.0 & 59.4 & 37.1 & 61.4 & 58.5 & 39.5 & 52.9 & 60.5 & 33.0 & 21.4 & 57.1 & 11.9 & 30.0 & 60.8 & 17.5 & 48.6 & 72.1 & 37.9 \\
\arrayrulecolor{gray!50}\cline{3-20}\cline{21-32}\arrayrulecolor{black}
 & Avg. & 77.6 & 68.2 & \hl{hlbestb}{56.0} & 57.6 & 62.8 & 40.8 & 58.7 & 61.8 & 42.5 & 73.3 & 66.3 & \hl{hlsecb}{52.5} & 69.0 & 63.9 & 47.1 & 65.2 & 62.3 & 45.0 & 61.4 & 68.2 & 44.4 & 28.6 & 57.5 & 17.0 & 35.7 & 63.9 & 24.4 & 53.8 & 74.6 & 42.6 \\
\arrayrulecolor{gray!50}\midrule\midrule\arrayrulecolor{black}
\multicolumn{2}{c}{\textbf{Overall}} & 70.5 & 68.6 & \hl{hlbestb}{51.0} & 52.2 & 64.0 & 36.7 & 46.9 & 60.6 & 32.0 & 63.0 & 66.3 & 44.0 & 50.8 & 64.3 & 35.7 & 54.1 & 61.9 & 36.6 & 56.5 & 74.2 & \hl{hlsecb}{44.4} & 29.0 & 61.8 & 19.1 & 32.8 & 59.8 & 21.6 & 52.1 & 73.3 & 40.2 \\
\bottomrule
\end{tabular}%
}
\caption{Full per-metric evaluation results of video generation models across our reasoning categories at Easy/Mid/Hard difficulty plus a per-category average, scored by three metrics: Completeness (Comp.), Rubric Score (Rub.), and the final aggregated score (Final). The main-paper Table~\ref{tab:main_res} reports the final aggregated score only. Higher is better; the \colorbox{hlbestb}{\strut best} and \colorbox{hlsecb}{\strut second-best} Final scores in each row are highlighted. HY1.5 for \texttt{HunyuanVideo-1.5}, Sdce2.0 for \texttt{Seedance2.0}, Mnx-H3 for \texttt{MiniMax-H3}. Note that, within each cell, generally,  $\textrm{Comp.}\times\textrm{Rub.}\neq\textrm{Final}$: this is because Comp.\ and Rub.\ are the mean Completeness and Rubric scores aggregated independently across instances, whereas Final is the mean of the per-instance product $\textrm{Comp.}\times\textrm{Rub.}$ (i.e., the mean of products, not the product of means).}
\label{tab:full_video_res}
\end{table*}

\paragraph{Strict success rate} Table~\ref{tab:apdx_strict_sr} additionally reports the \textbf{strict success rate}, where an instance is counted as a success only if the judge marks it as fully completing the task (perfect Completeness) without violating any rules (perfect Rubric Score). The numbers are sharply lower than the aggregated scores in Table~\ref{tab:main_res}, reflecting how rare end-to-end success still is for current video models under this strict criterion. We also include the human ceiling performance for reference, demonstrating the gap of visual intelligence between current models and average humans, as detailed in Appendix~\ref{apdx:human_ceiling}. 

\begin{table*}[t]
\centering
\setlength{\tabcolsep}{8pt}
\renewcommand{\arraystretch}{1.1}
\resizebox{\textwidth}{!}{%
% \newcolumntype{G}{>{\columncolor{gray!8}[2pt][2pt]}c}  % alternating grey shading (disabled; restore this line to re-enable)
\newcolumntype{G}{c}
% Highlight: pill-shaped \colorbox painted ONLY around the number, not the
% whole cell. Tight \fboxsep so the pill hugs the digits.
\newcommand{\hl}[2]{{\setlength{\fboxsep}{2pt}\colorbox{#1}{#2}}}
\begin{tabular}{ll c c G c G c G c G c G}
\toprule
\multirow{2}{*}{\textbf{Category}} & \multirow{2}{*}{\textbf{Level}}
 & \multirow{2}{*}{\textcolor{gray}{\textbf{Human}}}
 & \multicolumn{6}{c}{\textit{Commercial}}
 & \multicolumn{4}{c}{\textit{Open Source}} \\
\cmidrule(lr){4-9} \cmidrule(lr){10-13}
 & &
 & \textbf{Sdce2.0} & \textbf{Sora2} & \textbf{Veo3.1} & \textbf{Kling3.0} & \textbf{Wan2.7} & \textbf{Gen4.5} & \textbf{Mnx-H3} & \textbf{HY1.5} & \textbf{Wan2.2} & \textbf{VBVR-Wan2.2} \\
\arrayrulecolor{gray!50}\midrule\midrule\arrayrulecolor{black}
\multirow{4}{*}{\makecell{Visual Org-\\anization}} & Easy & \textcolor{gray}{100.0} & \hl{hlbestb}{40.0} & 16.0 & 4.0 & \hl{hlsecb}{24.0} & 8.0 & 8.0 & 12.0 & 4.0 & 4.0 & 16.0 \\
 & Mid & \textcolor{gray}{97.5} & \hl{hlbestb}{16.0} & 4.0 & 4.0 & \hl{hlsecb}{12.0} & 4.0 & 8.0 & 8.0 & 0.0 & 0.0 & 4.0 \\
 & Hard & \textcolor{gray}{92.5} & \hl{hlbestb}{12.0} & 0.0 & 4.0 & \hl{hlbestb}{12.0} & 4.0 & \hl{hlsecb}{8.0} & \hl{hlsecb}{8.0} & 0.0 & 4.0 & 4.0 \\
\arrayrulecolor{gray!50}\cline{3-3}\cline{4-9}\cline{10-13}\arrayrulecolor{black}
 & Avg. & \textcolor{gray}{96.7} & \hl{hlbestb}{22.7} & 6.7 & 4.0 & \hl{hlsecb}{16.0} & 5.3 & 8.0 & 9.3 & 1.3 & 2.7 & 8.0 \\
\arrayrulecolor{gray!50}\midrule\arrayrulecolor{black}
\multirow{4}{*}{\makecell{Spatiotemporal\\Dynamics}} & Easy & \textcolor{gray}{100.0} & 5.4 & 5.7 & 0.0 & \hl{hlbestb}{10.0} & 5.7 & 5.0 & \hl{hlsecb}{8.6} & 0.0 & 2.9 & 5.7 \\
 & Mid & \textcolor{gray}{98.0} & 5.7 & \hl{hlsecb}{8.6} & 0.0 & 2.6 & \hl{hlbestb}{11.4} & 0.0 & \hl{hlsecb}{8.6} & 5.7 & 2.9 & \hl{hlsecb}{8.6} \\
 & Hard & \textcolor{gray}{92.0} & \hl{hlbestb}{5.7} & \hl{hlsecb}{2.9} & \hl{hlsecb}{2.9} & 2.8 & 0.0 & 0.0 & \hl{hlbestb}{5.7} & 0.0 & \hl{hlsecb}{2.9} & \hl{hlsecb}{2.9} \\
\arrayrulecolor{gray!50}\cline{3-3}\cline{4-9}\cline{10-13}\arrayrulecolor{black}
 & Avg. & \textcolor{gray}{96.7} & 5.6 & \hl{hlsecb}{5.7} & 0.9 & 5.2 & \hl{hlsecb}{5.7} & 1.8 & \hl{hlbestb}{7.6} & 1.9 & 2.9 & \hl{hlsecb}{5.7} \\
\arrayrulecolor{gray!50}\midrule\arrayrulecolor{black}
\multirow{4}{*}{\makecell{Structured\\Puzzles}} & Easy & \textcolor{gray}{100.0} & 11.4 & 11.4 & 11.4 & 8.6 & 11.4 & 0.0 & \hl{hlbestb}{22.9} & 2.9 & 2.9 & \hl{hlsecb}{20.0} \\
 & Mid & \textcolor{gray}{100.0} & 11.4 & 2.9 & 0.0 & 2.9 & 11.4 & 0.0 & \hl{hlbestb}{20.0} & 0.0 & 0.0 & \hl{hlsecb}{14.3} \\
 & Hard & \textcolor{gray}{84.4} & \hl{hlsecb}{11.4} & 2.9 & 0.0 & 2.9 & 5.7 & 0.0 & \hl{hlsecb}{11.4} & 0.0 & 0.0 & \hl{hlbestb}{17.1} \\
\arrayrulecolor{gray!50}\cline{3-3}\cline{4-9}\cline{10-13}\arrayrulecolor{black}
 & Avg. & \textcolor{gray}{94.8} & 11.4 & 5.7 & 3.8 & 4.8 & 9.5 & 0.0 & \hl{hlbestb}{18.1} & 1.0 & 1.0 & \hl{hlsecb}{17.1} \\
\arrayrulecolor{gray!50}\midrule\arrayrulecolor{black}
\multirow{4}{*}{\makecell{Physical\\Manipulation}} & Easy & \textcolor{gray}{100.0} & \hl{hlbestb}{40.0} & 14.3 & 20.0 & \hl{hlsecb}{28.6} & 20.0 & 22.9 & \hl{hlsecb}{28.6} & 0.0 & 8.6 & 17.1 \\
 & Mid & \textcolor{gray}{100.0} & \hl{hlsecb}{11.4} & 2.9 & \hl{hlbestb}{14.7} & 8.6 & 2.9 & 8.6 & 5.7 & 0.0 & 0.0 & \hl{hlsecb}{11.4} \\
 & Hard & \textcolor{gray}{98.3} & \hl{hlsecb}{5.7} & 2.9 & 2.9 & \hl{hlsecb}{5.7} & \hl{hlbestb}{8.6} & \hl{hlsecb}{5.7} & 2.9 & 0.0 & 0.0 & \hl{hlbestb}{8.6} \\
\arrayrulecolor{gray!50}\cline{3-3}\cline{4-9}\cline{10-13}\arrayrulecolor{black}
 & Avg. & \textcolor{gray}{99.4} & \hl{hlbestb}{19.0} & 6.7 & 12.5 & \hl{hlsecb}{14.3} & 10.5 & 12.4 & 12.4 & 0.0 & 2.9 & 12.4 \\
\arrayrulecolor{gray!50}\midrule\midrule\arrayrulecolor{black}
\multicolumn{2}{c}{\textbf{Overall}} & \textcolor{gray}{97.1} & \hl{hlbestb}{\textbf{14.0}} & 6.2 & 5.3 & 9.5 & 7.9 & 5.3 & \hl{hlsecb}{12.1} & 1.0 & 2.3 & 11.0 \\
\bottomrule
\end{tabular}%
}
\caption{\textbf{Strict success rate} (\%) of video generation models on \bench. Layout mirrors Table~\ref{tab:main_res}. Per row, the \colorbox{hlbestb}{\strut best} and \colorbox{hlsecb}{\strut second-best} scores among the video models are highlighted (Human reference excluded); tied bests share the darker shade. Cells marked ``-'' have no judged instances yet. HY1.5 for \texttt{HunyuanVideo-1.5}, Sdce2.0 for \texttt{Seedance2.0}, Mnx-H3 for \texttt{MiniMax-H3}.}
\label{tab:apdx_strict_sr}
\end{table*}

% \begin{figure}[ht]
%     \centering
%     \begin{subfigure}[t]{0.48\linewidth}
%         \centering
%         \includegraphics[width=\linewidth]{Figures/apdx_class_corr.pdf}
%         \caption{Domain $\times$ Domain.}
%         \label{fig:apdx_class_corr}
%     \end{subfigure}\hfill
%     \begin{subfigure}[t]{0.48\linewidth}
%         \centering
%         \includegraphics[width=\linewidth]{Figures/apdx_skill_corr.pdf}
%         \caption{Skill-tag $\times$ Skill-tag.}
%         \label{fig:apdx_skill_corr}
%     \end{subfigure}
%     \caption{Within-axis pairwise Pearson correlations across all tasks in \bench: domains (left) and skill tags (right).}
%     \label{fig:apdx_taxonomy_corr}
% \end{figure}

% \paragraph{Within-taxonomy correlations.}
% To complement the per-domain / per-skill numbers above, Figure~\ref{fig:apdx_taxonomy_corr} shows the pairwise Pearson correlations within each axis of the taxonomy. The Domain heatmap (left) is constrained by mutual exclusivity (every task lives in exactly one domain) so all off-diagonal entries are mildly negative. The Skill-tag heatmap (right) reveals the strongest negative coupling between \textit{Planning} and \textit{Physics} ($-0.43$) and between \textit{Spatial} and \textit{Planning} ($-0.39$), indicating that planning-heavy tasks tend to live apart from both physics-driven and purely spatial tasks; most remaining pairs sit close to zero, suggesting that the seven skill tags carve up the suite along largely independent axes. 

\subsection{Human Ceiling Evaluation}
\label{apdx:human_ceiling}

\paragraph{Setup.}
To estimate a human performance ceiling, we conduct a human study on our benchmark. Since the tasks are designed to rely on everyday visual and physical intuition, the study aims to verify task solvability and clarity rather than measure human response speed. For each task, pilot instances are randomly sampled across different difficulty levels. We first run a pilot study and compute the average response time for each task family; the response-time limit is then set to the 90th percentile of these averaged pilot response times. This yields a time budget that covers most normal human solutions while avoiding excessive delays.

For each task and each difficulty level, we randomly sample half of the instances to form the evaluation set. Accuracy is averaged across participants and instances. Due to the heterogeneous nature of the tasks, we allow different response formats. For example, some tasks naturally require spatial annotation, such as maze solving and Euler path tracing, so participants can annotate the original image, while for tasks that are more naturally solved through verbal reasoning, participants describe their solution orally to the interviewer.

\paragraph{Metrics of Human Response.}
Video model outputs can exhibit a wide range of unintended behaviors, such as altering the task structure, changing the background, or producing physically inconsistent transitions. We therefore use fine-grained rubrics to evaluate generated videos. For human participants, however, these video-specific checklist violations almost never occur. As a result, checklist-based scoring is less meaningful for human ceiling evaluation. We instead report the strict success rate, where each task instance is judged with a binary label indicating whether the participant successfully solves the task or substantially achieves the intended objective. All participants are undergraduate or graduate students with a computer science background, and receive brief instructions and task-specific examples before the study. The human ceiling evaluation results are in Table~\ref{tab:apdx_strict_sr}, together with the strict success rate of video models. 

\begin{figure}[ht]
\begin{tcolorbox}[
  enhanced, frame hidden, boxrule=0pt,
  colback=black!6, arc=6pt,
  left=6pt, right=6pt, top=6pt, bottom=6pt
]
\centering
\begin{subfigure}{0.48\linewidth}
  \centering
  \includegraphics[width=\linewidth]{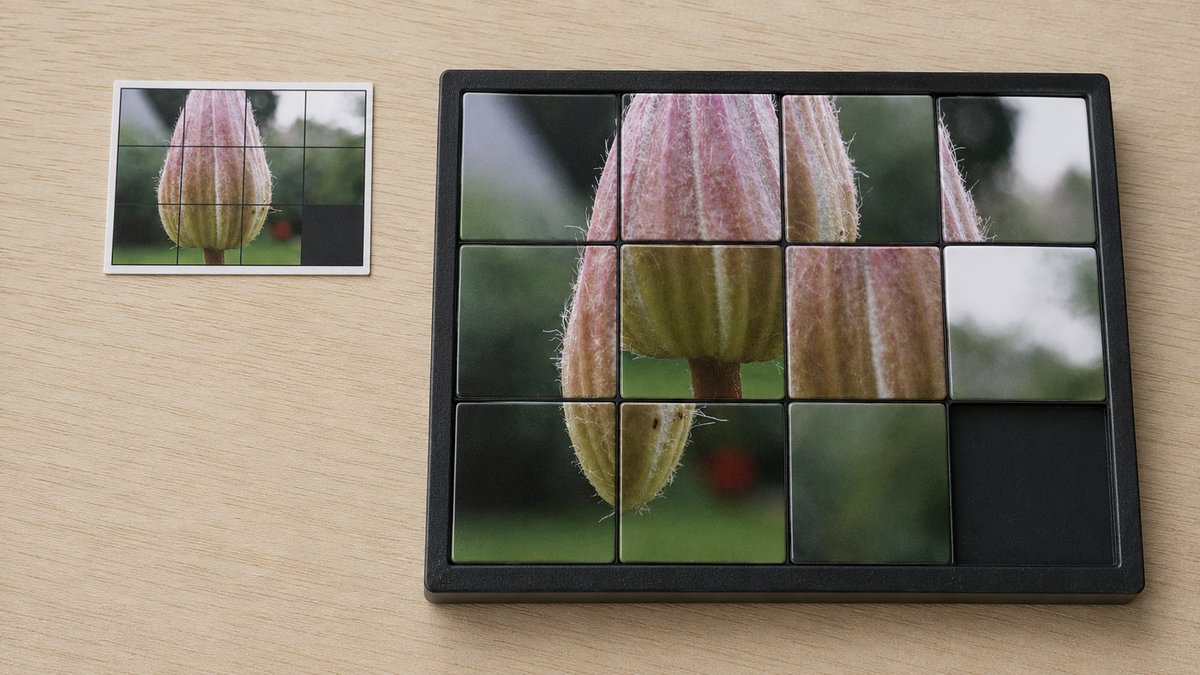}
  \caption{\textsc{sliding\_puzzle}}
\end{subfigure}\hfill
\begin{subfigure}{0.48\linewidth}
  \centering
  \includegraphics[width=\linewidth]{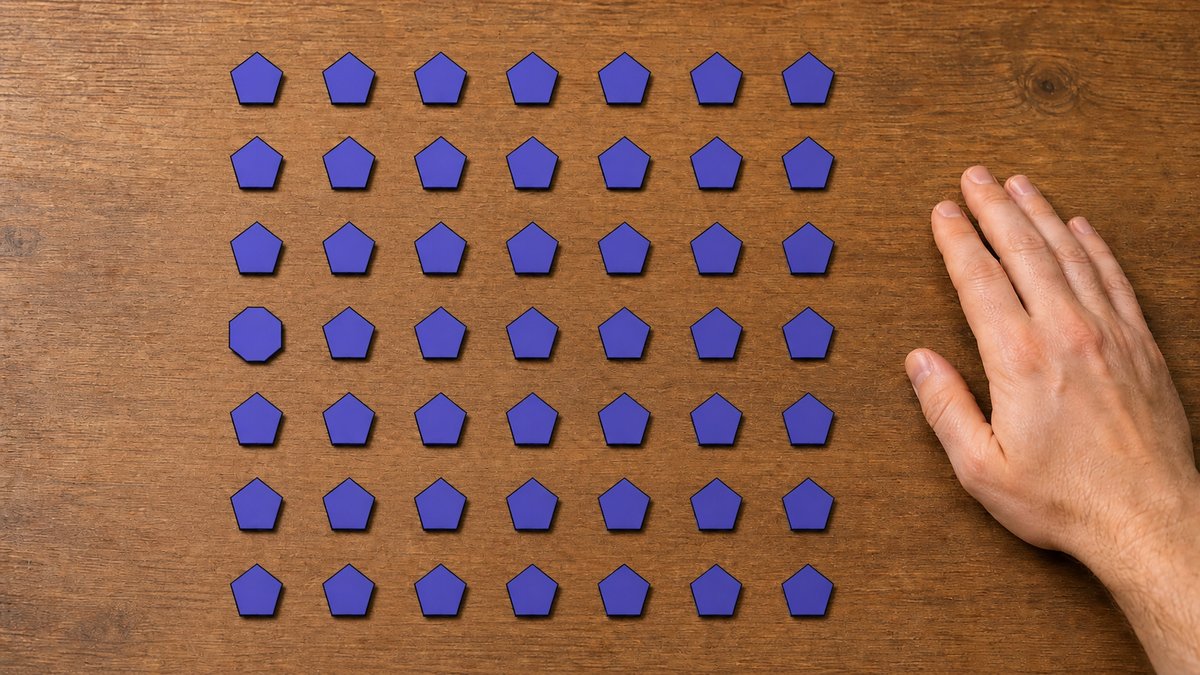}
  \caption{\textsc{pick\_unique}}
\end{subfigure}

\vspace{0.6em}

\begin{subfigure}{0.48\linewidth}
  \centering
  \includegraphics[width=\linewidth]{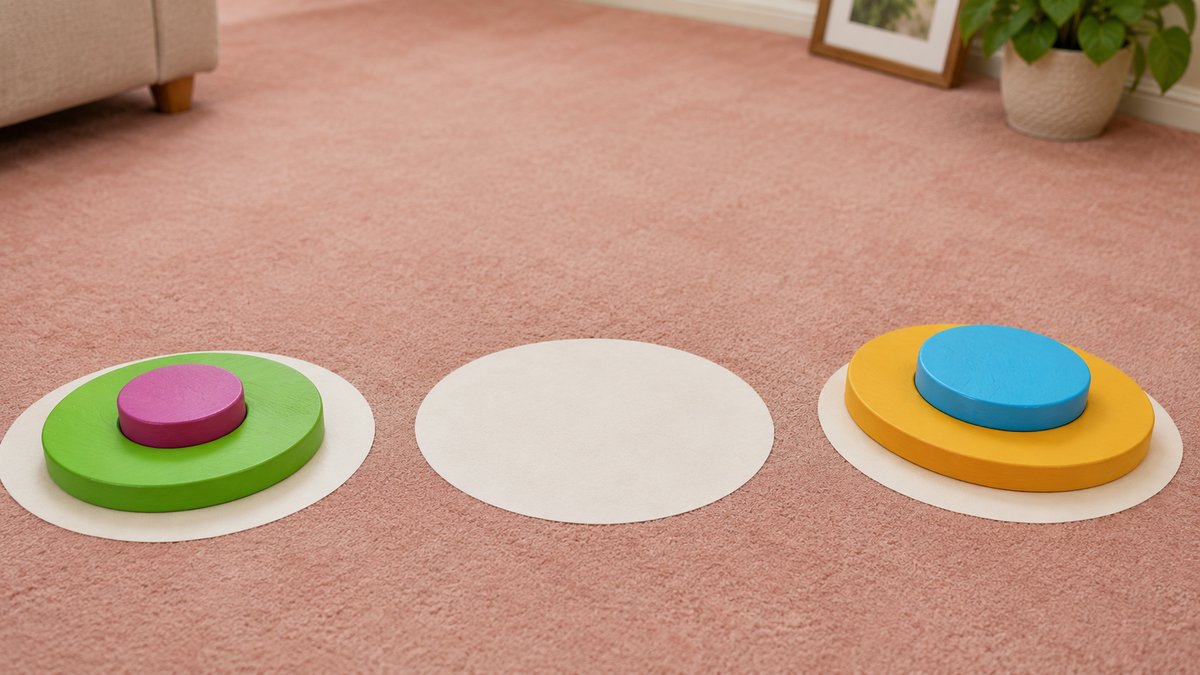}
  \caption{\textsc{hanoi\_tower}}
\end{subfigure}\hfill
\begin{subfigure}{0.48\linewidth}
  \centering
  \includegraphics[width=\linewidth]{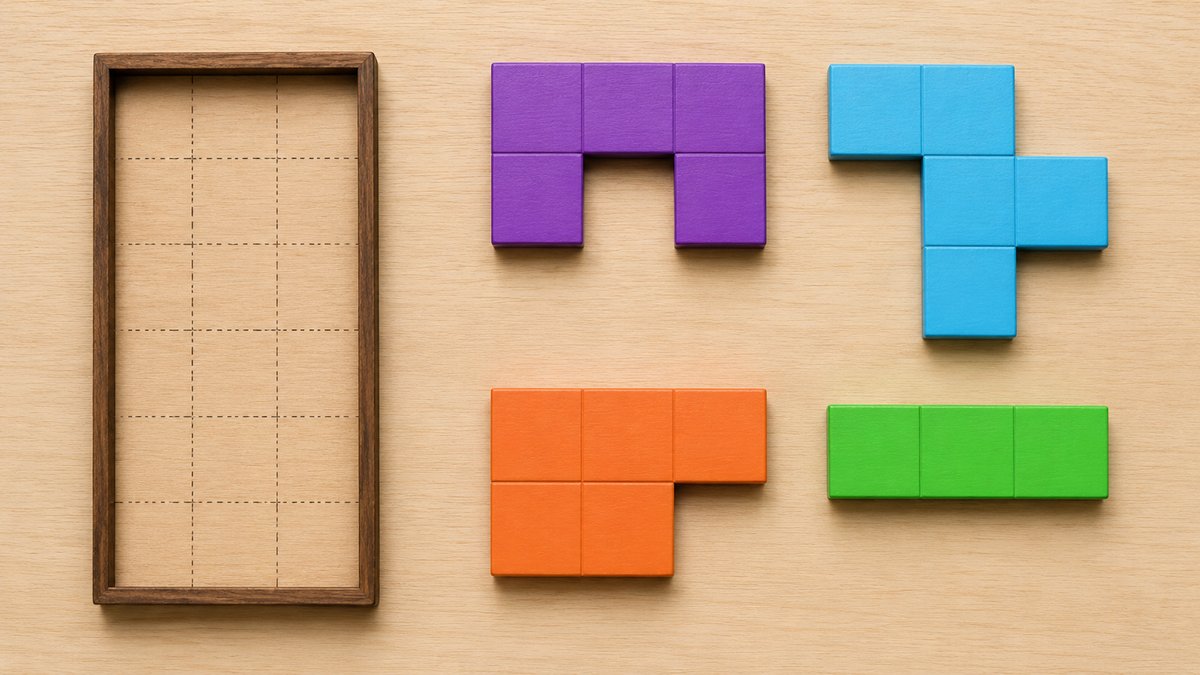}
  \caption{\textsc{polyform\_tiling}}
\end{subfigure}
\end{tcolorbox}
\caption{Representative human failure cases, most cases are sourced from the highest difficulty level of the task.}
\label{fig:human_failure_examples}
\end{figure}

\paragraph{Failure Examples.}
The ceiling is not perfect: even at the highest difficulty of each task, a non-trivial fraction of participants fail within the response time budget. Fig.~\ref{fig:human_failure_examples} collects representative failure cases of several tasks. the failures include often miscounted moves (\textsc{hanoi tower}, \textsc{sliding puzzle}), missed identification of the unique cell among many distractors (\textsc{pick unique}), or geometric mis-fits in tiling (\textsc{polyform tiling}).

\subsection{Stability Analysis under Full-Set Evaluation}
\label{apdx:stable_eval_subset}

The main results in Section~\ref{4_main_results} are computed on a fixed subset of \bench: for each (model, task) cell, we generate and evaluate videos for \textbf{half of the instances at each difficulty level}, keeping the video-generation cost manageable. To verify that this half-subset is a reliable proxy for the full pool, we additionally run a full-instance evaluation on a representative slice: 5 video models and 6 tasks spanning the four domains. For this slice, we generate videos for all the instances per level and evaluate the complete set with the same judge protocol. We summarize the half-subset scores with the corresponding full-set scores in Table~\ref{tab:apdx_eval_stability}. Across the 5 models, the half-subset score tracks the full-set score within $5$ percentage points on average and the model ranking is preserved, suggesting that the main paper's half-subset evaluation is a faithful proxy for the full pool.

\begin{table}[h]
\centering
\setlength{\tabcolsep}{8pt}

\renewcommand{\arraystretch}{1.1}
\small
\resizebox{\linewidth}{!}{%
\begin{tabular}{l ccc}
\toprule
\textbf{Video Model} & \textbf{Full (N=10)} & \textbf{Half (N=5, main)} & \textbf{$\Delta$ (Half$-$Full)} \\
\midrule
\texttt{Kling3.0}       & 34.8 & 33.2 & $-1.6$ \\
\texttt{Sora2}          & 30.9 & 27.2 & $-3.7$ \\
\texttt{Veo3.1}         & 27.8 & 25.1 & $-2.7$ \\
\texttt{Wan2.2}         & 14.3 & 14.0 & $-0.3$ \\
\texttt{HY1.5}          & 14.7 & 11.7 & $-3.0$ \\
\midrule
\textbf{Mean}           & \textbf{24.5} & \textbf{22.2} & \textbf{$-2.3$} \\
\bottomrule
\end{tabular}
}
\caption{Stability of half-subset evaluation against full-instance evaluation. Scores are Final Score (rubric $\times$ completeness) averaged per task then over the 6 tasks.}
\label{tab:apdx_eval_stability}
\end{table}

% Auto-generated by stats_fig_tab/apdx_gallery_eval_res/build_gallery_eval_res.py

\subsection{Evaluation Results Gallery}
\label{apdx:gallery_eval_res}

\begin{figure*}[ht]
\centering
\begin{tcolorbox}[
  title={\textsc{maze}: drive the toy car through the maze corridors and stop on the red goal cell},
  colback=GreenLight, colframe=Green,
  boxrule=0.6pt, arc=2pt, left=4pt, right=4pt, top=4pt, bottom=4pt,
  enhanced, breakable=false,
  width=0.9\textwidth, halign=center
]
\includegraphics[width=\linewidth]{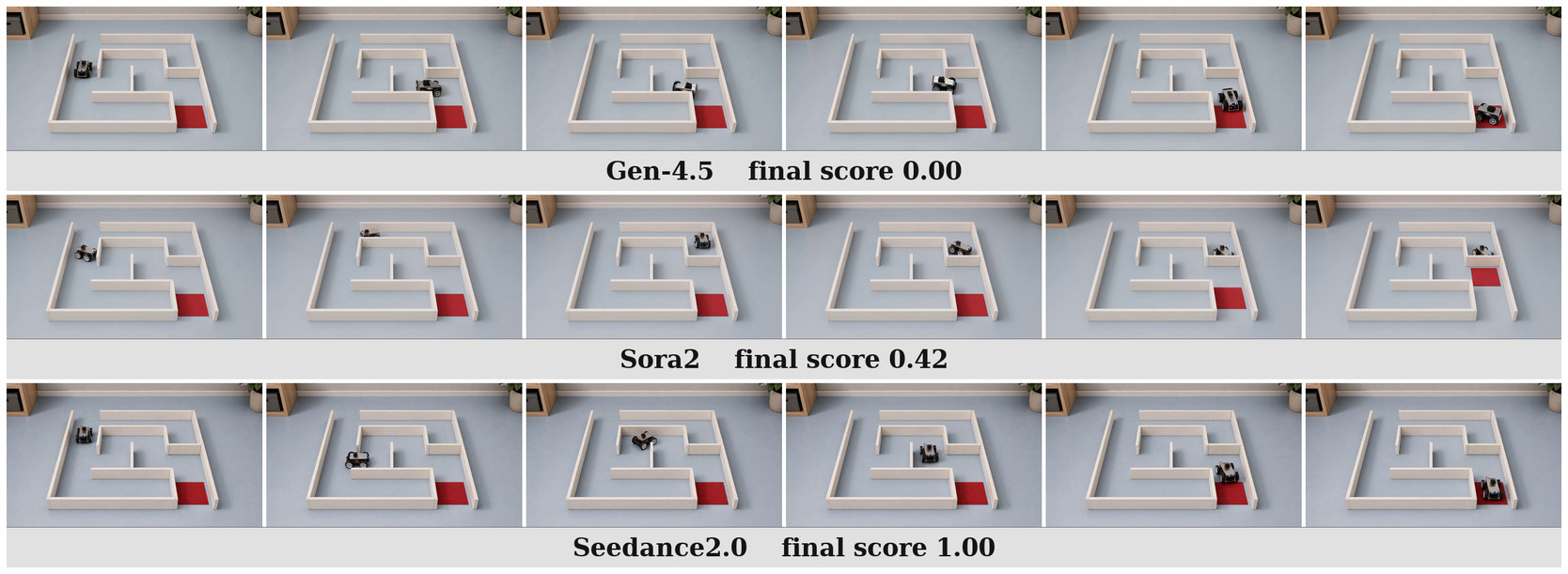}
\end{tcolorbox}
\caption{Generated result videos for the \textsc{maze} task, from three video models ordered low to high by final score.}
\label{fig:eval_gallery_maze_square}
\end{figure*}

\begin{figure*}[ht]
\centering
\begin{tcolorbox}[
  title={\textsc{recover\_2d\_net}: fold the flat 2D net up into the complete 3D solid},
  colback=GreenLight, colframe=Green,
  boxrule=0.6pt, arc=2pt, left=4pt, right=4pt, top=4pt, bottom=4pt,
  enhanced, breakable=false,
  width=0.9\textwidth, halign=center
]
\includegraphics[width=\linewidth]{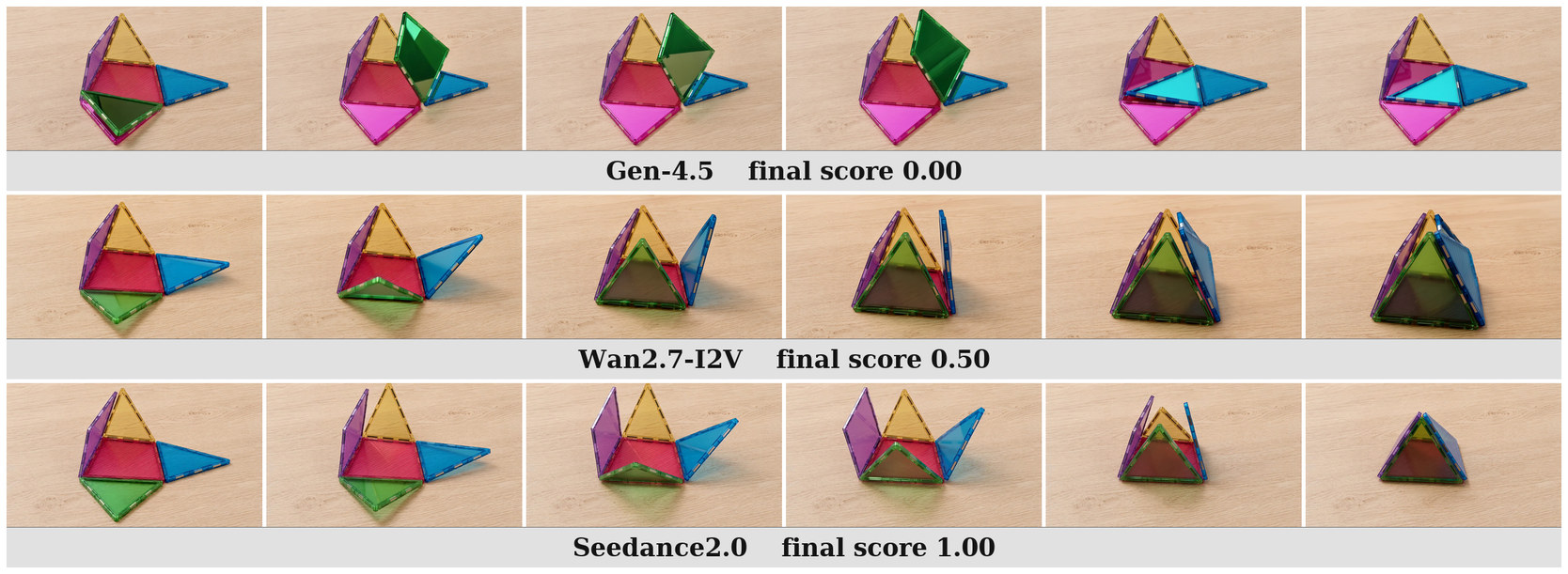}
\end{tcolorbox}
\caption{Generated result videos for the \textsc{recover\_2d\_net} task, from three video models ordered low to high by final score.}
\label{fig:eval_gallery_recover_2d_net}
\end{figure*}

\begin{figure*}[ht]
\centering
\begin{tcolorbox}[
  title={\textsc{section\_3d\_figure}: separate the translucent solid into two pieces along the dashed cross-section seam},
  colback=GreenLight, colframe=Green,
  boxrule=0.6pt, arc=2pt, left=4pt, right=4pt, top=4pt, bottom=4pt,
  enhanced, breakable=false,
  width=0.9\textwidth, halign=center
]
\includegraphics[width=\linewidth]{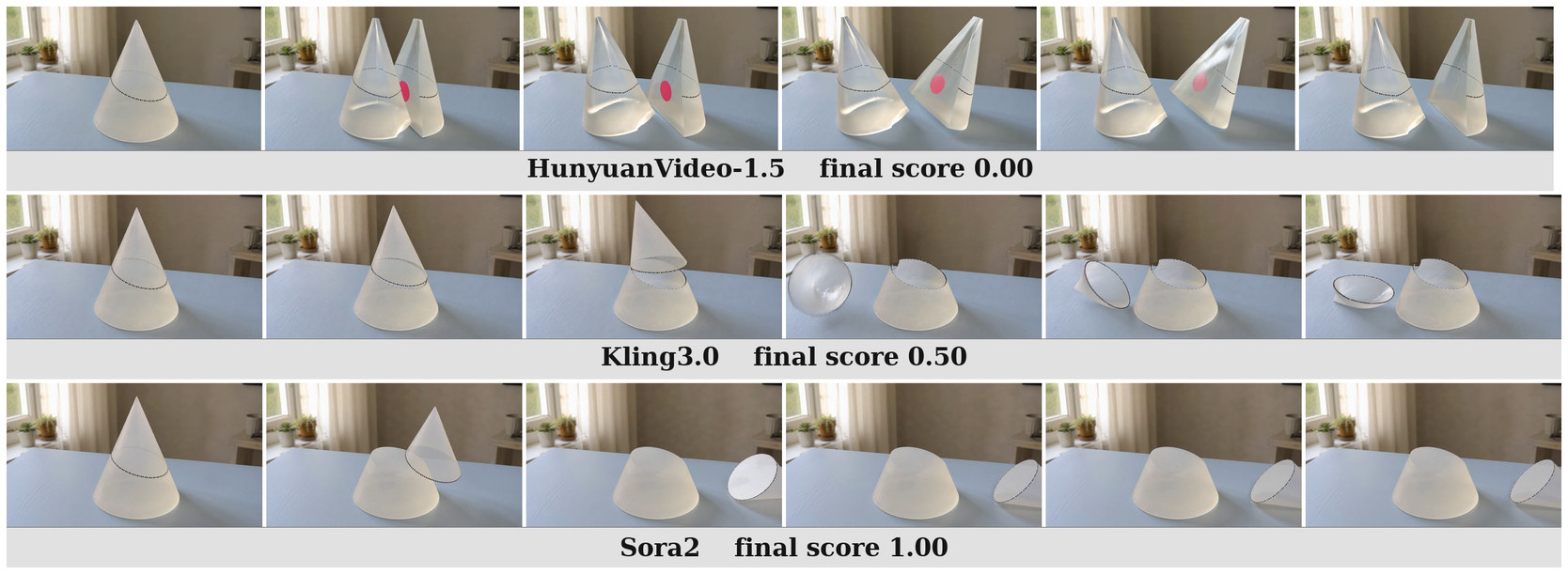}
\end{tcolorbox}
\caption{Generated result videos for the \textsc{section\_3d\_figure} task, from three video models ordered low to high by final score.}
\label{fig:eval_gallery_section_3d_figure}
\end{figure*}

\begin{figure*}[ht]
\centering
\begin{tcolorbox}[
  title={\textsc{block\_assembly}: assemble the scattered blocks into the 3D structure shown in the inset preview},
  colback=GreenLight, colframe=Green,
  boxrule=0.6pt, arc=2pt, left=4pt, right=4pt, top=4pt, bottom=4pt,
  enhanced, breakable=false,
  width=0.9\textwidth, halign=center
]
\includegraphics[width=\linewidth]{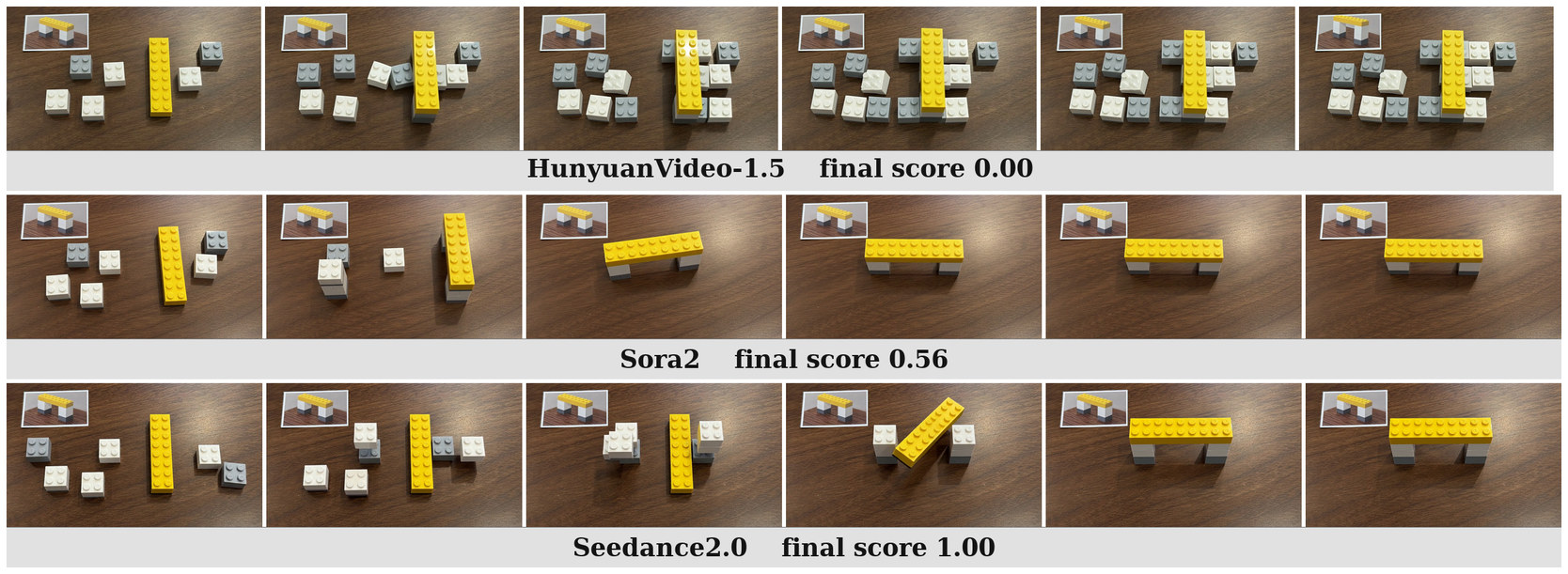}
\end{tcolorbox}
\caption{Generated result videos for the \textsc{block\_assembly} task, from three video models ordered low to high by final score.}
\label{fig:eval_gallery_block_assembly}
\end{figure*}

\begin{figure*}[ht]
\centering
\begin{tcolorbox}[
  title={\textsc{object\_packing}: pack the size- and meaning-appropriate items into the box, leaving the rest out},
  colback=GreenLight, colframe=Green,
  boxrule=0.6pt, arc=2pt, left=4pt, right=4pt, top=4pt, bottom=4pt,
  enhanced, breakable=false,
  width=0.9\textwidth, halign=center
]
\includegraphics[width=\linewidth]{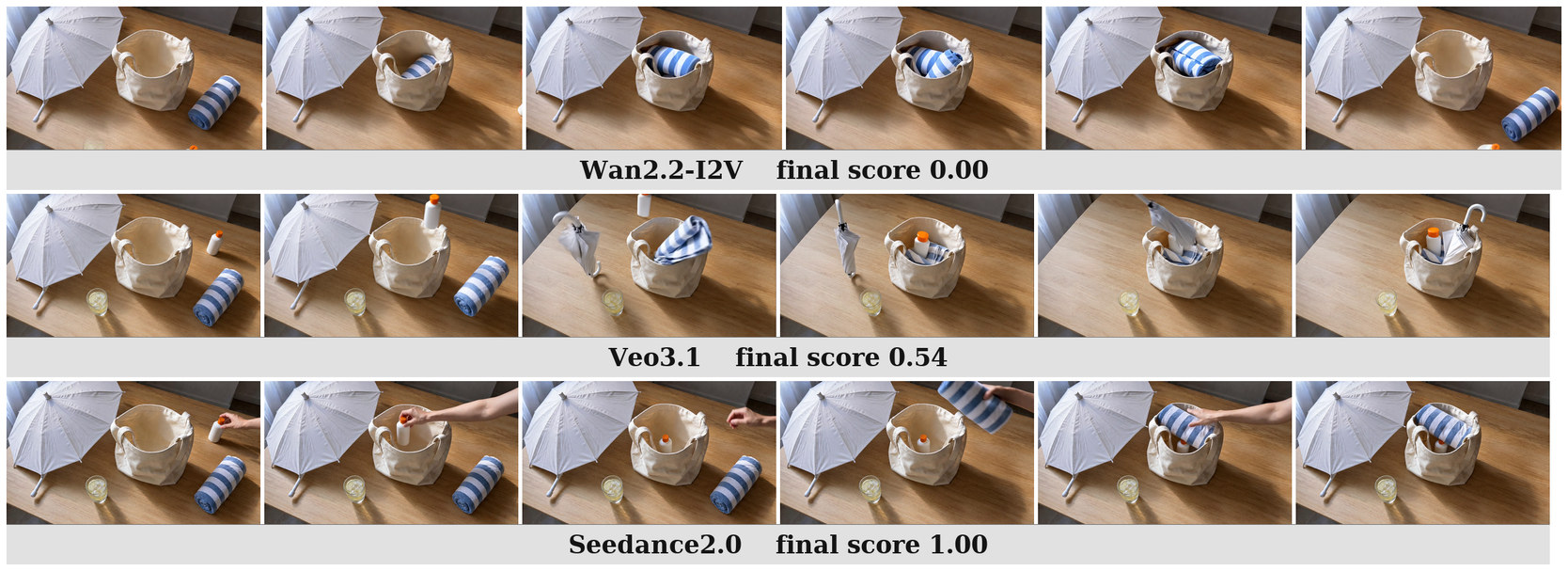}
\end{tcolorbox}
\caption{Generated result videos for the \textsc{object\_packing} task, from three video models ordered low to high by final score.}
\label{fig:eval_gallery_object_packing}
\end{figure*}

\begin{figure*}[ht]
\centering
\begin{tcolorbox}[
  title={\textsc{sorting}: arrange the pieces into the cells in order of colour darkness},
  colback=GreenLight, colframe=Green,
  boxrule=0.6pt, arc=2pt, left=4pt, right=4pt, top=4pt, bottom=4pt,
  enhanced, breakable=false,
  width=0.9\textwidth, halign=center
]
\includegraphics[width=\linewidth]{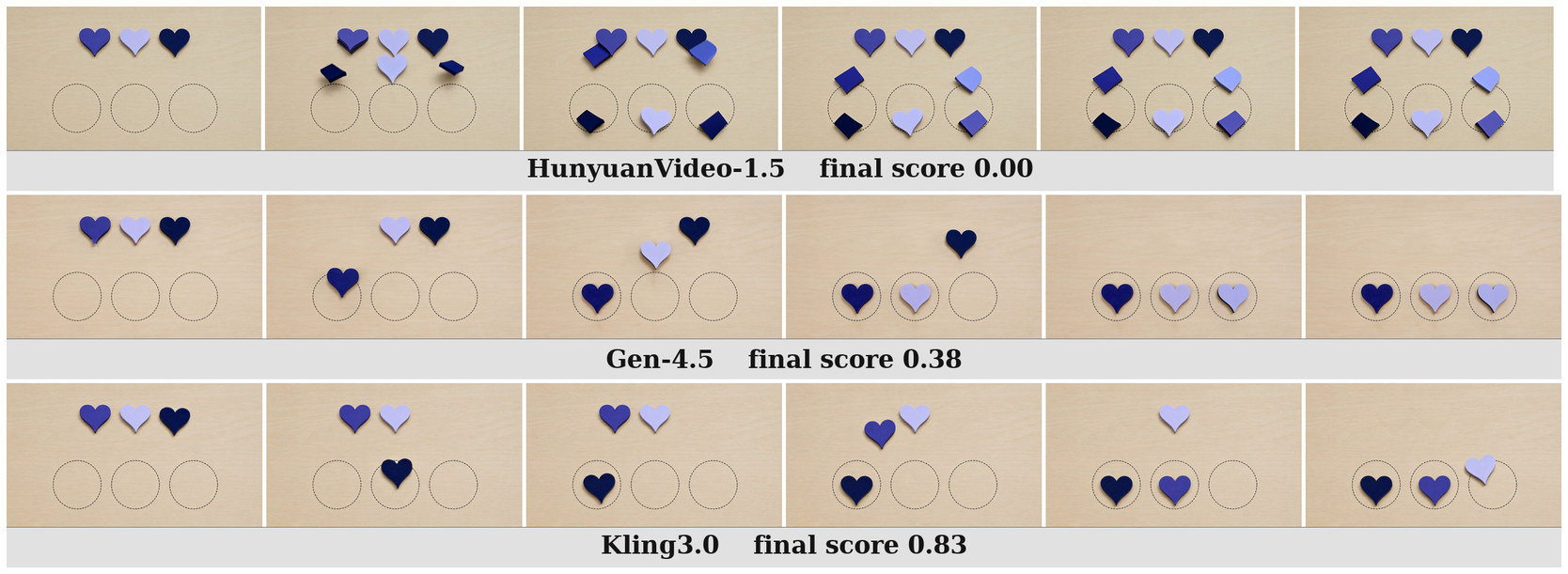}
\end{tcolorbox}
\caption{Generated result videos for the \textsc{sorting} task, from three video models ordered low to high by final score.}
\label{fig:eval_gallery_sorting}
\end{figure*}

For each task we show three result videos from different video models, ordered low to high by final score. Each row stitches six uniformly sampled frames from that model's generation; the caption bar reports the model name and its final score. Figures~\ref{fig:eval_gallery_maze_square}--\ref{fig:eval_gallery_sorting} present one such gallery per task.

\twocolumn
\clearpage
\section{VLM-as-Judge Details}
\label{apdx:judge_details}
\subsection{Implementation}
\label{apdx:vlm_as_judge}

Every generated video is scored by two complementary metrics: a global \textbf{Completeness} and a local \textbf{Rubric Score}, their product becomes the per-instance \textbf{Final Score} of Section~\ref{3_eval_criteria}. Both passes share the same underlying VLM (\texttt{gemini-3-flash-preview}) but differ in what they see and what they return; the per-pass prompts are in Tables~\ref{tab:apdx_completeness_prompt} and~\ref{tab:apdx_rubric_prompt}, with \texttt{[BRACKETS]} placeholders filled per instance; the paragraphs below expand each stage.

\paragraph{Reference routing.}
Which ground-truth signal is fed to the judge is task-dependent: some tasks supply a reference image, some a textual description, some both, and a few neither (judged on the checklist alone). Each task is routed to its available signal(s), and \texttt{[BRACKETS]} for an absent signal are simply omitted from the prompt.

% Declared here, several pages before its C.3 reference, on purpose: a
% double-column float can only be placed at a page top *after* the page it was
% declared on. Sitting next to its reference it always missed the last top and
% got flushed onto a page of its own by the \onecolumn \clearpage below.
\begin{figure*}[!t]
  \centering
  \includegraphics[width=\linewidth]{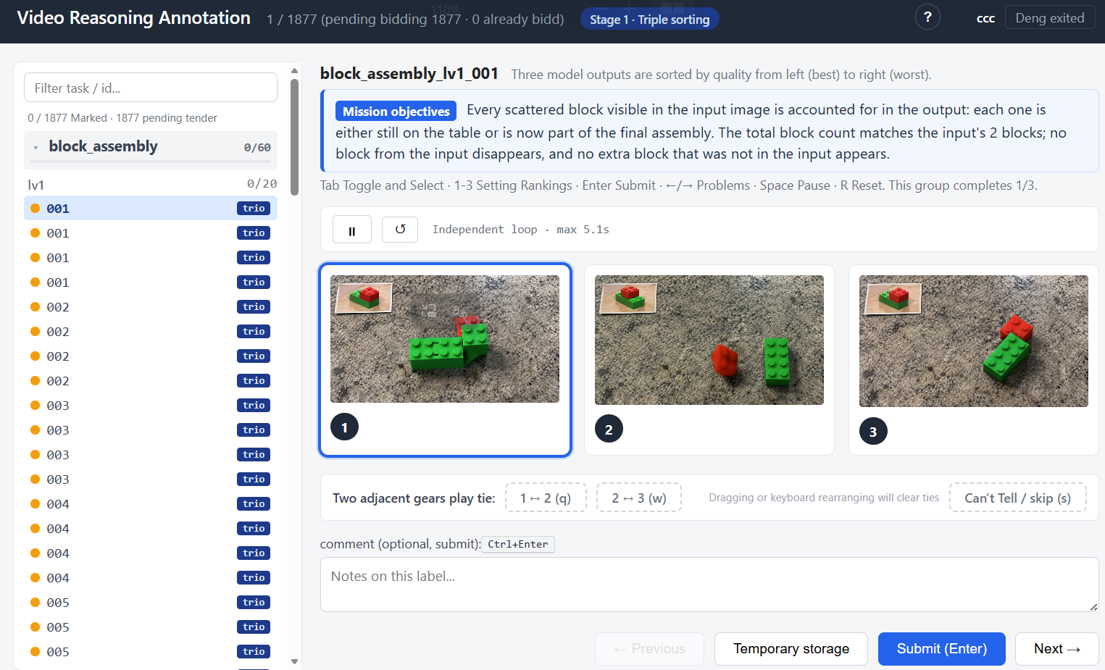}
  \caption{Screenshot of our human-annotation interface for resulting video preference. For each instance, the annotator watches the task instructions alongside three model-generated videos and either ranks them by reasoning progress from best to worst, or flags the triple as a tie / skip when no clear ordering exists.}
  \label{fig:apdx_anno_ui}
\end{figure*}

\paragraph{Completeness.}
The Completeness pass evaluates the global progress of a generated video toward the task goal. The full judge prompt example is provided in Table~\ref{tab:apdx_completeness_prompt}. 

We uniformly sample the full clip at a low frame rate (2fps) and provide the sampled frames, the input image, optional ground-truth references, and the task-specific tiered standard to the VLM judge.  The judge assigns one of three tiers: \texttt{complete}, \texttt{partial}, or \texttt{failed}, which are mapped to scores of $1$, $0.5$, and $0$, respectively. 
A \texttt{complete} judgment allows minor visual imperfections as long as the intended procedure is substantially completed; \texttt{partial} indicates meaningful progress without reaching the correct final state; and \texttt{failed} covers cases with little progress, wrong task execution, or severe drift from the input. This metric captures the macro-level trajectory of the video, while finer-grained process violations are handled by the Rubric Score below.

\paragraph{Rubric pass.}
The Rubric pass evaluates fine-grained process validity using the task-specific checklist. The full judge prompt example is provided in Table~\ref{tab:apdx_rubric_prompt}.

We use an adaptive coarse-to-fine procedure. The judge first inspects the full video at a coarse sampling rate (\emph{low fps}, $4$\,fps) with a sliding focus window of $10$ frames and a $1$-frame overlap between consecutive windows, so that each VLM call only covers a short local segment rather than the entire frame sequence. While checking the rubric items, the judge also nominates suspicious frames that require closer inspection. The time intervals around these frames are then resampled at a finer rate (\emph{high fps}, $8$\,fps) and re-evaluated with the same checklist. This design allows the evaluator to capture transient violations without densely sampling the whole video.

After all inspected segments are judged, we aggregate the segment-level comments into a violation count for each rubric item. If item $i$ is violated $x_i$ times, its item-level score is defined as $s_i = 1/(x_i+1)$. This inverse decay strongly penalizes repeated violations while giving diminishing marginal penalty when the same violation has already occurred many times. The final Rubric Score is the arithmetic mean over all checklist items:
$ \mathrm{Rub} = \frac{1}{N}\sum_{i=1}^{N} s_i. $

The per-item violation counts $x_i$ are produced by a final, image-free \emph{polish} call. This call sees no frames: it is given the full checklist together with all per-window evaluation comments from both the coarse and the refined passes, grouped by frame range and tagged with their time stamps. It deduplicates this evidence---a persistent violation spanning a window counts as a single event, and the same incident reported by two overlapping windows is merged---and returns the final $x_i$ for every checklist item (plus a short overall reasoning). Only this polish call assigns the violation counts; the per-window calls only produce localized comments and nominate frames to zoom into. The polish prompt is given in Table~\ref{tab:apdx_polish_prompt}.

\paragraph{Aggregation to Final Score and Reported Numbers.}
As discussed in Section~\ref{3_eval_criteria}, for each instance, we combine the two video-evaluation passes multiplicatively:
$\mathrm{Final}=\mathrm{Comp.}\times\mathrm{Rub.}$. 

Reported video results are averaged over all judged instances; a domain score pools every instance of the tasks belonging to that domain and averages over them, as shown in Tables~\ref{tab:main_res} and Table~\ref{tab:full_video_res}.

\subsection{Adaptation for Image Outputs}
\label{apdx:judge_image}

Image-output models produce a single still image, so we use a lightweight image judge. The judge receives the original input image, a reference image, the generated image, the task specification, and the textual ground-truth description when available. It then evaluates two criteria: \textbf{task correctness}, i.e., whether the task-relevant content satisfies the specification, and \textbf{background preservation}, i.e., whether non-task regions remain roughly consistent with the input image.
We allow minor background drift, color or layout changes, and style variation. A generated image is marked as failed when it contains task-critical errors, such as missing, wrong, unreadable, or structurally broken content. The judge returns a binary success signal. Since image outputs have no temporal axis, we do not apply Completeness or Rubric Score; the binary verdict serves as the final per-instance metric.

\subsection{Human Correlation}
\label{apdx:judge_human_corr}

To validate the reliability of our VLM-based evaluator, we collect human preference annotations over generated videos. For each selected task instance, we construct triplets consisting of three response videos from different models. Annotators are asked to rank the three videos by considering task progress, rule violations, and overall correctness. The interface (shown in Figure~\ref{fig:apdx_anno_ui}) also allows annotators to mark a tie between two videos, or skip a triplet when the three responses are indistinguishable; the skip rate is small in practice. We have four annotators and ensure that each triplet is annotated by two annotators.

We convert the triplet rankings into pairwise preferences and apply two filtering steps before computing human correlation. First, we remove pairs where the two annotators give contradictory preferences. Second, we remove pairs that would introduce preference cycles, such as $A>B>C>D>A$. After filtering and preprocessing, we obtain 1100+ human preference pairs, which serve as the human reference set to verify our evaluator.

\begin{table}[ht]
\centering
\setlength{\tabcolsep}{8pt}
\renewcommand{\arraystretch}{1.0}
\resizebox{0.8\linewidth}{!}{%
\begin{tabular}{l cc}
\toprule
\textbf{Judge Design} & \textbf{AUC} & \textbf{Pairwise Acc.} \\
\midrule
Main (w/\;  \texttt{Gemini-3-Flash})        & \textbf{0.803} & \textbf{73.2\%} \\
\midrule
\;\;w/o adaptive fps  & 0.772          & 69.5\%          \\
\;\;w/o focus window  & 0.753          & 68.3\%          \\
\midrule
\;\;w/\;  \texttt{GPT-5-mini}  & 0.690          & 64.3\%          \\
\;\;w/\;  \texttt{Claude-Haiku-4.5}  & 0.478          & 47.9\%          \\
\;\;w/\;  \texttt{Qwen3.6-Plus}  & 0.624          & 59.1\%          \\
% \;\;w/\;  \texttt{Qwen3-VL-235B-A22B}  & 0.634          & 59.0\%          \\

\bottomrule
\end{tabular}
}
\vspace{-0.4em}
\caption{
Reliability of our VLM-based evaluator compared with human annotations.
% \textbf{AUC} for the ROC-AUC over the strict-preference pairs (label $\in\{+1,-1\}$), using the raw judge-score difference as the ranking scalar.
% \textbf{Pairwise Acc.} is judge-vs-human ternary agreement (\textgreater, \textless, $=$).
}
\label{tab:judge_pairwise_auc_apdx}
\end{table}

We report two correlation metrics. The first is pairwise agreement: given a human preference pair, we use the evaluator scores of the two videos to predict which one is preferred. Since the evaluator score is normalized to $[0,1]$, we treat score differences smaller than $0.05$ as ties. The second metric is ROC-AUC over strict human preference pairs. For each pair, we compute the score difference between the two videos and measure whether this continuous signal ranks human-preferred videos higher. Unlike pairwise agreement, AUC does not depend on a tie threshold and is therefore less sensitive to score calibration.

\paragraph{Breakdown by domain and skill tag.}
Beyond the aggregate numbers, we further decompose the human-correlation pool along the two levels of our taxonomy (Section~\ref{3_taxonomy}), by routing every human preference pair to the domain and the skill tag(s) of its underlying task; since skill tags are non-exclusive, a task annotated with $k$ tags contributes its pairs to $k$ skill pools. The results are reported in Table~\ref{tab:judge_corr_breakdown_apdx}.

Agreement is stable across the taxonomy: every domain and every skill tag stays within roughly $0.73$--$0.85$ AUC and $69\%$--$80\%$ pairwise accuracy, so no single slice of the benchmark is driving the aggregate reliability. At the domain level, \textbf{Structured Puzzles} shows the strongest agreement with human preferences, which we attribute to its explicit rules, constrained solution space, and correspondingly verifiable outcomes. Agreement is lowest for \textbf{Physical Manipulation}, where success hinges on subtle details of real-world physical interaction that are more easily missed by a frame-sampling evaluator. The skill-level view echoes the same pattern: \textbf{Topology} and \textbf{Physics} are the weakest-aligned tags. Topology requires tracking fine-grained changes in connectivity and containment, while Physics depends on fine-grained interactions and dynamics spread across frames; both are harder to read off sampled frames than the discrete, checkable state changes that dominate the better-aligned tags such as \textbf{Affordance}.

% Judge-human correlation broken down by domain (top) and skill tag (bottom).
% Source: stats_fig_tab/4_judge_corr/{auc,pairwise_acc}_breakdown.txt,
% judge = main (comp x rub). Skill buckets overlap (a task with k skill
% tags feeds k buckets), so per-skill pools sum to more than the overall pool.
\providecommand{\skicon}[2]{%
  \raisebox{-0.2em}{\includegraphics[height=1.1em]{Figures/skill_icons/skill_#1.png}}\,\textbf{#2}%
}
\begin{table}[ht]
\centering
\setlength{\tabcolsep}{4pt}
\renewcommand{\arraystretch}{1.15}
\resizebox{\columnwidth}{!}{%
\begin{tabular}{l c c c c}
\toprule
\textbf{Metric}
 & \makecell{Structured\\Puzzles}
 & \makecell{Visual\\Organization}
 & \makecell{Spatiotemporal\\Dynamics}
 & \makecell{Physical\\Manipulation} \\
\midrule
\textbf{AUC}           & \textbf{0.851} & 0.800 & 0.814 & 0.783 \\
\textbf{Pairwise Acc.} & \textbf{75.0\%} & 69.2\% & 74.1\% & 70.1\% \\
\bottomrule
\end{tabular}%
}
\\[0.9em]
\resizebox{\columnwidth}{!}{%
\begin{tabular}{l c c c c}
\toprule
\textbf{Metric}
 & \skicon{planning}{Planning}
 & \skicon{spatial}{Spatial}
 & \skicon{temporal}{Temporal}
 & \skicon{attribute_grounding}{\makecell{Attribute\\Grounding}} \\
\midrule
\textbf{AUC}           & 0.815 & 0.802 & 0.819 & 0.800 \\
\textbf{Pairwise Acc.} & 73.0\% & 71.0\% & 75.6\% & 73.2\% \\
\midrule
\textbf{Metric}
 & \skicon{spring}{Physics}
 & \skicon{affordance}{Affordance}
 & \skicon{topology}{Topology}
 & \\
\midrule
\textbf{AUC}           & 0.799 & \textbf{0.842} & 0.731 & \\
\textbf{Pairwise Acc.} & 69.2\% & \textbf{80.3\%} & 69.6\% & \\
\bottomrule
\end{tabular}%
}
\vspace{-0.4em}
\caption{Judge--human agreement broken down by domain (top) and skill tag (bottom), for the main judge configuration of Table~\ref{tab:judge_pairwise_auc_apdx}. Skill tags are non-exclusive, so a task with $k$ tags contributes its pairs to $k$ skill pools.}
\label{tab:judge_corr_breakdown_apdx}
\end{table}

\onecolumn
\begin{tcolorbox}[
  enhanced, breakable, contbreak, colback=black!4, colframe=black!55,
  boxrule=0.6pt, arc=4pt,
  left=10pt, right=10pt, top=8pt, bottom=8pt,
  width=\textwidth,
  title=\textbf{Prompt: Rubric pass (one call per batch in the adaptive / sliding-window loop)},
  fonttitle=\bfseries, coltitle=white, colbacktitle=black!55
]
\scriptsize\ttfamily
\textbf{[SYSTEM]}\\[2pt]
You are a visual rubric judge for video generation outputs. You are judging a VIDEO (delivered as a small batch of frames sampled at a given fps), NOT a single image. Each frame is preceded by a metadata line giving the time in seconds and the original-video global frame index. 
\\[4pt] 
Rules:\\[4pt]
\hspace*{1em} - Reference frames by their global\_frame number when describing what you see.\\[4pt]
\hspace*{1em} - EVERY frame in this batch must appear in at least one evaluation's frame\_indices. If a frame is uneventful, group it with the nearest informative comment instead of skipping.\\[4pt]
\hspace*{1em} - Flag substantive rubric violations and ignore minor noise. Substantive means the kind of failure the rubric is actually targeting (for a maze task: the car clearly clipping through or driving over a wall, a teleport or a duplicate car, a missing required turn, etc.). IGNORE minor near-rubric deviations that are NOT the kind of failure the rubric is checking for: sub-pixel jitter, a small-part sliver of overlap during fast motion, faint compression artifacts, lighting flicker, and so on.\\[4pt]
\hspace*{1em} - DO NOT hallucinate events between adjacent sampled frames. At the current fps, consecutive frames in this batch are separated by ${\sim}1/\textit{fps}$ seconds of native video that you cannot see. If you SUSPECT something happened in that gap (two objects collided / phased, a piece teleported, the car briefly crossed a wall, etc.) but you cannot see it actually, you MUST list both flanking global\_frame numbers in "frames\_to\_inspect\_closely", so the next round samples higher fps in that gap. Do NOT write comments like ``they collide between frame X and frame Y'' or ``the object phases through between X and Y'' based on inference, that is hallucination. Only flag a collision / phase / teleport as a violation when you can point to one or more sampled frames that visibly show the overlap, the impossible state, or the object in two places at once.\\[4pt]

Each turn you are also shown the original input image and the reference ground-truth image (when available) BEFORE the sampled frames; use them as references for what the input scene contains and what one acceptable output looks like.\\[4pt]

Always reply with valid JSON matching the schema in the user message.\\[6pt]
\textbf{[USER]}\\[2pt]
Task: \textcolor{black!70}{[TASK NAME]} \\[4pt]
Task main goal: \textcolor{black!70}{[RUBRIC MAIN GOAL]} \\[4pt]
Detailed checklist (each item is scored independently; reference these 1-based indices when you flag violations):\\
\hspace*{1em} 1. \textcolor{black!70}{[CHECKLIST ITEM 1]} \\
\hspace*{1em} 2. \textcolor{black!70}{[CHECKLIST ITEM 2]} \\
\hspace*{1em} ... \\
\hspace*{1em} K. \textcolor{black!70}{[CHECKLIST ITEM K]}\\[4pt]

Ground-truth description (success / final state to reach): \# Optional \\
""" \\
\textcolor{black!70}{[GT DESCRIPTION]} \\
""" \\[4pt]

Below: the original input image then the reference ground-truth image (if available), followed by THIS BATCH's sampled video frames.\\[2pt]

Input Image: 
$\langle$INPUT IMAGE$\rangle$ \\

Ground Truth Image: 
$\langle$GT IMAGE$\rangle$ \\

Sampled Frames: \\
$\langle$FRAME 1$\rangle$  t=t\_1  global\_frame=g\_1 \\
\hspace*{1em} ... \\
$\langle$FRAME n$\rangle$  t=t\_n  global\_frame=g\_n \\[6pt]
Output STRICT JSON (no markdown, no prose outside JSON):\\
\{\\
\hspace*{1em} "evaluations": [\\
\hspace*{2em} \{"frame\_indices": [$\langle$global\_frame$\rangle$, ...], "comment": "$\langle$one or two sentences$\rangle$"\}\\
\hspace*{1em} ],\\
\hspace*{1em} "frames\_to\_inspect\_closely": [$\langle$global\_frame$\rangle$, ...]   // 0..$M$ entries; can be [] \\
\}\\[4pt]

\end{tcolorbox}
\nopagebreak[4]%
\refstepcounter{table}\label{tab:apdx_rubric_prompt}%
{\small Table~\thetable: Prompt used for the \textbf{Rubric Score} judgment. Per-batch \texttt{frames\_to\_inspect\_closely} drives the adaptive refinement loop; the localized comments it returns are later turned into violation counts by the polish call (Table~\ref{tab:apdx_polish_prompt}).\par}
\bigskip

\clearpage
\begin{tcolorbox}[
  enhanced, breakable, contbreak, colback=black!4, colframe=black!55,
  boxrule=0.6pt, arc=4pt,
  left=10pt, right=10pt, top=8pt, bottom=8pt,
  width=\textwidth,
  title=\textbf{Prompt: Completeness pass (single call per video)},
  fonttitle=\bfseries, coltitle=white, colbacktitle=black!55
]
\scriptsize\ttfamily
\textbf{[SYSTEM]}\\[2pt]
You are a visual judge for task completeness in a generated VIDEO. You are shown the INPUT image (the video's first frame / starting state), optionally a GT image and/or a GT description (one acceptable final state / goal), and the whole clip sampled as still frames in time order. Using ONLY the supplied tiered standard, decide how far the task's GOAL was actually carried out and return a single integer tier. Judge the procedure and the final state you can see across the frames; do not invent events between frames. Reply ONLY with the JSON schema given in the user message.\\[6pt]
\textbf{[USER]} 

Task: [TASK NAME]  \\
You will be shown: the INPUT image (starting state at t=0); a GT image (one acceptable final state); then the sampled video frames in time order.\\[4pt]
Tiered standard (score by how far the goal was carried out):\\
""" \\
\textcolor{black!70}{[TIERED COMPLETENESS STANDARD]} \\
""" \\[4pt]
Ground-truth description (the goal / acceptable final state):\\
""" \\
\textcolor{black!70}{[GT DESCRIPTION]} \\
""" \\[4pt]
Input Image (Start Frame): 
$\langle$INPUT IMAGE$\rangle$ \\
Ground Truth Image \# Optional
$\langle$GT IMAGE$\rangle$ \\
Sampled frames from the video: 
$\langle$FRAME 1$\rangle$  t=t\_1  global\_frame=g\_1 \\
$\langle$FRAME 2$\rangle$  t=t\_2  global\_frame=g\_2 \\
\hspace*{1em} ... \\
$\langle$FRAME N$\rangle$  t=t\_N  global\_frame=g\_N\\[4pt]
Reply ONLY with valid JSON:\\
\hspace*{1em} \{"tier": 0|1|2, "reasoning": "1-2 sentences"\}\\
tier is the single integer from the tiered standard above (0, 1, or 2).
\end{tcolorbox}
\nopagebreak[4]%
\refstepcounter{table}\label{tab:apdx_completeness_prompt}%
{\small Table~\thetable: Prompt used for the \textbf{Completeness} judgment. The tier is mapped to $\{0, 0.5, 1\}$ and becomes the first factor of \textit{Final Score}.\par}
\bigskip
% Do not let the page end here: the boxes are breakable, so forcing the next one
% to start on the current page (and split) keeps these pages full instead of
% leaving a short box alone with two thirds of the page blank.
\nopagebreak[4]

\begin{tcolorbox}[
  enhanced, breakable, contbreak, colback=black!4, colframe=black!55,
  boxrule=0.6pt, arc=4pt,
  left=10pt, right=10pt, top=8pt, bottom=8pt,
  width=\textwidth,
  title=\textbf{Prompt: Polish pass (single image-free call per video)},
  fonttitle=\bfseries, coltitle=white, colbacktitle=black!55
]
\scriptsize\ttfamily
\textbf{[SYSTEM]}\\[2pt]
You are a careful video-judging assistant. You are finalising the per-rubric-item violation counts for a VIDEO based on sampled-frame evidence collected by per-batch judges.\\[6pt]
\textbf{[USER]}\\[2pt]
You are finalising the rubric violation counts for a VIDEO. The per-batch evidence below was collected from multiple LLM calls, each shown a contiguous window of sampled frames (round 1 at low fps over the whole clip; round 2 at high fps zoomed into the frames round 1 flagged). Your job is to deduplicate that evidence into a final violation count per rubric checklist item.\\[4pt]
Task main goal:\\
\textcolor{black!70}{[RUBRIC MAIN GOAL]} \\[4pt]
Detailed checklist (reference these 1-based indices):\\
\hspace*{1em} 1. \textcolor{black!70}{[CHECKLIST ITEM 1]} \hspace*{1em} ... \hspace*{1em} K. \textcolor{black!70}{[CHECKLIST ITEM K]}\\[4pt]
Per-batch evidence (each batch saw a contiguous window at the given fps; consecutive batches share their boundary frame as overlap):\\
\hspace*{1em} - round $r$ fps=$f$ batch $b$ (global $g_{lo}$-$g_{hi}$, t $\dots$s):\\
\hspace*{2em} comments:\\
\hspace*{3em} - global\_frames $\langle$a-b$\rangle$: \textcolor{black!70}{[COMMENT]} \\
\hspace*{2em} ... \hspace*{1em} (repeated for every batch of every round) \\[4pt]
Tolerance: only count something as a violation if it is defensible to a human reviewer as a real failure; ignore minor near-rubric noise (sub-pixel jitter, a one-pixel sliver during fast motion, lighting flicker, micro-drift). A borderline / cosmetic deviation does NOT count.\\[4pt]
Output STRICT JSON only:\\
\{\\
\hspace*{1em} "by\_range": [\{"global\_frames": "$\langle$a-b$\rangle$", "summary": "$\langle$one sentence$\rangle$"\}, ...],\\
\hspace*{1em} "checklist": [\\
\hspace*{2em} \{"index": $\langle$1..K$\rangle$, "n\_violations": $\langle$int $\geq$ 0$\rangle$, "evidence": "$\langle$one sentence citing time ranges$\rangle$"\}\\
\hspace*{2em} // EXACTLY K entries, indexes 1..K in order \\
\hspace*{1em} ],\\
\hspace*{1em} "overall\_reasoning": "$\langle$two or three sentences$\rangle$"\\
\}\\[4pt]
Per-item rule: \texttt{n\_violations} is the FINAL deduplicated count of DISTINCT substantive violations of that item across the whole video, after integrating evidence from ALL batches (round 1 + round 2). Deduplication: (a) a persistent violation spanning a continuous window counts as ONE; (b) two reports in adjacent batches within $0.5$\,s across the boundary count as ONE; (c) the same incident in two overlapping batches is ONE. Per-item score $= 1/(\texttt{n\_violations}+1)$; \texttt{rubric\_score} is the arithmetic mean over items.
\end{tcolorbox}
\nopagebreak[4]%
\refstepcounter{table}\label{tab:apdx_polish_prompt}%
{\small Table~\thetable: Prompt used for the polishing and summarization of Rubric Score per-batch judge results.\par}
\bigskip

\begin{tcolorbox}[
  enhanced, breakable, contbreak, colback=black!4, colframe=black!55,
  boxrule=0.6pt, arc=4pt,
  left=10pt, right=10pt, top=8pt, bottom=8pt,
  width=\textwidth,
  title=\textbf{Prompt: Image-output judge (single call per generated image)},
  fonttitle=\bfseries, coltitle=white, colbacktitle=black!55
]
\scriptsize\ttfamily
\textbf{[SYSTEM]}\\[2pt]
You are a visual judge for image-generation outputs. The model produces a SINGLE generated image; judge only that still image.\\[4pt]
You will be shown, in order: the original INPUT image, optionally a reference GROUND-TRUTH image showing one acceptable answer, and the MODEL's generated image. A textual ground-truth description may also appear in the prompt.\\[4pt]
Success requires BOTH criteria:\\
\hspace*{1em} (1) \textbf{Task correctness} — the task-relevant content satisfies the spec. The GT image or JSON answer gives one acceptable solution; semantically equivalent alternatives also count.\\
\hspace*{1em} (2) \textbf{Background preservation} — outside the task-relevant region, the generated image should remain consistent with the INPUT: same scene, objects, layout, framing, and major colors. Substantial changes such as adding/removing objects, recoloring or distorting the background, rebuilding the scene, or changing style/medium count as failures. Background is judged against INPUT, not GT.\\[4pt]
OVERALL RULE: focus on key task-critical content. Mark success=true if the key content is correct despite minor issues, such as small background drift, slight non-task repositioning, minor color/lighting changes, small rendering artifacts, or light style shifts. Fail only if the key content is wrong, missing, unreadable, or structurally broken, or if the background is substantively rebuilt/replaced/restyled. If no INPUT is provided, judge only task correctness.\\[4pt]
Reply ONLY with valid JSON in the schema specified in the user message.\\[6pt]
\textbf{[USER]}\\[2pt]
Task: [TASK NAME] \\[4pt]
Task spec (what the model was asked to generate):\\
""" \\
\textcolor{black!70}{[IMAGE PROMPT]} \\
""" \\[4pt]
You will see N image(s), in this order:\\
\hspace*{1em} 1. INPUT (the image the model was given)\\
\hspace*{1em} 2. GROUND-TRUTH image (one acceptable answer)\\
\hspace*{1em} 3. MODEL's GENERATED image (to be judged)\\[4pt]
Ground-truth description (success / final state to reach):\\
\textcolor{black!70}{[GT DESCRIPTION]}\\[4pt]
Input Image: 
$\langle$INPUT IMAGE$\rangle$ \\
Ground Truth Image: 
$\langle$GT IMAGE$\rangle$ \\
Generated Image: 
$\langle$GENERATED IMAGE$\rangle$\\[4pt]
Decide if the generated image substantively satisfies the task spec. Both must hold for success: (1) task correctness and (2) background preservation (skip if no INPUT). Apply the key-thing rule: ignore minor secondary issues; only fail on substantive task-critical errors or wholesale background rebuild.\\[4pt]
Reply ONLY with valid JSON matching this schema:\\
\hspace*{1em} \{"success": true|false, "reasoning": "2--3 sentences explaining the verdict"\}
\end{tcolorbox}
\nopagebreak[4]%
\refstepcounter{table}\label{tab:apdx_image_judge_prompt}%
{\small Table~\thetable: Prompt used for the \textbf{Image-output judge}. Mirrors the video-judge two-criteria structure (task correctness $\wedge$ background preservation), reduced to a single still-image call: the model output is one image, so no adaptive frame sampling or sliding-window loop is needed.\par}
\bigskip
\clearpage
\onecolumn
\section{Details of Discussion and Analysis}
\label{apdx:discussion_details}
\subsection{Failure Modes: More Examples}
\label{apdx:failure_modes}
% Per-mode palette uses the global predefined colors:
%   physical collapse     -> Green  / GreenLight
%   rule violation        -> Yellow / YellowLight
%   object/state inconsist -> Blue   / BlueLight

\medskip

\newcommand{\fmcell}[2]{%
  \begin{minipage}[t]{0.49\linewidth}
    \includegraphics[width=\linewidth]{#1}
    \par\smallskip
    #2
  \end{minipage}%
}

\begin{center}
\scalebox{0.98}{%
\begin{minipage}{\linewidth}
\centering

% ---------------- Mode 1: Physical collapse ----------------
\begin{tcolorbox}[
  enhanced, boxrule=0.9pt, colframe=Green, colback=GreenLight,
  title=\textbf{Failure Mode: Physical Collapse}, coltitle=white,
  fonttitle=\bfseries, colbacktitle=Green, fontupper=\footnotesize,
  arc=4pt, left=3pt, right=3pt, top=2pt, bottom=2pt
]
  \fmcell{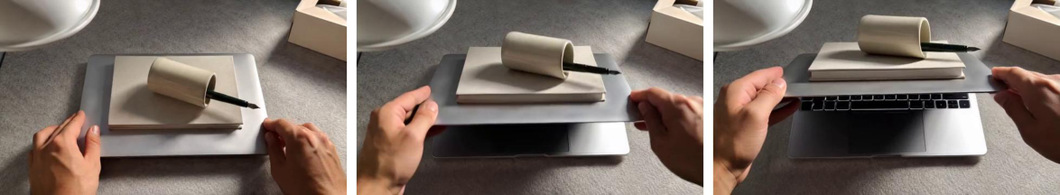}{The laptop is tilted up at a noticeable angle, yet \textbf{the cup lying on its lid does not roll down}, appearing \textbf{glued to the surface} and violating gravity.}\hfill
  \fmcell{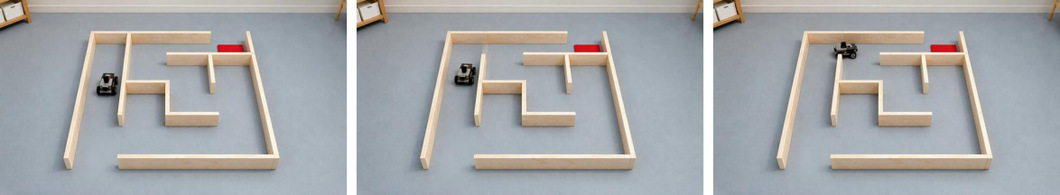}{The maze layout abruptly changes mid-video: a \textbf{previously solid wall opens up into a gap}.}
  \par\smallskip
  \fmcell{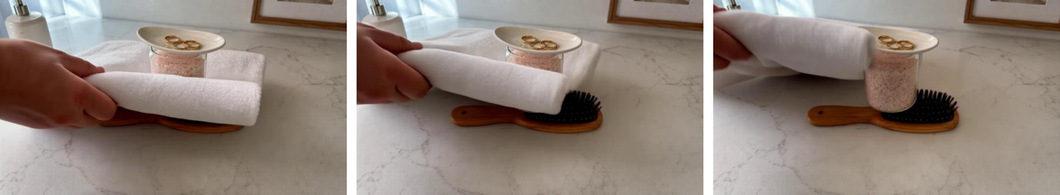}{The towel \textbf{passes straight through the bottle}; in the intermediate frame the towel and the bottle are visibly \textbf{entangled rather than colliding}.}\hfill
  \fmcell{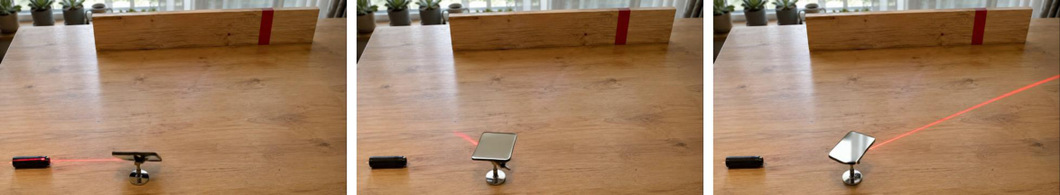}{When the laser hits the mirror, \textbf{the incident ray is sometimes missing} and \textbf{the reflected ray is sometimes missing}, a clear violation of optical physics.}
\end{tcolorbox}

\medskip
% ---------------- Mode 2: Rule violation ----------------
\begin{tcolorbox}[
  enhanced, boxrule=0.9pt, colframe=Yellow, colback=YellowLight,
  title=\textbf{Failure Mode: Rule Violation}, coltitle=white,
  fonttitle=\bfseries, colbacktitle=Yellow, fontupper=\footnotesize,
  arc=4pt, left=3pt, right=3pt, top=2pt, bottom=2pt
]
  \fmcell{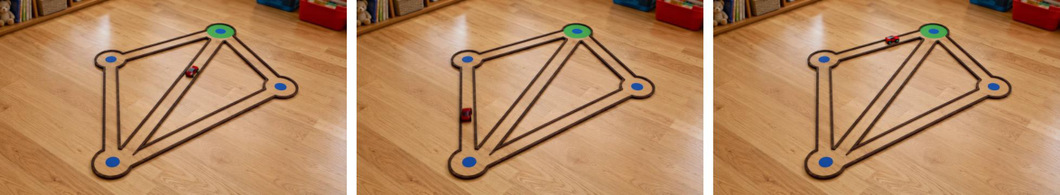}{An Eulerian path requires traversing every edge exactly once, but the trajectory in the video \textbf{terminates without covering all edges}.}\hfill
  \fmcell{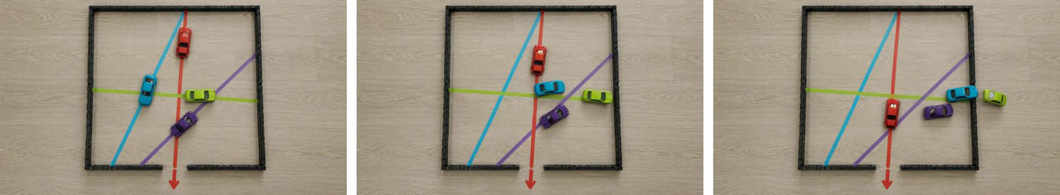}{The \textsc{leave\_parking\_lot} task imposes two rules: (i) each car may only move along the track segment of its own color, and (ii) no car may cross over the walls. \textbf{The video violates both}.}
  \par\smallskip
  \fmcell{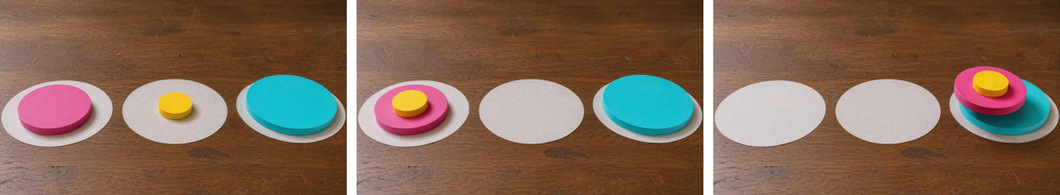}{The Tower of Hanoi allows only one disk to be moved at a time, yet \textbf{the video moves two disks simultaneously}.}\hfill
  \fmcell{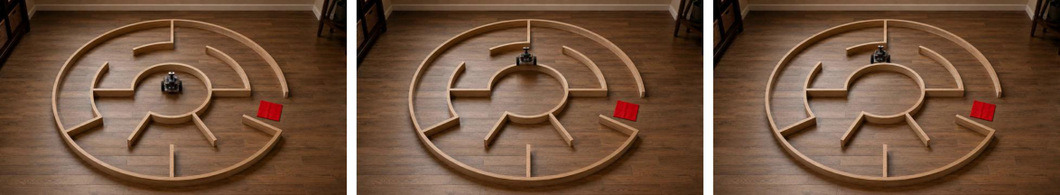}{Cars navigating the maze are not allowed to jump over the walls, yet \textbf{the car in the video does exactly that}.}
\end{tcolorbox}

\medskip
% ---------------- Mode 3: Object/state inconsistency ----------------
\begin{tcolorbox}[
  enhanced, boxrule=0.9pt, colframe=Blue, colback=BlueLight,
  title=\textbf{Failure Mode: Object/State Inconsistency}, coltitle=white,
  fonttitle=\bfseries, colbacktitle=Blue, fontupper=\footnotesize,
  arc=4pt, left=3pt, right=3pt, top=2pt, bottom=2pt
]
  \fmcell{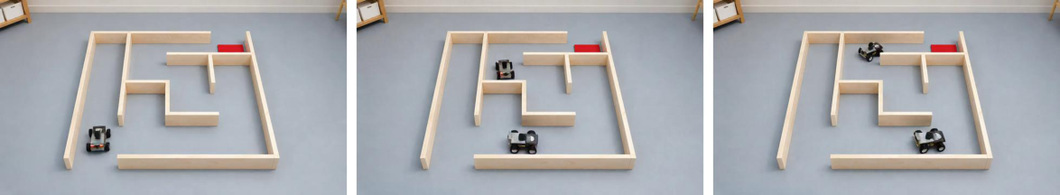}{While the car is navigating through the maze, \textbf{it suddenly splits into two cars}.}\hfill
  \fmcell{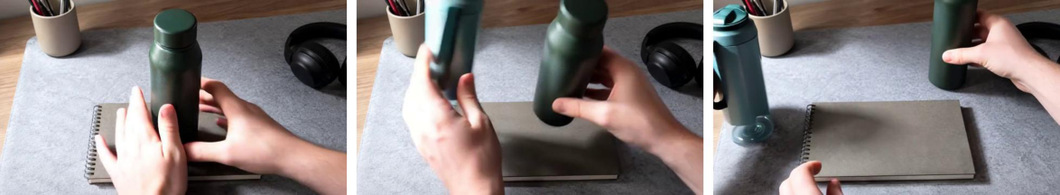}{As the cup is being picked up, \textbf{a second identical cup splits off from it}.}
  \par\smallskip
  \fmcell{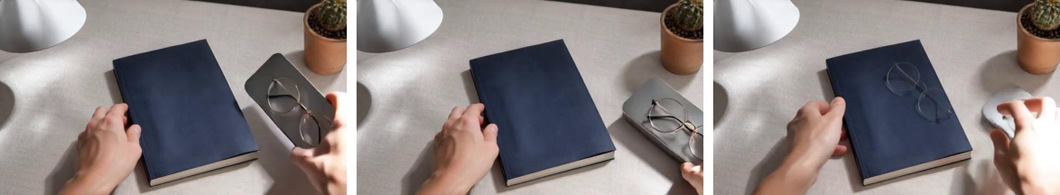}{The glasses have just been moved to the right side, yet in the very next moment \textbf{they reappear on the book in the middle}.}\hfill
  \fmcell{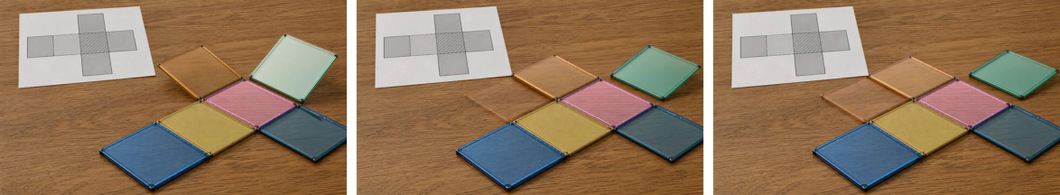}{After the acrylic panels finish unfolding, \textbf{an extra panel suddenly appears} that was not present before.}
\end{tcolorbox}

\end{minipage}}
\par\medskip
\captionof{figure}{Representative failure cases for the three failure modes: \textbf{Physical Collapse}, \textbf{Rule Violation}, and \textbf{Object/State Inconsistency}.}
\label{fig:fail_modes}
\end{center}
\twocolumn
\newpage

\subsection{Input Condition Sensitivity}
\label{apdx:sens_analysis}

\subsubsection{Oracle Prompting.}
Prompt optimization is a common test-time strategy for improving video generation quality and controllability, and recent studies have explored similar directions ~\citep{tong2025thinking, chen2025tivibench, 
cheng2026vlmsgoodteachersvideo, kim2026collabvrcollaborativevideoreasoning}. VideoThinkBench~\citep{tong2025thinking} studies how prompt rewriting and related prompting strategies affect video-based reasoning, while TiVi-Bench~\citep{chen2025tivibench} introduces VideoTPO as a training-free test-time optimization method for improving reasoning performance. Motivated by these works, we examine the performance ceiling when the intended solution is made explicit in the prompt.

We construct an \textbf{oracle prompt} for a subset of 7 tasks by augmenting the canonical task rules with a ground-truth step-by-step solution. Each prompt contains (i) a concrete, executable solution, such as an exact move sequence or target assignment, followed by (ii) the original task rules. This reduces the reasoning burden and tests whether the model can follow detailed instructions and render a valid visual process under all constraints. In the examples below, we show the solution explicitly and abbreviate the retained content as \texttt{[original task goal and rules]}.

We focus on open-source models and commercial systems that expose prompt enhancement as a configurable option. Since many proprietary systems may internally rewrite user prompts, isolating the effect of oracle prompting is difficult. We therefore conduct the controlled comparison using \texttt{\textbf{HunyuanVideo-1.5}}, \texttt{\textbf{Veo 3.1}}, and \texttt{\textbf{Wan 2.7}}.

\begin{tcolorbox}[
  title=Oracle Prompt: \textsc{sorting} example 1,
  colback=GreenLight,
  colframe=Green,
  breakable, contbreak
]
\linespread{1.18}\scriptsize\selectfont
Move every top-row piece one at a time, in darkness order (darkest first, lightest last), into a bottom-row cell so the bottom row reads darkest $\rightarrow$ lightest left to right. Per-instance assignment: \\
\quad step 1: top-row piece at position 1 (rank-1 darkest) $\rightarrow$ slot 1 \\
\quad step 2: top-row piece at position 3 (rank-2 darkest) $\rightarrow$ slot 2 \\
\quad step 3: top-row piece at position 2 (rank-3 darkest) $\rightarrow$ slot 3 \\

Each move is one continuous human-hand pick-and-place from the top-row position to the target slot. Once placed, a piece does not move again; cells themselves never move. Piece colours, shapes, sizes unchanged. Exactly one bare human hand operates the pieces; no tools or other agents enter the frame. Keep background, lighting, and camera fixed. No glitches.

\medskip
\texttt{[original task rules]}
\end{tcolorbox}

\begin{tcolorbox}[
  title=Oracle Prompt: \textsc{sorting} example 2,
  colback=GreenLight,
  colframe=Green,
  breakable, contbreak
]
\linespread{1.18}\scriptsize\selectfont
Move every top-row piece one at a time, in darkness order (darkest first, lightest last), into a bottom-row cell so the bottom row reads darkest $\rightarrow$ lightest left to right. Per-instance assignment: \\
\quad step 1: top-row piece at position 5 (rank-1 darkest) $\rightarrow$ slot 1 \\
\quad step 2: top-row piece at position 2 (rank-2 darkest) $\rightarrow$ slot 2 \\
\quad step 3: top-row piece at position 3 (rank-3 darkest) $\rightarrow$ slot 3 \\
\quad step 4: top-row piece at position 4 (rank-4 darkest) $\rightarrow$ slot 4 \\
\quad step 5: top-row piece at position 1 (rank-5 darkest) $\rightarrow$ slot 5 \\

Each move is one continuous human-hand pick-and-place from the top-row position to the target slot. Once placed, a piece does not move again; cells themselves never move. Piece colours, shapes, sizes unchanged. Exactly one bare human hand operates the pieces; no tools or other agents enter the frame. Keep background, lighting, and camera fixed. No glitches.

\medskip
\texttt{[original task rules]}
\end{tcolorbox}

\begin{tcolorbox}[
  title=Oracle Prompt: \textsc{maze} example 1,
  colback=GreenLight,
  colframe=Green,
  breakable, contbreak
]
\linespread{1.18}\scriptsize\selectfont
The blue dot (or toy car) drives smoothly through the 4x4 maze corridors from its starting cell to the centre of the red goal cell, taking exactly 6 single-cell steps. Each step moves to the adjacent corridor cell in the named direction: \\
\quad down, down, right, right, right, down

Stay strictly inside corridors; never cross, climb, or pass through any wall. Motion is continuous from start to goal: no teleport, no jumps, no airborne motion. Maze layout, walls, dot/car, and goal square are pixel-identical to the input.

Use exactly the input scene; do not add, remove, recolour, or resize any object beyond what the solution above moves. No hands or external agents enter the frame; motion is autonomous. Keep background, lighting, and camera fixed. No glitches or artifacts.

\medskip
\texttt{[original task rules]}
\end{tcolorbox}

\begin{tcolorbox}[
  title=Oracle Prompt: \textsc{maze} example 2,
  colback=GreenLight,
  colframe=Green,
  breakable, contbreak
]
\linespread{1.18}\scriptsize\selectfont
The blue dot (or toy car) drives smoothly through the 5x5 maze corridors from its starting cell to the centre of the red goal cell, taking exactly 10 single-cell steps. Each step moves to the adjacent corridor cell in the named direction: \\
\quad right, down, left, down, down, down, right, right, right, right

Stay strictly inside corridors; never cross, climb, or pass through any wall. Motion is continuous from start to goal: no teleport, no jumps, no airborne motion. Maze layout, walls, dot/car, and goal square are pixel-identical to the input.

Use exactly the input scene; do not add, remove, recolour, or resize any object beyond what the solution above moves. No hands or external agents enter the frame; motion is autonomous. Keep background, lighting, and camera fixed. No glitches or artifacts.

\medskip
\texttt{[original task rules]}
\end{tcolorbox}

\begin{tcolorbox}[
  title=Oracle Prompt: \textsc{rope\_untangle} example 1,
  colback=GreenLight,
  colframe=Green,
  breakable, contbreak
]
\linespread{1.18}\scriptsize\selectfont
Two human hands enter from the sides and cooperatively untangle the cables shown in the input. By the final frame every cable lies in its own region as a single roughly-straight line with zero crossings: no cable passes over, under, or through another. Resolve each crossing by lifting one cable up and over the other (visible 3D motion), never by phasing through. Loops shrink and open as cables slide free. Same cable count, same individual colours, same connectors at both ends of every cable as the input; nothing is added, removed, recoloured, cut, spliced, or merged.

Two normal human hands only (five fingers each, natural look, no morphing, visible contact). Keep the floor / table background, lighting, and top-down perspective fixed. No extra cables or objects. No glitches.

\medskip
\texttt{[original task rules]}
\end{tcolorbox}

\subsubsection{Visual Style.}
\begin{figure*}[!t]
    \centering
    \begin{tcolorbox}[
      colback=RedLight,
      colframe=Red,
      breakable,
      left=2pt, right=2pt, top=2pt, bottom=2pt
    ]
    \begin{subfigure}{\linewidth}
        \centering
        \includegraphics[width=\linewidth]{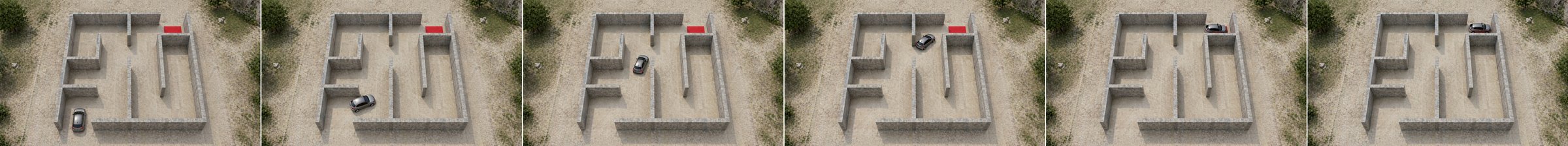}
        \caption{Real-scene: succeeds.}
        \label{fig:style_mismatch_maze_real}
    \end{subfigure}
    \\[0.5em]
    \begin{subfigure}{\linewidth}
        \centering
        \includegraphics[width=\linewidth]{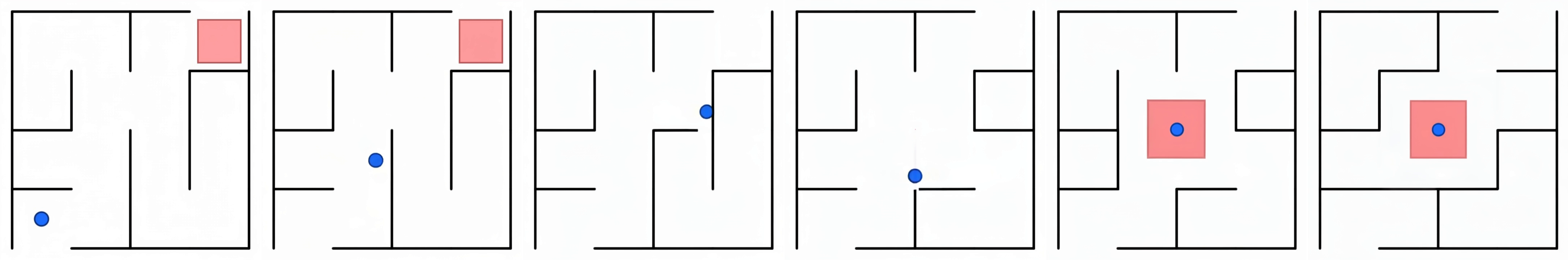}
        \caption{Line-art: fails and the maze structure is not preserved, with prompt explicitly instructing the model to preserve the maze structure and follow the maze rules.}
        \label{fig:style_mismatch_maze_line}
    \end{subfigure}
    \end{tcolorbox}
    \caption{Effect of appearance and style difference. Example task: \textsc{maze}.}
    \label{fig:style_mismatch_eg1_maze}
\end{figure*}

\begin{figure*}[!t]
    \centering
    \begin{tcolorbox}[
      colback=RedLight,
      colframe=Red,
      breakable,
      left=2pt, right=2pt, top=2pt, bottom=2pt
    ]
    \begin{subfigure}{\linewidth}
        \centering
        \includegraphics[width=\linewidth]{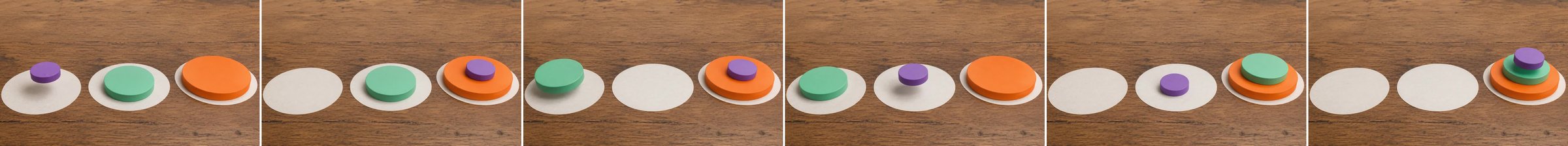}
        \caption{Real-scene: succeeds.}
        \label{fig:style_mismatch_hanoi_tower_real}
    \end{subfigure}
    \\[0.5em]
    \begin{subfigure}{\linewidth}
        \centering
        \includegraphics[width=\linewidth]{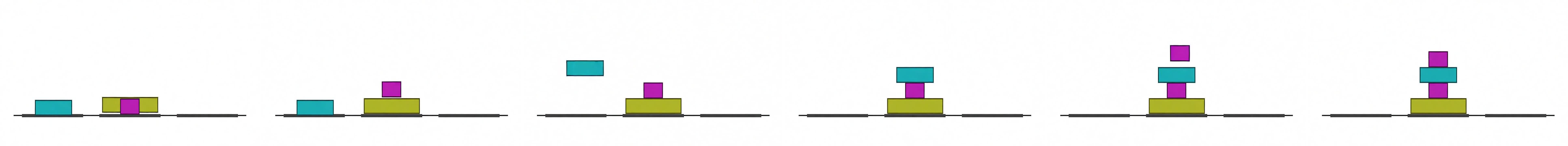}
        \caption{Line-art: fails and rules are not followed.}
        \label{fig:style_mismatch_hanoi_tower_line}
    \end{subfigure}
    \end{tcolorbox}
    \caption{Effect of appearance and style difference. Example task: \textsc{hanoi tower}.}
    \label{fig:style_mismatch_eg2_hanoi_tower}
\end{figure*}

We evaluate the effect of visual style on 5 video models: \texttt{\textbf{Sora2}}, \texttt{\textbf{Veo3.1}}, \texttt{\textbf{Kling3.0}}, \texttt{\textbf{Wan2.2}}, \texttt{\textbf{HunyuanVideo-1.5}}, and 7 selected tasks: \textsc{clock\_running}, \textsc{hanoi\_tower}, \textsc{jigsaw\_puzzle}, \textsc{leave\_parking\_lot}, \textsc{maze}, \textsc{pick\_unique}, and \textsc{polyform\_tiling}. Below We show some generated videos of the same task in photorealistic style and line-art style, including \textsc{maze} (Figure~\ref{fig:style_mismatch_eg1_maze}) and \textsc{hanoi tower} (Figure~\ref{fig:style_mismatch_eg2_hanoi_tower}).

% Declared here, next to the D.2.2 text that references them, rather than at the
% end of the file behind a \newpage: there they were flushed onto a float page
% of their own (spread apart by the float-page glue) three pages after the
% reference, and left the preceding paragraph alone on a near-empty page.

\subsection{Scaling Fine-tuning on Synthetic Data}
\label{apdx:vbvr_sft}

Complementing Section~\ref{5_sft_transfer}, we provide further details on how our \bench \ tasks are grouped according to their structural overlap with the VBVR training set distribution~\citep{vbvr2026}. We identify \textbf{overlap} tasks as those whose goals and overall task structures closely match a VBVR training task, with the main difference being the visual domain: VBVR uses abstract synthetic scenes, whereas our benchmark uses realistic inputs. Representative examples include \textsc{maze} and \textsc{clock\_running}. \textbf{Semi-overlap} tasks share only part of the task structure or require action patterns similar to those seen during VBVR training. \textbf{Non-overlap} tasks have no clear structural counterpart in VBVR; this group includes, for example, tasks requiring 3D imagination and understanding, capabilities largely absent from its training distribution. The qualitative examples below use the \texttt{VBVR-Wan2.2} / \texttt{Wan2.2-I2V} pair.

The following examples illustrate these three groups. For overlap and semi-overlap tasks, each pair places our real-scene input on the left and the most related VBVR training input on the right, each labelled with its own task name; only our input is shown for non-overlap tasks. Each example is annotated with its task-level final score averaged across difficulty levels. The value before the arrow corresponds to the base \texttt{Wan2.2-I2V}, the value after it to \texttt{VBVR-Wan2.2}, and the highlighted delta indicates a gain in \textcolor{Green}{green} or a drop in \textcolor{Red}{red}.

% One (ours, VBVR) example pair in a colored box. Each image carries its own
% task name underneath -- ours on the left, the VBVR training task on the right
% -- so the reader can see which VBVR task the pair is claiming to match.
% [#1 extra tcolorbox options (e.g. group title)]
% #2 frame color  #3 back color  #4 image stem
% #5 our task name  #6 score annotation  #7 VBVR task name
\newcommand{\vbvrPairBox}[7][]{%
  \begin{tcolorbox}[
    enhanced, boxrule=0.9pt, colframe=#2, colback=#3,
    arc=4pt, left=3pt, right=3pt, top=3pt, bottom=3pt,
    coltitle=white, colbacktitle=#2, fonttitle=\bfseries\small, #1
  ]
    \centering
    \begin{minipage}[c]{0.615\linewidth}\centering
      \includegraphics[width=\linewidth]{Figures/apdx_vbvr_overlap/ours_#4.jpg}%
    \end{minipage}\hfill
    \begin{minipage}[c]{0.345\linewidth}\centering
      \includegraphics[width=\linewidth]{Figures/apdx_vbvr_overlap/vbvr_#4.png}%
    \end{minipage}\\[3pt]
    \begin{minipage}[t]{0.615\linewidth}\centering
      {\scriptsize\textit{ours:}\ #5\\[1pt] #6}
    \end{minipage}\hfill
    \begin{minipage}[t]{0.345\linewidth}\centering
      {\scriptsize\textit{VBVR:}\ #7}
    \end{minipage}
  \end{tcolorbox}%
}
% A VBVR training-task name. Deliberately NOT \textsc: small caps are reserved
% for our own task names, so a VBVR task is set as a plain identifier instead.
% The names are long and underscore-separated, so the call sites put an
% \allowbreak after every underscore rather than letting them run past the column.
\newcommand{\vbvrtask}[1]{\texttt{#1}}
% One ours-only example cell (used by the non-overlap 2x2 grid).
% #1 image stem  #2 task name  #3 score annotation
\newcommand{\vbvrSoloCell}[3]{%
  \begin{minipage}[t]{0.48\linewidth}\centering
    \includegraphics[width=\linewidth]{Figures/apdx_vbvr_overlap/ours_#1.jpg}\\[1pt]
    {\scriptsize #2\quad #3}
  \end{minipage}%
}
% Score annotation: base -> SFT with a color-backed delta.
% #1 base score  #2 SFT score  #3 delta
\newcommand{\vbvrUp}[3]{$#1$\,{\tiny(base)}\,$\rightarrow$\,$#2$\,{\tiny(SFT)}\;{\setlength{\fboxsep}{1.5pt}\colorbox{green!22}{\textcolor{Green}{$\uparrow$#3}}}}
\newcommand{\vbvrDn}[3]{$#1$\,{\tiny(base)}\,$\rightarrow$\,$#2$\,{\tiny(SFT)}\;{\setlength{\fboxsep}{1.5pt}\colorbox{red!22}{\textcolor{Red}{$\downarrow$#3}}}}

% Boxes are placed inline (no float) so they stay adjacent to the text and
% can break between boxes across columns/pages; the first box of each group
% carries the group name on its title bar.

\vbvrPairBox[title=\textbf{Overlap}]{Green}{GreenLight}{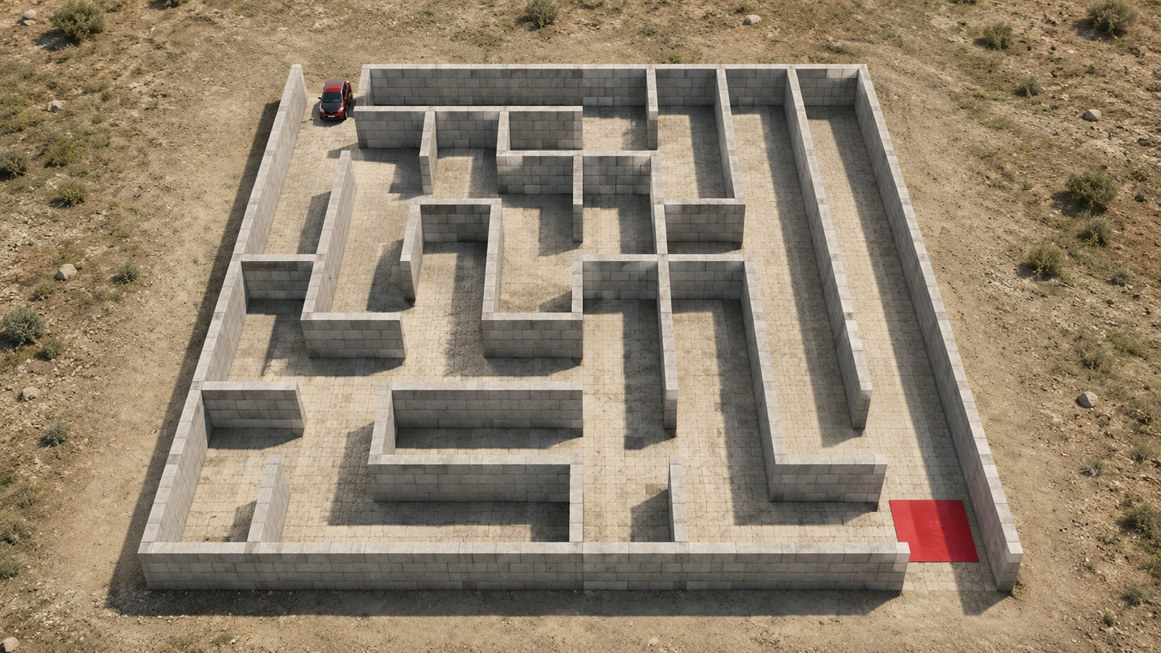}{\textsc{maze\_square}}
  {\vbvrUp{29.0}{87.8}{58.7}}{\vbvrtask{maze}}
\vbvrPairBox[title=\textbf{Overlap}]{Green}{GreenLight}{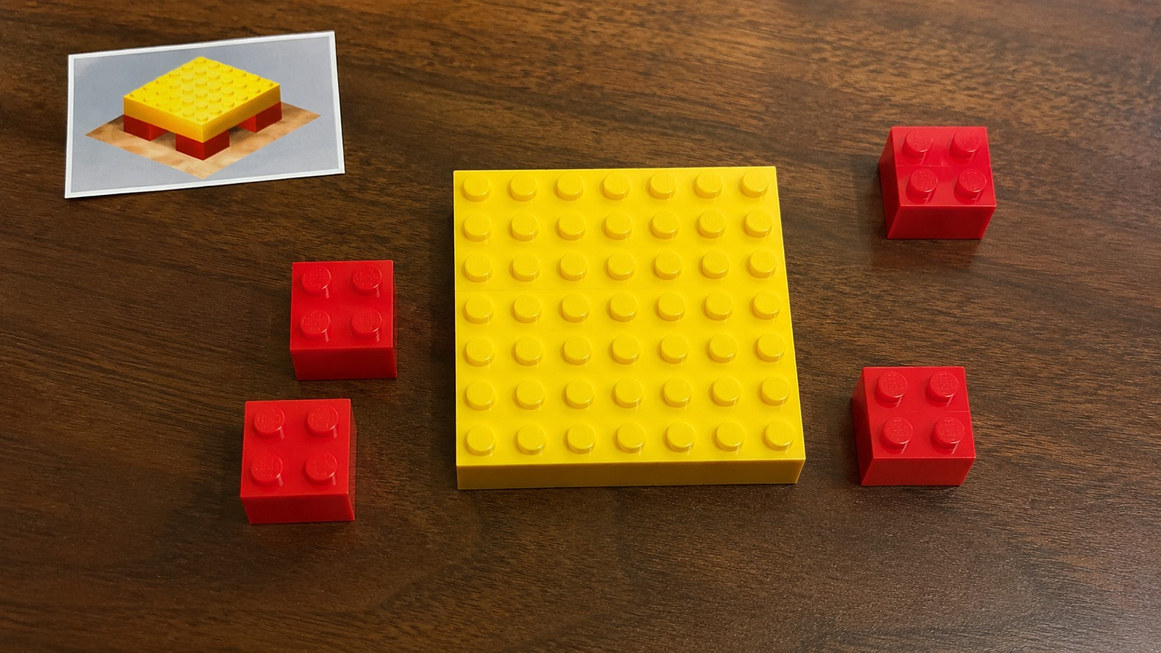}{\textsc{block\_assembly}}
  {\vbvrUp{7.1}{51.7}{44.6}}{\vbvrtask{LEGO\_\allowbreak construction\_\allowbreak assembly}}
\vbvrPairBox[title=\textbf{Overlap}]{Green}{GreenLight}{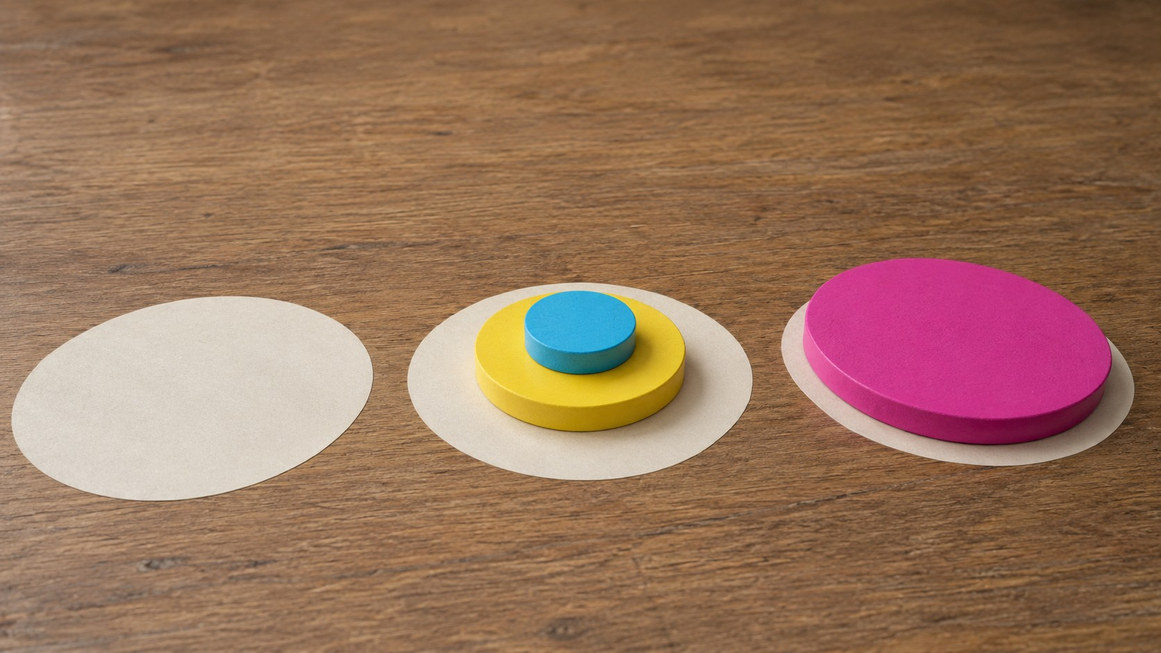}{\textsc{hanoi\_tower}}
  {\vbvrUp{3.2}{52.9}{49.6}}{\vbvrtask{construction\_\allowbreak stack}}
\vbvrPairBox[title=\textbf{Overlap}]{Green}{GreenLight}{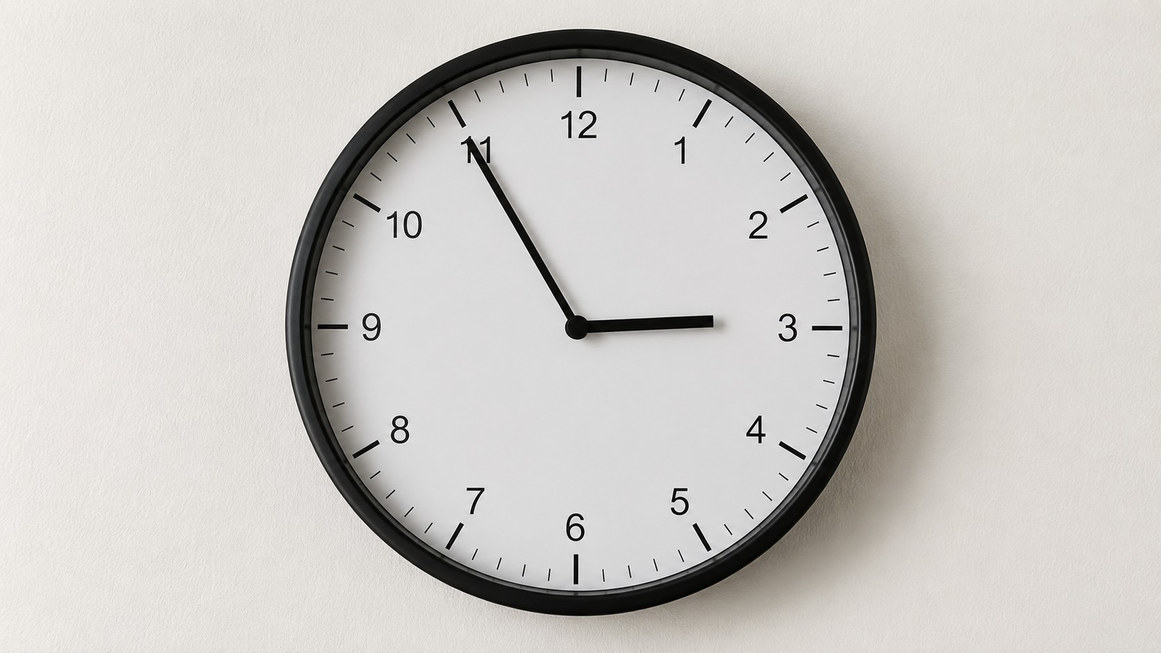}{\textsc{clock\_running}}
  {\vbvrDn{32.5}{10.0}{22.5}}{\vbvrtask{clock}}
\medskip

\noindent
\vbvrPairBox[title=\textbf{Semi-overlap}]{Yellow}{YellowLight}{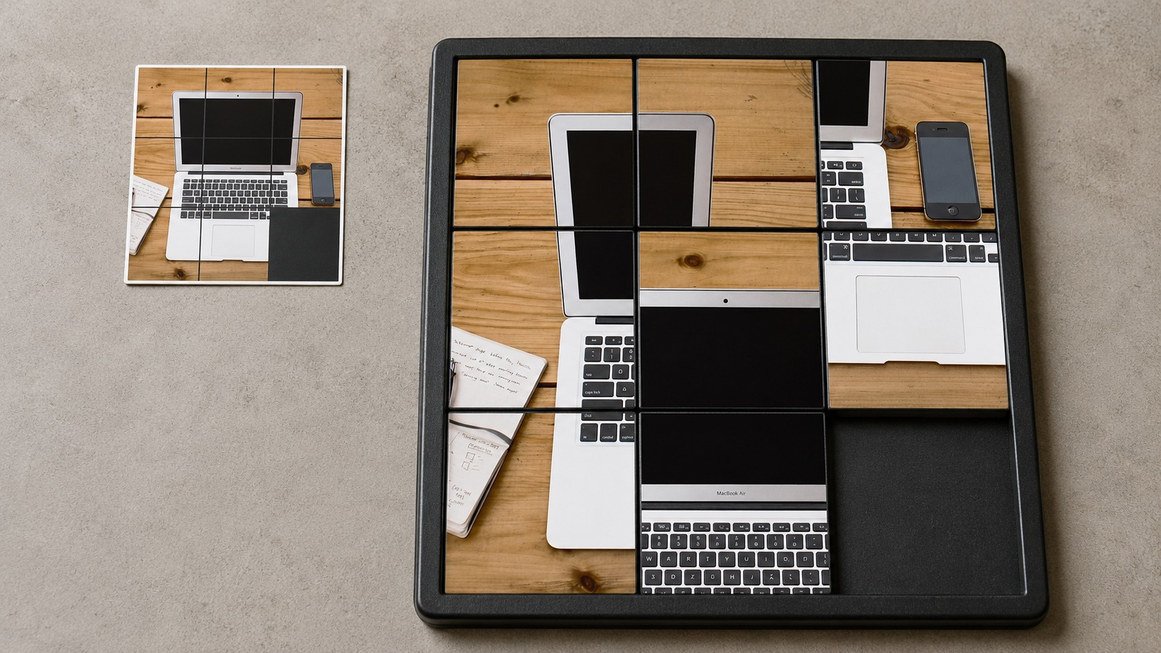}{\textsc{sliding\_puzzle}}
  {\vbvrUp{0.0}{7.6}{7.6}}{\vbvrtask{sliding\_\allowbreak puzzle}}
\vbvrPairBox[title=\textbf{Semi-overlap}]{Yellow}{YellowLight}{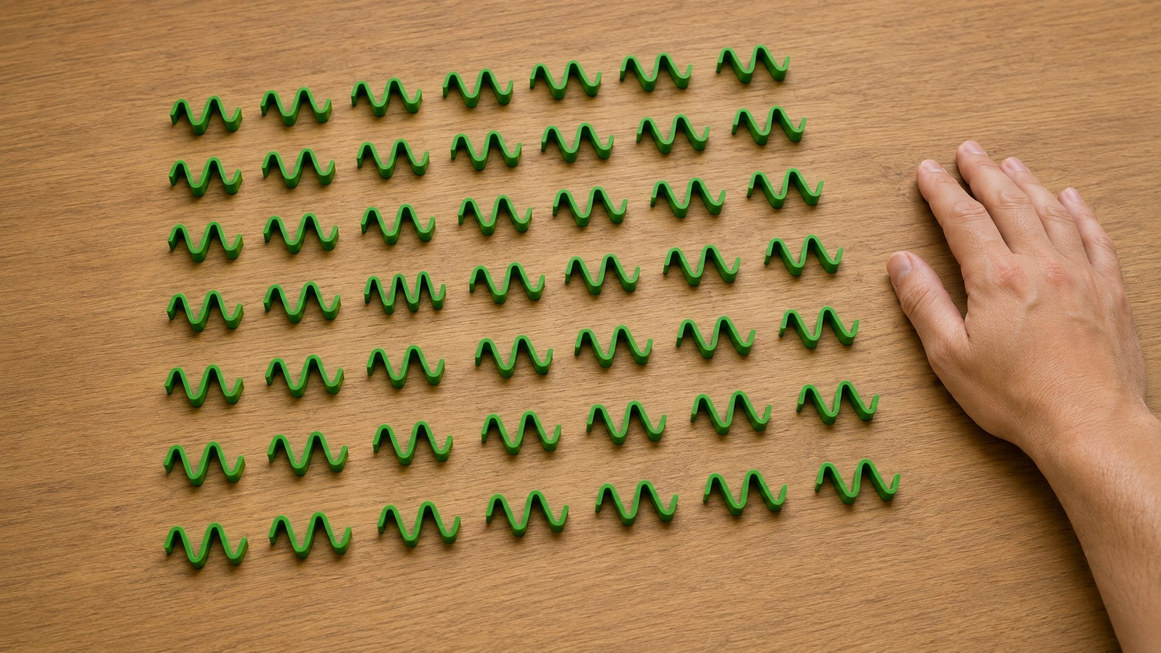}{\textsc{pick\_unique}}
  {\vbvrUp{53.9}{55.0}{1.1}}{\vbvrtask{spot\_\allowbreak unique\_\allowbreak non\_\allowbreak repeated\_\allowbreak color}}
\vbvrPairBox[title=\textbf{Semi-overlap}]{Yellow}{YellowLight}{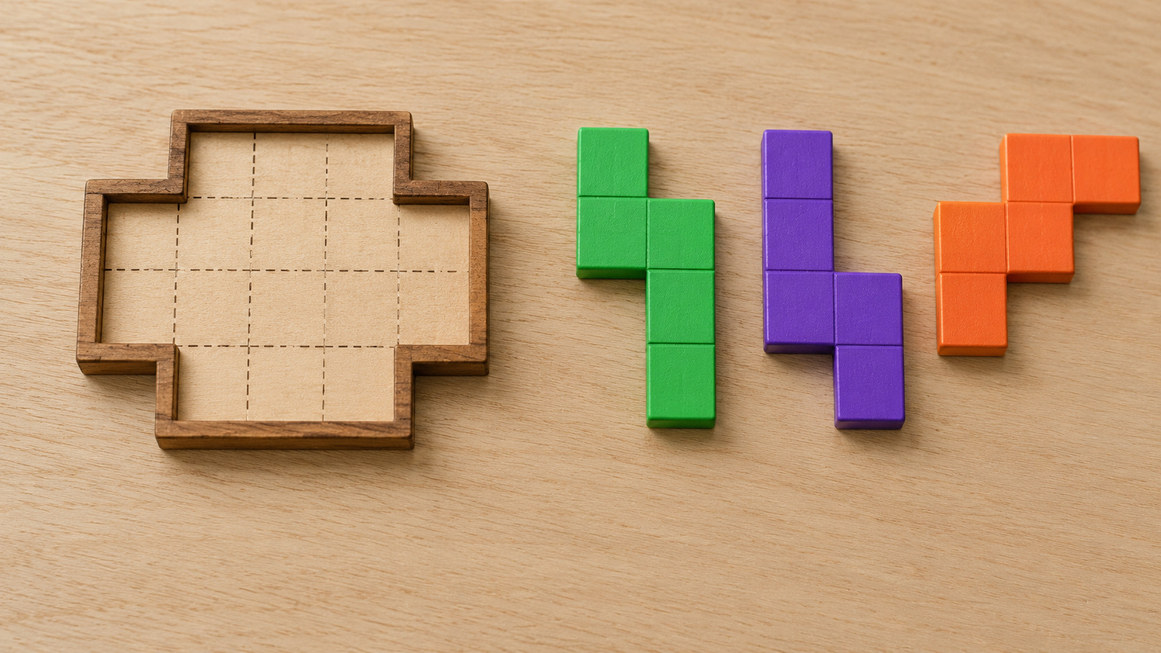}{\textsc{polyform\_tiling}}
  {\vbvrUp{15.8}{49.2}{33.3}}{\vbvrtask{construction\_\allowbreak blueprint}}
\medskip

\noindent
\begin{tcolorbox}[
  enhanced, boxrule=0.9pt, colframe=Red, colback=RedLight,
  arc=4pt, left=3pt, right=3pt, top=3pt, bottom=3pt,
  title=\textbf{Non-overlap}, coltitle=white, colbacktitle=Red,
  fonttitle=\bfseries\small
]
  \centering
  \vbvrSoloCell{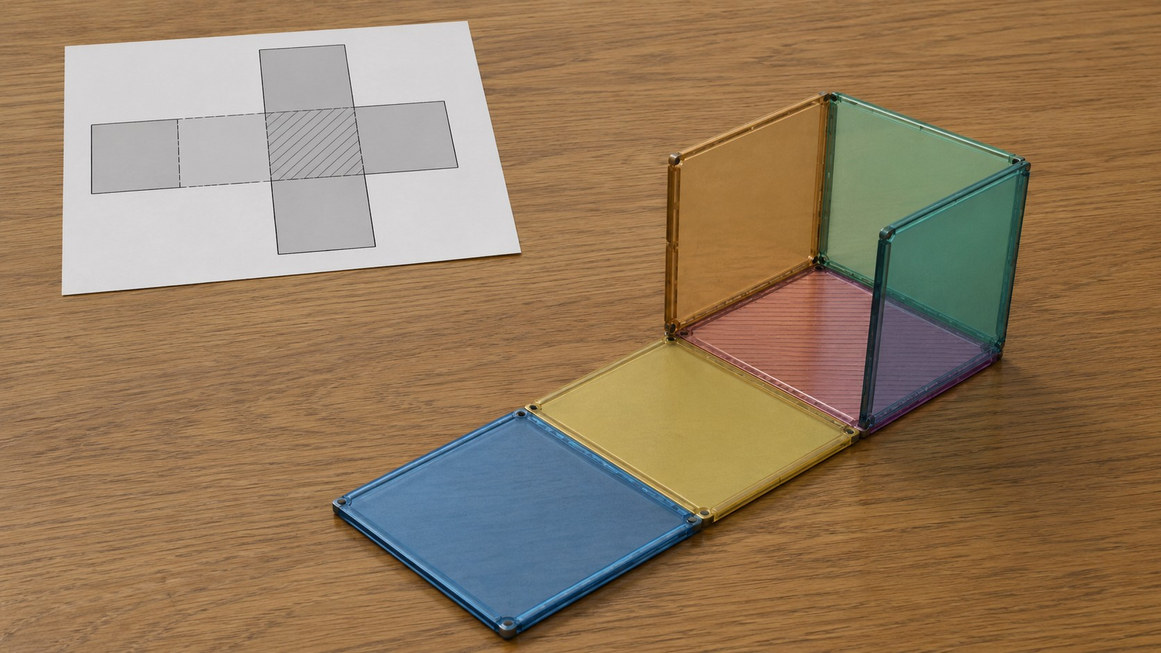}{\textsc{expand\_3d\_to\_2d}}
    {\vbvrUp{16.3}{16.7}{0.4}}\hfill
  \vbvrSoloCell{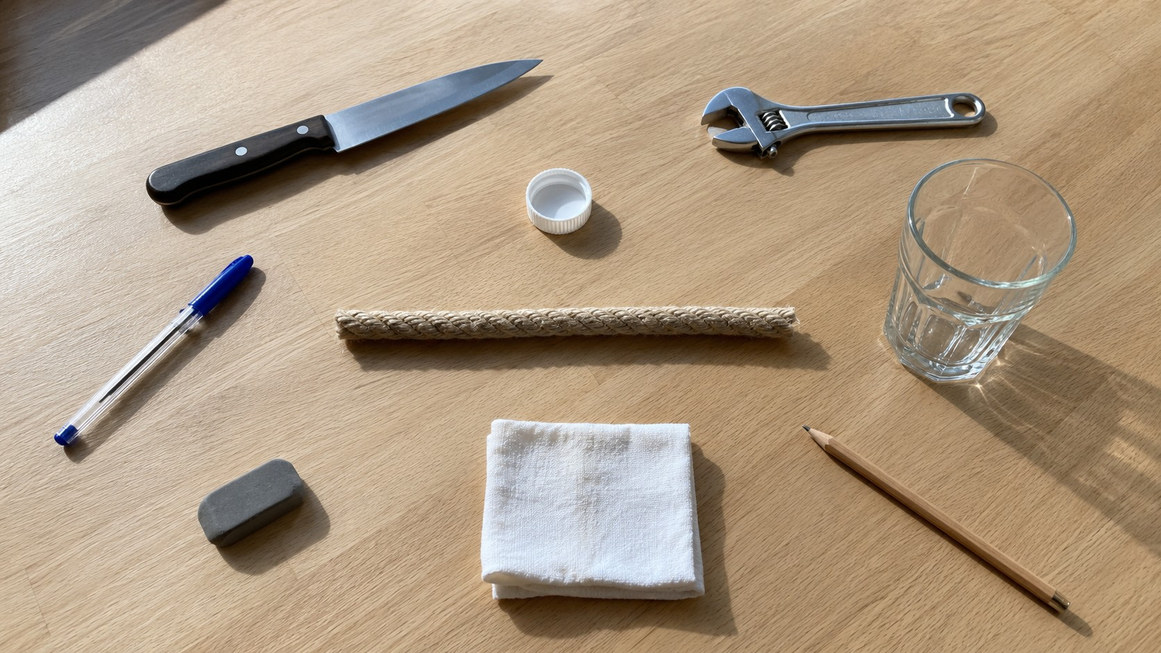}{\textsc{tool\_use\_common}}
    {\vbvrDn{58.8}{44.0}{14.8}}\\[4pt]
  \vbvrSoloCell{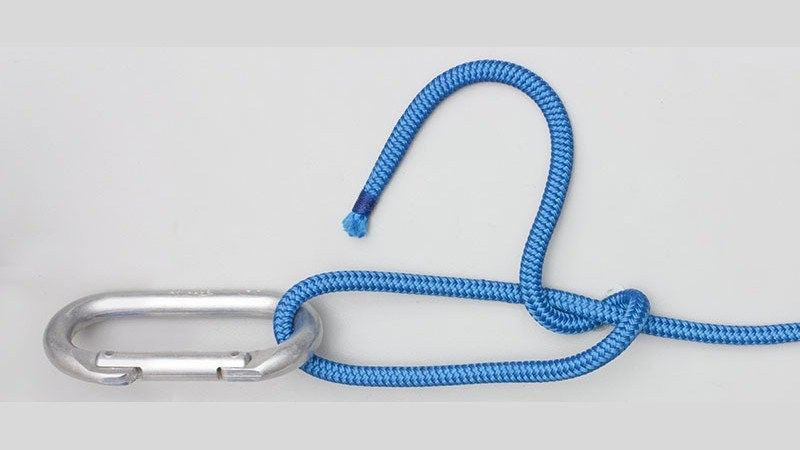}{\textsc{untie\_knot}}
    {\vbvrUp{1.8}{34.0}{32.2}}\hfill
  \vbvrSoloCell{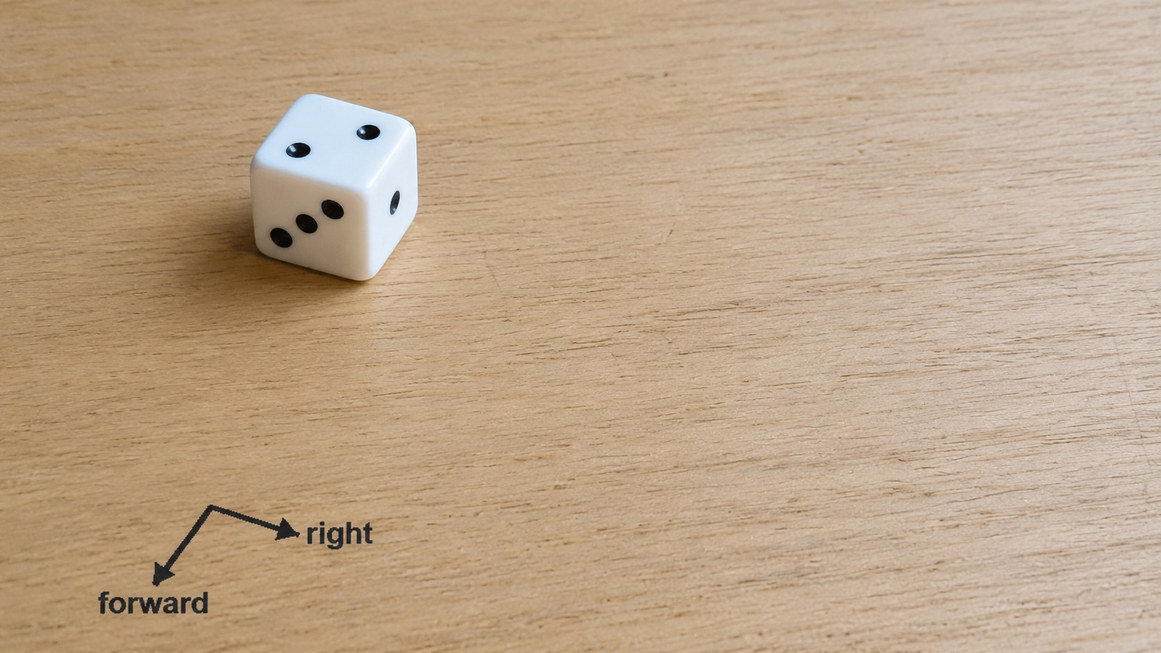}{\textsc{rolling\_dice}}
    {\vbvrDn{30.6}{22.2}{8.4}}
\end{tcolorbox}
\medskip

At the group level, the average improvement decreases from overlap to semi-overlap and then to non-overlap tasks, as shown in Table~\ref{tab:vbvr_sft_res}, consistent with the intuition that training on structurally aligned examples helps models follow task rules, reach the intended goal, and avoid rubric violations. Smaller gains or occasional degradation are expected when the required structures and skills are weakly represented in the training data. 

However, this aggregate trend does not hold uniformly at the task level: some overlap tasks still degrade, while some non-overlap tasks improve. Degradation on overlap tasks may arise from the remaining synthetic-to-real domain gap or execution requirements that are not captured by structural similarity alone. Meanwhile, the improvements on non-overlap tasks often come from more controlled and less aggressive generation after fine-tuning, which reduces large-scale rule violations observed in the base model. Basic spatial and logical capabilities learned during fine-tuning may also transfer to tasks with different surface structures. A representative example here is \textsc{untie\_knot}: although its success rate changes little, the metric score ($final\_score = completeness \times rubric\_score$) increases substantially after fine-tuning (from 0.02 to 0.34) since the generated actions become more restrained and controllable.

% \input{Tables/apdx_vbvr_sft_per_domain}

% Overall, large-scale fine-tuning on synthetic data shows substantial potential for improving performance on realistic downstream tasks. Compared with its base model, \texttt{VBVR-Wan2.2} improves the average final score from \textbf{22.3} to \textbf{40.7}, corresponding to an absolute gain of \textbf{+18.4} points and a relative improvement of \textbf{82.5\%}. On our leaderboard, it ranks third in average final score, behind only \texttt{Seedance 2.0} and \texttt{Kling 3.0}, while outperforming several strong closed-source models, including \texttt{Sora 2}, \texttt{Veo 3.1}, and \texttt{Wan 2.7}. This result is particularly encouraging because the fine-tuning data are entirely synthetic and abstract in style, whereas our benchmark evaluates realistic downstream scenarios. It suggests that scaling readily available and cost-controllable synthetic data can provide an effective path toward improving the reasoning performance of video generation models beyond the synthetic training domain. However, the gains remain uneven across tasks and skills, and transfer is strongly bounded by the structural and capability coverage of the training distribution. Future work may investigate more effective scaling strategies, expand the coverage of underrepresented skills, characterize the limits of cross-domain transfer, and better understand the mechanisms underlying these gains.

\subsection{Reasoning along Denoising Trajectory}
\label{apdx:denoising}

This section gives the protocol behind Table~\ref{tab:denoise_transition} of Section~\ref{5_reasoning_denoising}, which quantifies how often a model revises and potentially corrects its solution states during the denoising process.

\paragraph{Operational definition.}
Self-correction is only meaningful if we can say what the model's current answer \emph{is} at an intermediate step, so we define it over decoded intermediates rather than over latents. For an open-source model we decode the video at several intermediate denoising steps and compare consecutive decoded checkpoints pairwise. Each earlier$\to$later pair is assigned exactly one label. \textbf{\catOne\ Unrecognizable}: at least one of the two videos is too unresolved to read off a solution state, which is concentrated in the earliest steps. \textbf{\catTwo\ Stable}: the solution state is readable and does not change. \textbf{\catThree\ Changed}: the solution state changes, sub-divided into (a) correct\,$\to$\,wrong, (b) wrong\,$\to$\,correct, and (c) wrong\,$\to$\,another wrong solution; only (b) counts as self-correction.

Labelling the \emph{transition} rather than the endpoint state is what separates the two failure narratives: a model that never revises and a model that revises into another wrong answer both end up incorrect, but only the latter is evidence of the search-like behavior reported for dLLMs.

\paragraph{Setup.}
We evaluate four open-source models under their default generation settings ($40$ denoising steps): \texttt{Wan2.2-I2V}, \texttt{VBVR-Wan2.2}, \texttt{HunyuanVideo-1.5}, and \texttt{LTX2.3}. Because early denoising carries the large structural decisions while later steps mainly refine appearance, we sample the trajectory non-uniformly and use the step pairs $1\!\to\!2$, $2\!\to\!3$, $3\!\to\!4$, $4\!\to\!10$, $10\!\to\!20$, and $20\!\to\!40$. We sample $117$ task instances and decode their intermediate denoising products, giving $117 \times 6 = 702$ annotated video pairs; Table~\ref{tab:denoise_transition} reports the label distribution within each step pair, so every column sums to $100\%$.

\paragraph{Interpretation.}
Two observations follow from the distribution. First, the solution state changes often: outside the unresolved early steps, up to a quarter of all transitions fall into the \emph{changed} categories. Second, these changes are almost never corrections. Wrong\,$\to$\,correct transitions stay at or below $0.9\%$ in every step pair and stop occurring after step~4, the first $10\%$ of the trajectory, whereas wrong\,$\to$\,wrong$'$ remains common well into the middle of denoising. Since these models solve few instances to begin with, a change of state is much more likely to move between two wrong solutions than to reach the right one. The answer is largely settled within the first few steps, and the remaining steps refine it.

\clearpage
\section{Other Details}
\label{apdx:other_details}

\subsection{Detailed Comparison with Related Works}
\label{apdx:bench_comparisons}

Table~\ref{tab:related_works_apdx} (same as Table~\ref{tab:related_works}) compares related video benchmarks along three desiderata central to our motivation: \boxednum{1} \textbf{input appearance}, \boxednum{2} \textbf{process-sensitivity}, and \boxednum{3} \textbf{difficulty control}. For benchmarks with data released, human inspection is conducted, while for benchmarks that have not publicly released their data (e.g., TiVi-Bench), we take the statements in paper as the source of truth. The three desiderata are discussed in turn below.

\begin{table}[ht]
\centering
\setlength{\tabcolsep}{4pt}
\renewcommand{\arraystretch}{1.15}
\resizebox{\columnwidth}{!}{%
\begin{tabular}{l c c c c}
\toprule
\textbf{Bench} & \hdr{Reasoning}{Demand} & \hdr{Appearance}{\textcolor{piered}{Abst.}, \textcolor{pieblue}{Real.}} & \hdr{Process-sensitive}{\textcolor{piered}{No}, \textcolor{pieblue}{Yes}} & \hdr{Difficulty}{Control} \\
\midrule
PhysGenBench     & \textcolor{piered}{Low}/\textcolor{pieyellow}{Mid}   & \pie{4}{4}  & \pie{4}{4} & \pmark \\
WorldSimBench  & \textcolor{piered}{Low}/\textcolor{pieyellow}{Mid}     & \pie{4}{4}  & \pie{4}{4} & \cmark \\
TiVi-Bench$^{\dag}$ & \textcolor{pieblue}{High}     & \pie{1}{4}  & \pie{15}{24} & \pmark \\
V-ReasonBench            & \textcolor{pieblue}{High}    & \pie{38}{328}  & \pie{108}{328}  & \xmark \\
VBVR-Bench       & \textcolor{pieblue}{High}        & \pie{0}{4}  & \pie{47}{100}    & \xmark \\
\midrule
\textbf{Ours}    & \textcolor{pieblue}{High}        & \pie{2}{2}  & \pie{2}{2} & \cmark \\
\bottomrule
\end{tabular}%
}

\begin{tablenotes}[flushleft]
\scriptsize
\item \parbox{0.48\textwidth}{
$^{\dag}$Data has not been publicly released, the entries are inferred from the paper.\par
}
\end{tablenotes}
\vspace{-0.5em}

\caption{Each pie encodes fraction of tasks that \textcolor{pieblue}{satisfy a desideratum} versus \textcolor{piered}{do not}.
For input appearance, \textcolor{pieblue}{Real.} denotes photorealistic-style inputs, while \textcolor{piered}{Abs.} denotes non-photorealistic inputs such as line-art or schematic renderings.}
\label{tab:related_works_apdx}
\end{table}

\vspace{-0.7em}

\paragraph{Input Appearance.}
For each dataset, we count the ratio of input images that depict \textbf{realistic scenarios}, as opposed to line-art, schematic, or otherwise abstract styles. PhysGenBench and WorldSimBench are fully realistic; TiVi-Bench contains roughly 1/4 and 1/9 realistic or near-realistic inputs, respectively; VBVR-Bench is entirely script-generated for controllability and scalability, without photographic inputs. Overall, reasoning-heavy video generation benchmarks still largely rely on abstract or line-art inputs, which may deviate from the visual distribution of modern video models. This concern is supported by our style-sensitivity study in Section~\ref{5_sensitivity}: holding the task fixed, switching an instance from a realistic scene to a line-art or low-poly rendering leads to substantially different performance, especially for models with lower overall performance, and abstract variants more often violate task constraints.
% Figures~\ref{fig:style_mismatch_eg1_maze} and~\ref{fig:style_mismatch_eg2_hanoi_tower} illustrate two such per-instance comparisons.

\vspace{-0.5em}

\paragraph{Process-sensitivity.}
We identify a task as process-sensitive if \textbf{it requires a non-trivial sequence of state transitions, regardless of whether the final outcome can be verified from a single frame.} For example, maze solving and Hanoi Tower have final states that are easy to inspect, but a valid video must still reach them through legal intermediate steps. This differs from one-shot visual reasoning tasks, such as Raven-style matrices, visual analogy, or even multiple-choice VQA, where the expected answer can be produced without temporal state evolution.

Under this definition, PhysGenBench and WorldModelBench are effectively process-sensitive, as they aim to evaluate physical coherence and world simulation. TiVi-Bench contains $15$ process-sensitive tasks out of $24$, while the remaining are closer to one-shot reasoning. For V-ReasonBench ($\sim$33\%) and VBVR-Bench (47\%), the portion of both parts are roughly comparable. In our benchmark, tasks are designed with explicit state transitions, and the rubrics inspect intermediates and transition validity (Section~\ref{3_benchmark}), making shortcut-to-final-state behavior detectable as a violation.

\vspace{-0.5em}

\paragraph{Difficulty Control.}
Task difficulty would become less interpretable when tasks are either saturated by current models or far beyond their feasible regime. For example, long-horizon tasks may require trajectories exceeding the practical generation duration, while knowledge-heavy tasks may rely on advanced scientific, cultural, or historical expertise rather than visually grounded reasoning. Although such tasks can serve as stress tests, diagnostic benchmarks should also produce informative differences among current models. We therefore consider two criteria to control difficulty: \ding{172} \textbf{feasibility calibration}, where we verifies that tasks can be meaningfully attempted by current video models without failures dominated by excessive duration or non-visual expertise; and \ding{173} \textbf{multi-level difficulty design}, where tasks are organized into explicit levels to support graded performance analysis as procedural complexity increases.

Under these criteria, PhysGenBench and WorldSimBench are treated as difficulty-calibrated since their tasks primarily target physical world simulation, with prompts manually reviewed for feasibility and clarity. PhysGenBench is marked as ``partial'' for not providing explicit difficulty splits. In contrast, reasoning-heavy benchmarks like TiVi-Bench, V-ReasonBench, and VBVR-Bench do not include a model-facing feasibility calibration stage; among them, only TiVi-Bench provides multi-level difficulty design. Our benchmark satisfies both criteria through pre-generation and manual review for feasibility calibration, together with explicit multi-level construction. All above are shown in the ``Difficulty Ctrl.'' column of Table~\ref{tab:related_works_apdx}.

\clearpage
\section{Submission Checklist}
\label{apdx:submission_checklist}

\subsection{Potential Risks}
The potential risks are minimal. Our benchmark does not involve sensitive domains, personal data, or identity-related attributes. Human studies only collect anonymized task responses and preference annotations. We also screen task descriptions, prompts, and visual contents to avoid offensive or sensitive material.

\subsection{Licenses and Terms of Use}

We use both commercial video generation APIs and open-source model weights in our experiments. For commercial APIs, we follow the corresponding provider terms of service and use the generated outputs only for research evaluation. For open-source models and tools, we use them under their released licenses and cite the original creators. We do not redistribute third-party model weights or proprietary API outputs beyond what is permitted by their terms.

For the benchmark artifact created in this work, we will release the data and evaluation code under a research-friendly license, with the intended use limited to research evaluation and diagnostic analysis of video generation models.

\subsection{Artifact Use Consistent With Intended Use}
We use existing artifacts, including video/image generation models, VLMs, evaluation tools, and packages like FFmpeg, only for research evaluation and data processing, following their intended use and applicable terms. We do not use them for commercial training, individual profiling, or real-world decision-making.

Our benchmark is intended for academic research and diagnostic evaluation of procedural and visual reasoning in video generation models. Human annotations are used only for estimating human-ceiling performance and validating VLM-as-Judge, and are anonymized before analysis.

\subsection{Personally Identifying Information and Offensive Content}

Our benchmark is constructed from task-level metadata and synthetic visual instructions, and does not include personal data or real-world records that identify individuals. For human studies, we collect only task responses for the human-ceiling evaluation and preference annotations for validating the reliability of VLM-as-Judge. We do not collect names, contact information, demographic attributes, or other sensitive personal information. All human annotations are anonymized and reported only in aggregate. We manually inspect task descriptions, prompts, and visual contents to remove offensive, hateful, sexually explicit, or otherwise sensitive content.

\subsection{Human Annotation Recruitment, Payment, and Data Consent}
The human annotations were provided by the paper co-authors as part of the research process. We did not recruit external participants through crowdsourcing platforms or student pools, and no separate compensation was provided. All annotators were aware that their task responses and preference annotations would be used for benchmark validation, including estimating human-ceiling performance and validating VLM-as-Judge reliability.

\subsection{Ethics Review}
The human-ceiling evaluation and preference annotations were conducted by the paper co-authors as part of the research process. We did not obtain formal ethics review board approval or exemption. No external participants were recruited, and we did not collect names, contact information, demographic attributes, or other sensitive personal information. The annotations were used only for benchmark validation and reported in aggregate.

\subsection{Information About Use Of AI Assistants}

We used AI assistants for grammar checking, language polishing, and debugging assistance. The authors reviewed and verified all technical content, experimental design, analyses, and final manuscript decisions.

\subsection{Package Usage and Parameters}

We use FFmpeg for video preprocessing. Specifically, we extract frames from each generated video at a fixed frame rate using the following command:

\begin{center}
\texttt{ffmpeg -i input.mp4 -vf fps=\textit{n} frames/\%04d.png}
\end{center}

\noindent
This extracts \textit{n} frames per second (we use $n=2$, $4$, or $8$ depending on the judging pass, see Appendix~\ref{apdx:vlm_as_judge}) and saves them as PNG. No additional normalization or filtering is applied.

\end{document}